\documentclass[11pt, a4paper]{article}
\usepackage{latexsym}
\usepackage[fleqn]{amsmath}
\usepackage{graphicx}
\usepackage[fleqn]{amsmath}
\usepackage[varg]{txfonts}
\usepackage[english]{babel}
\usepackage[hyperindex=false,
    pageanchor=false,
    pdfmenubar=false,
    pdfpagelabels=false, breaklinks=true]{hyperref}
\usepackage{apacite} 
\usepackage{setspace}
\usepackage{epstopdf}
\usepackage{setspace}
\usepackage{threeparttable}
\newcommand{\mb}{\mathbb} % Must use capital letter \mb{A} not \mb{a}
\newcommand{\vc}{\mathbf} % use smale letter\mbf{a}
\newcommand{\gvc}{\boldsymbol}

\newcommand{\mc}{\mathcal}

\newcommand\scalemath[2]{\scalebox{#1}{\mbox{\ensuremath{\displaystyle #2}}}}

\begin{document}

%\title{ Influence function of  Mean Element, Covariance operator and  Kernel Canonical Correlation Analysis}         
\title{Kernel Method for Detecting Higher Order Interactions   in   multi-view Data: An Application to Imaging, Genetics, and Epigenetics} %and its Influence Function}% for Imaging Genetics Analysis}   
\author{\textbf{  Md. Ashad Alam$^{1}$, Hui-Yi Lin$^{2}$,   Vince Calhoun$^{3}$ and Yu-Ping Wang$^{1}$}  \\
 $^{1}$Department of Biomedical Engineering, Tulane University\\
 New Orleans, LA 70118, USA\\
%$^2$Department of Statistics, Hajee Mohammad Danesh Science and Technology\\ University, Dinajpur 5200, Bangladesh\\
 $^{2}$ Biostatistics Program, Louisiana State University Health Sciences Center\\
 New Orleans, LA 70112, USA\\
$^3$Department of Electrical and Computer Engineering, The University of New Mexico\\ Albuquerque, NM 87131, USA}
    %Enter your name between curly braces
       % Enter your date or \today between curly braces
\date{}
\maketitle
\begin{abstract}
Technological advances are enabling us to collect multiple types of data at an  increasing depth and resolution while decreasing the labor needed to compile and analyze it.  A central goal of multimodal data integration is to understand the interaction effects of different features. Understanding the complex interaction among multimodal datasets, however, is challenging. In this study, we tested the interaction effect of multimodal datasets using a novel method called the kernel method for detecting higher order interactions among biologically relevant mulit-view data. Using a semiparametric method on a reproducing kernel Hilbert space (RKHS), we used a standard mixed-effects linear model and derived a score-based variance component statistic that  tests for  higher order interactions between multi-view data.  The proposed method offers an intangible framework for the identification of higher order  interaction effects (e.g., three way interaction) between genetics, brain imaging, and epigenetic data. Extensive numerical simulation studies were first conducted to evaluate the performance of this method.  Finally, this method was evaluated using data from the Mind  Clinical Imaging Consortium (MCIC) including single nucleotide polymorphism (SNP) data, functional magnetic resonance imaging (fMRI) scans, and deoxyribonucleic acid (DNA) methylation data, respectfully, in schizophrenia patients and healthy controls. We treated each gene-derived SNPs, region of interest (ROI) and gene-derived DNA methylation as a single testing unit, which are combined into triplets for evaluation. In addition, cardiovascular disease risk factors such as age, gender, and body mass index were assessed as covariates on hippocampal volume  and compared between triplets. Our method identified  $13$-triplets ($p$-values $\leq 0.001$) that included  $6$ gene-derived SNPs,  $10$  ROIs, and $6$ gene-derived DNA methylations that correlated with changes in hippocampal volume, suggesting that these triplets may be important in explaining  schizophrenia-related neurodegeneration. With strong evidence ($p$-values $\leq 0.000001$), the triplet ({\bf MAGI2, CRBLCrus1.L, FBXO28}) has the potential to distinguish  schizophrenia  patients from the    healthy control variations.  This novel method may shed light on other disease processes in the same manner, which may benefit from this type of multimodal analysis.
\end{abstract}  
keywords:  Multimodal data,  Higher order interaction,  Kernel methods,   Imaging genetics, Imaging epigenetics, and Schizophrenia.
   
\section{Introduction}
\label{sec:Intro}
The advancements in data science technology over the last decade has rapidly evolved to collect multi-view data, which has emerged to provide a comprehensive way to explore statistical structures and information embedded in the relationship between datasets.  The integration of imaging and genetic information into a format capable of predicting disease phenotypes, however, continues to be challenging problem.

\begin{figure}
\begin{center}
\includegraphics[width=10cm, height=8cm]{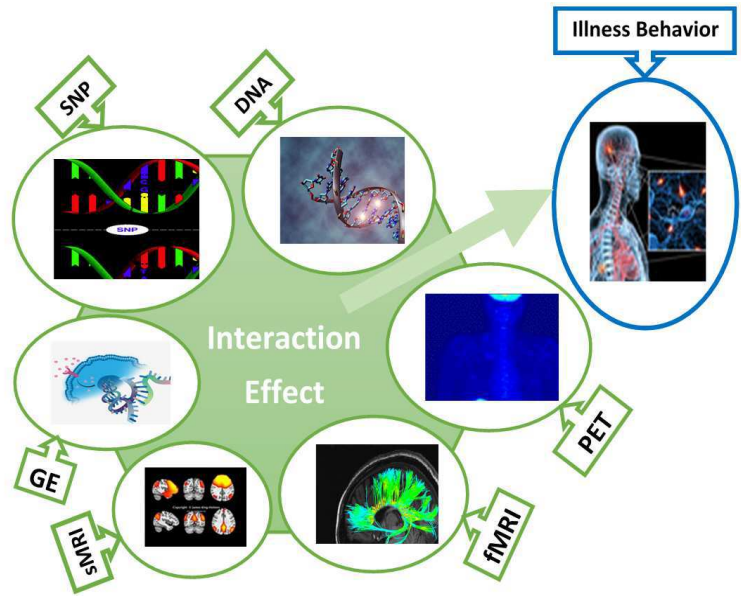}
\caption{An illustration of  the different  imaging genetics and epigenetics  data along  with their interaction effect on human  behavior. Note, SNP: single nucleotide polymorphism, DNA: deoxyribonucleic acid methylations, PET: positron emission tomography (PET),  fMRI: functional magnetic resonance imaging (fMRI),  sMRI: structural MRI, GE: gene expression.}
\label{fig:mvd}
\end{center}
\end{figure}

% 2nd para Imaging genetis with figure
One of the goals of  imaging genetics is the modeling and understanding  of how genetic  variations influence the structure and function of  brain disease. This goal can be achieved by collating multimodal data including functional magnetic resonance imaging (fMRI), structural MRI (sMRI), and  positron emission tomography (PET) scans with single nucleotide polymorphisms (SNPs), deoxyribonucleic acid  (DNA) methylations, gene expression (GE),  transcriptomics, epigenomics, and proteomics factors. Numerous studies have suggested that these different factors do not act in isolation, but rather they interact at multiple levels and depend on one another in an intertwined manner 
\cite{Calhoun-16, Pearlson-15}.  Extracting the interaction effects from within and among data sets, however, remains a challenge for multi-view data analysis \cite{Li-15, Chekouo-16, Zhen-15, Zhao-16, Liu-16}. Figure~\ref{fig:mvd} illustrates how the interaction effects of different data sets can be used to model and predict human illness.

% 3rd  advantages of  Imaging genetis  data From imagilys.com.
To date, both  genetic  techniques and brain imaging have played a substantial role  in  detecting  disease phenotypes.  For example, by correlating imaging and  genetic data, it has been shown that certain genes affect specific brain functions, connectivity, and serve as risk predictors for certain diseases. \cite{Jahanshad-12, Dongdong-14, Bis-12, Jahanshad-13}. Additionally, \cite{Bis-12} have identified genetic variants affecting the volume of the hippocampus, which could be used as predictors of cognitive decline and dementia \cite{Jahanshad-13}. As shown in \cite{Wen-17}, accurate identification of Tourette's syndrome  in children has notably improved using  multi-view features as compared to relying solely on one view. Accumulating evidence also shows that the inherent genetic variations for complex traits can sometimes be explained by the joint analysis of multiple genetic features with environmental factors.

%data application
Schizophrenia (SZ) is a complex brain disorder that  affects how a person thinks, feels and acts, which is thought to be caused through  an interplay of genetic effects, brain region, and DNA methylation abnormalities \cite{Richfield-17}.  Studies using neurological tests and brain imaging technologies (fMRI and PET) have been used to examine functional differences in brain activity  that  seem to   arise within the frontal lobes, hippocampus and temporal lobes \cite{Van-09, Kircher-05}.  Many researchers have shown that genetic alterations at the mRNA and SNP level, however, also play a significant role in SZ \cite{ Chang-13, Lencz-07}.  Thus, only focusing on  brain imaging data  is not sufficient in the  identification of the related risk factors for SZ \cite{Potkin-15}. To address this,  \cite{Chekouo-16} have developed  the ROI-SNP network  for the selection of discriminatory markers using  brain imaging  and genetics information.  

A number of studies suggest that epigenetics also has a role in  SZ disease susceptibility. Genome-wide DNA methylation  analysis of human brain tissue from SZ patients   shows a heritable epigenetic modification, which  can regulate gene expression. The cell specific differences in chromatin structure that influence cell development, including DNA methylation, have emerged as a potential explanation for the non-Mendelian inheritance of SZ  \cite{Wockner-14}. There is also evidence on epigenetic alterations in the blood and central nervous system of patients with SZ, and it has been shown that methylation status in brain tissue from SZ patients varies  significantly  from controls \cite{Aberg-14, Montano-16}.  In this paper, we consider the interaction effects among the genetics, brain imaging, and epigenetics data on hippocampal volume measurements between SZ patients and healthy controls using a novel kernel method for detecting these higher order interactions.

%4dr machine lerning + kernel mthods 

Many advancements in multimodal fusion methods have utilized such approaches as co-training, multi-view learning, subspace learning, multi-view embedding, and kernel multiple learning, to analyze multi-view data of biological relevance \cite{Xu-13}. However, due to the large number of  genes,  SNPs, DNA methylations and  different types of  imaging,  positive definite  kernel based  methods  have become a popular and effective tool for conducting  genome-wide association studies (GWASs) and  imaging genetics,  especially for identifying genes associated with diseases \cite{Li-12, Ge-15, Alam-16a, Alam-16b}. Kernel methods are emerging as innovative techniques that   map  data from  high dimension input spaces  to a kernel feature space using a nonlinear function. The main advantage of these methods is  to combine statistics and geometry in an effective way \cite{Hofmann-08}. Kernel methods  offer useful algorithms to learn how a large number of genetic variants are associated with complex phenotypes, to help explore the relationship between the genetic markers and the outcome of interest \cite{Camps-07, Yu-11, Ashad-14T, Ashad-15, Schlkof-kpca, Kung-14}.

 %5th Intercetion effiect wiht classical + kernel mthods
In genetics, the  detection of  gene-gene interactions or co-associations in most methods are divided into two types: SNP based and gene-based methods in GWASs. In the last decade, a number of statistical methods have been used to detect gene-gene interactions (GGIs). Logistic regression, multifactor dimensionality reduction, linkage disequilibrium and entropy based statistics are examples of such methods \cite{Hieke-14, Wan-10}.  While most of these methods are based on the unit association of the SNPs, testing the associations between the phenotype and SNPs has limitations and  is not sufficient for interpretation of GGIs \cite{Zhongshang-12}. In GWASs, gene-based methods are always more effective than the ones based  only  on a SNP, and powerful tools for multivariate gene-based genome-wide associations have been proposed \cite{Sluis-15}. 

In recent years,  linear, kernel, and robust canonical correlation based U statistic   have been   utilized  to identify  gene-gene co-associations \cite{Peng-10, Alam-16b}.  \cite{Li-12}  have proposed a model-based kernel machine method for GGIs.  In addition, \cite{Ge-15} have  also proposed a kernel machine method for detecting effects of interactions between multi-variable sets. This is an extended model of \cite{Li-12}  to jointly model the genetics and non-genetic features, and their interactions. While these methods could ultimately shed light on novel features of the etiology of complex diseases, they cannot be reliable used in multi-view data sets.  Thus, there exists a need to extend kernel machine based methods.

% par 7paper contribution
The contribution of this paper, therefore, is threefold. By examining the three-way interaction effects between triplet data sets combining genetics, imaging, and epigenetics, we hope to shed light on the phenotype features associated with disease mechanisms.  This is done iteratively. First,  we propose a novel   semiparametric method on a reproducing kernel Hilbert space (RKHS) to study the interaction effects among the multiple-view datasets. We  name  a  kernel method for detecting higher order interactions (KMDHOI)  and include  the pairwise  and higher order Hadamard product of the  features from different views.  Second, we formulate the problem as a standard mixed-effect linear model to derive a score-based variance component test for the higher order interactions. The proposed method  offers a flexible framework to account for the  main (single), pairwise, triplet, other higher order effects  and test for the overall higher order effects.   Finally,  we validate the proposed method  on both simulation and the Mind  Clinical Imaging Consortium (MCIC) data \cite{Chen-12MCIC, Gollub-13}.    
 
% par 8 organigation 
The remainder of this  paper is organized as follows. In Section~\ref{sec:methods}, we  propose  a standard mixed-effects linear model to derive score-based variance component test for higher order interaction.  In Section~\ref{sec:test}, we propose statistical testing for  higher order interaction effects. The  relevant methods are discussed in Section~\ref{sec:comt}.  In Section~\ref{sec:exp}, we describe  the experiments conducted on both synthesized  and  the imaging genetics data sets. We conclude the paper with a discussion of major findings and future research in Section~\ref{sec:cond}. Details of the  theoretical analysis for the proposed method, Satterthwaite approximation to the score test,  and supplementary tables and figures on   application to  imaging genetics and  epigenetics can be found in the appendix.

\section{Method}
\label{sec:methods}
In kernel methods, the nonlinear feature map is given by a {\bf positive definite kernel}, which provides nonlinear methods for data analysis.   It is known \cite{Aron-RKHS} that a positive definite kernel $k$ is associated with a Hilbert space $\mc{H}$, called {\bf reproducing kernel Hilbert space} (RKHS), consisting of functions on $\mathcal{X}$ so that the function value is reproduced by the kernel; namely,
for any function $f\in \mc{H}$ and  a point $X\in\mathcal{X}$, the function value $f(X)$ is $
f(X)=\langle f(\cdot),k(\cdot, X)\rangle_{\mc{H}},$ where $\langle ,\rangle_{\mc{H}}$ in the inner product of $\mc{H}$  is called the reproducing property.  Replacing $f$ with $k(\cdot,\tilde{X})$ yields $
k(X,\tilde{X})=\langle k(\cdot, X), k(\cdot, \tilde{X})\rangle_{\mc{H}}$ for any $X,\tilde{X} \in \mc{X}$.  A symmetric kernel $k(\cdot, \cdot)$ defined on a space $\mathcal{X}$ is called {\bf positive definite}, if  for an  arbitrary number of points $X_1,\ldots, X_n\in\mathcal{X}$ the Gram matrix $(k(X_i, Y_j))_{ij}$ is positive semi-definite.  To transform data for extracting nonlinear features, the mapping $\vc{\Phi}: \mc{X}\to \mc{H}$ is defined as $\vc{\Phi}(X)=k(\cdot, X),$
which is  a function of the first argument. This map is called the  {\bf feature map}, and the vector $\vc{\Phi}(X)$ in $\mc{H}$ is called  the {\bf feature vector}.  The inner product of two feature vectors is then $\langle \vc{\Phi}(X) ,\vc{\Phi}(\tilde{X} )\rangle_{\mc{H}}=k(X, \tilde{X}).$
This is known as the {\bf kernel trick}. By this trick the kernel can evaluate the inner product of any two feature vectors efficiently without knowing an explicit form of $\vc{\Phi}(\cdot)$.
 
\subsection{Model setting}
\label{sec:promethd}
Assuming that we have $n$ independent identical distributed (IID) subjects $y_i$ $(i = 1, 2, \cdots, n)$ with $(q-1)$ covariates $X_i= [X_{i1}, X_{i2}\cdots X_{i(q-1)}]^T$ and m-view datasets, $\vc{M}_{i}^{(1)}, \cdots, \vc{M}_{i}^{(m)}$. In the following  semiparametric model,  we associate the output $y_i$ with covariates including intercept and $m$-view datasets:
\begin{eqnarray}
\label{me1}
y_i= X_i^T\vc{\gvc{\beta}} + f(\vc{M}_{i}^{(1)}, \cdots, \vc{M}_{i}^{(m)}) + \epsilon_i,
\end{eqnarray}
 where  $X_i$ is a $q\times 1$ vector of covariates including intercept for the $i-$th subject,  $\gvc{\beta}$ is  a $q\times 1$ vector of fixed effects,  $f$ is an unknown function on the product domain, $\mc{M}=\mc{M}^{(1)} \otimes \mc{M}^{(2)}\otimes, \cdots, \otimes \mc{M}^{(m)}$ with  $\vc{M}_{i}^{(\ell)} \in \mc{M}_\ell, \ell= 1, 2, \cdots m$ and the error $\epsilon_i$'s are IID as normal with mean zero and  variance $\sigma^2$, $\epsilon_i\sim \rm{NIID}(0, \sigma^2)$.  According to the ANOVA decomposition, the function, $f$ can be extended  as:
\begin{multline}
\label{me2}
f(\vc{M}^{(1)}_{i}, \cdots, \vc{M}^{(m)}_{i}) = \sum_{\ell=1}^m h_{\vc{M}^{(\ell)}}(\vc{M}_{i}^{(\ell)}) +  \sum_{\ell_1 > \ell_2}h_{\vc{M}^{(\ell_1)}, \vc{M}^{(\ell_2)}} (\vc{M}^{(\ell_1)}_{i}, \vc{M}^{(\ell_2)}_{i}) +\\ \sum_{\ell_1 > \ell_2> \ell_3}h_{\vc{M}^{(\ell_1)}\times \vc{M}^{(\ell_2)}\times \vc{M}^{(\ell_3)}} (\vc{M}^{(\ell_1)}_{i}, \vc{M}^{(\ell_2)}_{i},  \vc{M}^{(\ell_3)}_{i}) + \cdots + h_{\vc{M}^{(1)}\times \vc{M}^{(2)}\times \vc{M}^{(3)} \times \cdots, \times \vc{M}^{(m)}} (\vc{M}^{(1)}_{i}, \vc{M}^{(2)}_{i}, \cdots, \vc{M}^{(m)}_{i}), 
\end{multline}
where   $h_{\vc{M}^{(\ell)}} (\vc{M_i}^{(\ell)})$'s ($\ell: 1, 2, \cdots m$) are the  main effects for the  respective dataset,  $h_{{\vc{M}^{(\ell_1)}}, {\vc{M}^{(\ell_2)}}}(\vc{M}_i^{(\ell_1)},\vc{M}_i^{(\ell_2)})$  are pairwise interactions effects, $h_{{\vc{M}^{(\ell_1)}}\times {\vc{M}^{(\ell_2)}}, {\vc{M}^{(\ell_3)}}}(\vc{M}_i^{(\ell_1)},\vc{M}_i^{(\ell_2)},\vc{M}_i^{(\ell_3)})$  are the interactions effects of the three dataset and so on.  The functional space, RKHS,  is decomposes as: 
\begin{multline}
\label{me3}
 \mc{H}= \mc{H}_{\vc{M}^{(1)}}\oplus\mc{H}_{\vc{M}^{(2)}} \oplus\cdots\oplus\mc{H}_{\vc{M}^{(m)}}\oplus\mc{H}_{\vc{M}^{(1)}\times \vc{M}^{(2)}}\oplus\mc{H}_{\vc{M}^{(1)}\times \vc{M}^{(3)}} \oplus\cdots\oplus\mc{H}_{\vc{M}^{(1)}\times \vc{M}^{(m)}}\oplus\mc{H}_{\vc{M}^{(2)}\times \vc{M}^{(3)}} \\ \oplus\cdots\oplus \mc{H}_{\vc{M}^{(1)}\times \vc{M}^{(2)}\times \vc{M}^{(3)}}  \oplus\cdots\oplus \mc{H}_{\vc{M}^{(1)}\times \vc{M}^{(2)}\times\cdots \times \vc{M}^{(m)}},
\end{multline}
 equipped with an inner product, $\langle \cdot, \cdot\rangle$ and a norm $\|\cdot\|_\mc{H}.$ 
If $m=1$,  Eq. (\ref{me1}) becomes simple semiparametric regression model as shown in \cite{Liu-07}. \cite{Li-12} and \cite{Ge-15} have proposed similar models (special case of Eq. (\ref{me1}), $m=2$) for   detecting interaction effects among  multidimensional variable sets.

 Specifically, in our case we have three data sets.  To do this,   we assume that  we have   $n$ IID subjects under investigation; $y_i\, (i =1, 2, \cdots n)$ is a quantitative phenotype for the $i$-th subject (say, hippocampal volume derived from structural MRI scan). We  associate the clinical covariates (e.g., age,  weight, height) with three views: genetics, imaging, and epigentics (gene-derived SNP, ROIs, and gene-derived DNA methylation). Let  $\vc{X}_i$ denote the  $(q-1)$ covariates, where $X_{ij}, j=1, 2, \cdots (q-1)$ is a measure of  the $i$-th subject. Let   $M_i^{(1)}=[M_{i1}^{(1)}, M_{i2}^{(1)}, \cdots,  M_{is}^{(1)}]$,  $M_i^{(2)}=[M_{i1}^{(2)}, M_{i2}^{(2)}, \cdots,  M_{is}^{(2)}]$ and $M_i^{(3)}=[M_{i1}^{(3)}, M_{i2}^{(3)}, \cdots,  M_{is}^{(3)}]$   be a genes-derived SNP with $s$ SNP markers,  a ROI with $r$ voxels of the fMRI scan,  and a gene-derived DNA methylation  with  $d$ methylation profiles of the $i$-th subject, respectively. Under this setting,  Eq. (\ref{me1}), Eq. (\ref{me2}) and Eq. (\ref{me3}) become:
\begin{eqnarray}
\label{me4}
y_i= X_i^T\vc{\gvc{\beta}} + f(\vc{M}_{i}^{(1)},\vc{M}_{i}^{(2)}, \vc{M}_{i}^{(3)}) + \epsilon_i,
\end{eqnarray}
\begin{multline}
\label{me5}
f(\vc{M}^{(1)}_{i}, \vc{M}^{(2)}_{i}, \vc{M}^{(3)}_{i}) =  h_{\vc{M}^{(1)}}(\vc{M}_{i}^{(1)}) +   h_{\vc{M}^{(2)}}(\vc{M}_{i}^{(2)}) + h_{\vc{M}^{(3)}}(\vc{M}_{i}^{(3)}) +  h_{\vc{M}^{(1)}\times \vc{M}^{(2)}}(\vc{M}_{i}^{(1)}, \vc{M}_{i}^{(2)})+  \\  h_{\vc{M}^{(1)}\times \vc{M}^{(3)}}(\vc{M}_{i}^{(1)}, \vc{M}_{i}^{(3)}) +  h_{\vc{M}^{(2)}\times \vc{M}^{(3)}}(\vc{M}_{i}^{(2)}, \vc{M}_{i}^{(3)}) +    h_{\vc{M}^{(1)}\times \vc{M}^{(2)}\times \vc{M}^{(3)}}(\vc{M}_{i}^{(1)}, \vc{M}_{i}^{(2)}, \vc{M}_{i}^{(3)}),  
\end{multline}
 and
\begin{multline}
\label{me6}
 \mc{H}= \mc{H}_{\vc{M}^{(1)}}\oplus\mc{H}_{\vc{M}^{(2)}} \oplus\mc{H}_{\vc{M}^{(3)}}\oplus\mc{H}_{\vc{M}^{(1)}\times \vc{M}^{(2)}}\oplus\mc{H}_{\vc{M}^{(1)}\times \vc{M}^{(3)}} \oplus\mc{H}_{\vc{M}^{(2)}\times \vc{M}^{(3)}}\oplus \mc{H}_{\vc{M}^{(1)}\times \vc{M}^{(2})\times \vc{M}^{(3)}}, 
\end{multline}
respectively. Here $\mc{H}_{\vc{M}^{(1)}}$,  $\mc{H}_{\vc{M}^{(2)}}$ and $\mc{H}_{\vc{M}^{(3)}}$, and $\mc{H}_{\vc{M}^{(1)}\times \vc{M}^{(2)}}$, $\mc{H}_{\vc{M}^{(1)}\times \vc{M}^{(3)}}$ and $\mc{H}_{\vc{M}^{(2)}\times \vc{M}^{(3)}}$, and   $\mc{H}_{\vc{M}^{(1)}\times \vc{M}^{(2)}\times \vc{M}^{(3)}}$ are RKHSs functions on  $\mc{M}_1$, $\mc{M}_2$ and $\mc{M}_3$, and   $\mc{M}_1\times\mc{M}_2$, $\mc{M}_1\times\mc{M}_3$ and $\mc{M}_2\times\mc{M}_3$ and  $\mc{M}_1, \mc{M}_2\times\mc{M}_3$, respectively. The notation $\oplus$ is  a direct sum  of RKHS.

\subsection{Model estimation}
\label{sec:modelest}
 We can estimate the function $f\in \mc{H}$ by minimizing the  penalized squared error loss function of Eq. (\ref{me4}) as:
\begin{eqnarray}
\label{mee1}
\mc{L}(\vc{y},\gvc{\beta}, f)= \frac{1}{2}\sum_{i=1}^n\left[ y_i- X_i^T\gvc{\beta}-f(\vc{M}_{i}^{(1)},\vc{M}_{i}^{(2)}, \vc{M}_{i}^{(3)})\right]^2+\frac{\lambda}{2}\mc{J}(f)
\end{eqnarray}
where $\mc{J}(\cdot)= \|\cdot\|_\mc{H}^2$ is a roughness penalty  with tuning parameter $\lambda$. It is known that the complete function space of  Eq. (\ref{me6}), $\mc{H}$, has the orthogonal decomposition. Hence the function  $\mc{J}(\cdot)$ can be decomposed accordingly. Eq. (\ref{mee1}) then becomes:
\begin{eqnarray}
\label{mee2}
\mc{L}(\vc{y},\gvc{\beta}, f) &=& \frac{1}{2}\sum_{i=1}^n\left[y_i - X_i^T\gvc{\beta} - h_{\vc{M}^{(1)}}(\vc{M}_{i}^{(1)}) - h_{\vc{M}^{(2)}}(\vc{M}_{i}^{(2)}) - h_{\vc{M}^{(3)}}(\vc{M}_{i}^{(3)}) -  h_{\vc{M}^{(1)}\times \vc{M}^{(2)}}(\vc{M}_{i}^{(1)}, \vc{M}_{i}^{(2)}) \right. \nonumber\\ &-& \left.  h_{\vc{M}^{(1)}\times \vc{M}^{(3)}}(\vc{M}_{i}^{(1)}, \vc{M}_{i}^{(3)}) - h_{\vc{M}^{(2)}\times \vc{M}^{(3)}}(\vc{M}_{i}^{(2)}, \vc{M}_{i}^{(3)}) -    h_{\vc{M}^{(1)}\times \vc{M}^{(2)}\times \vc{M}^{(3)}}(\vc{M}_{i}^{(1)}, \vc{M}_{i}^{(2)}, \vc{M}_{i}^{(3)})\right]^2 \nonumber \\&+& 
\frac{\lambda^{(1)}}{2}\|h_{\vc{M}^{(1)}}\|^2+\frac{\lambda^{(2)}}{2}\|h_{\vc{M}^{(2)}}\|^2 + \frac{\lambda^{(3)}}{2}\|h_{\vc{M}^{(3)}}\|^2+\frac{\lambda^{(1\times 2)}}{2}\|h_{\vc{M}^{(1)}\times \vc{M}^{(2)}}\|^2+\frac{\lambda^{(1\times 3)}}{2}\|h_{\vc{M}^{(1)}\times \vc{M}^{(3)}}\|^2\nonumber \\&+& \frac{\lambda^{(2\times 3)}}{2}\|h_{\vc{M}^{(2)}\times \vc{M}^{(3)}}\|^2+  \frac{\lambda^{(1\times 2\times 3)}}{2}\|h_{\vc{M}^{(1)}\times \vc{M}^{(2)}\times \vc{M}^{(3)}}\|^2\\&=& 
\left[\vc{y} - \vc{X}\gvc{\beta} - \vc{h}_{\vc{M}^{(1)}} - \vc{h}_{\vc{M}^{(2)}} - \vc{h}_{\vc{M}^{(3)}} -  \vc{h}_{\vc{M}^{(1)}\times \vc{M}^{(2)}}-\vc{h}_{\vc{M}^{(1)}\times \vc{M}^{(3)}}  \right. \nonumber\\ &-& \left. \vc{h}_{\vc{M}^{(2)}\times \vc{M}^{(3)}}-\vc{h}_{\vc{M}^{(1)}\times \vc{M}^{(2)}\times \vc{M}^{(3)}}\right]^2 +
\frac{\lambda^{(1)}}{2}\|h_{\vc{M}^{(1)}}\|^2 + \frac{\lambda^{(2)}}{2}\|h_{\vc{M}^{(2)}}\|^2 + \frac{\lambda^{(3)}}{2}\|h_{\vc{M}^{(3)}}\|^2\nonumber \\&+& \frac{\lambda^{(1\times 2)}}{2}\|h_{\vc{M}^{(1)}\times \vc{M}^{(2)}}\|^2+\frac{\lambda^{(1\times 3)}}{2}\|h_{\vc{M}^{(1)}\times \vc{M}^{(3)}}\|^2+\frac{\lambda^{(2\times 3)}}{2}\|h_{\vc{M}^{(2)}\times \vc{M}^{(3)}}\|^2+  \frac{\lambda^{(1\times 2\times 3)}}{2}\|h_{\vc{M}^{(1)}\times \vc{M}^{(2)}\times \vc{M}^{(3)}}\|^2, \nonumber
\end{eqnarray}
where $\vc{y}=[y_1, y_2, \cdots, y_n]^T$,\,  $\vc{X}= [X_1, X_2, \cdots, X_n]^T$,\, 
 $\vc{h}_{\vc{M}^{(1)}}=[h_{\vc{M}^{(1)}}(\vc{M}_{1}^{(1)}), h_{\vc{M}^{(1)}}(\vc{M}_{2}^{(1)}), \cdots, h_{\vc{M}^{(1)}}(\vc{M}_{n}^{(1)})]^T,$ \,
$\vc{h}_{\vc{M}^{(2)}}=[h_{\vc{M}^{(2)}}(\vc{M}_{1}^{(2)}), h_{\vc{M}^{(2)}}(\vc{M}_{2}^{(2)}), \cdots, h_{\vc{M}^{(2)}}(\vc{M}_{n}^{(2)})]^T$, \,
 $\vc{h}_{\vc{M}^{(3)}}=[h_{\vc{M}^{(3)}}(\vc{M}_{1}^{(3)}), h_{\vc{M}^{(3)}}(\vc{M}_{2}^{(3)}), \cdots, h_{\vc{M}^{(3)}}(\vc{M}_{n}^{(3)})]^T$, \,
 $\vc{h}_{\vc{M}^{(1)}\times \vc{M}^{(2)}}=[h_{\vc{M}^{(1)}\times \vc{M}^{(2)}}(\vc{M}_{1}^{(1)}, \vc{M}_{1}^{(2)}), h_{\vc{M}^{(1)}\times \vc{M}^{(2)}}(\vc{M}_{2}^{(1)}, \vc{M}_{2}^{(2)}), \cdots, h_{\vc{M}^{(1)}\times \vc{M}^{(2)}}(\vc{M}_{n}^{(1)}, \vc{M}_{n}^{(2)})]^T $,\\
 $\vc{h}_{\vc{M}^{(1)}\times \vc{M}^{(3)}}=[h_{\vc{M}^{(1)}\times \vc{M}^{(3)}}(\vc{M}_{1}^{(1)}, \vc{M}_{1}^{(3)}), h_{\vc{M}^{(1)}\times \vc{M}^{(3)}}(\vc{M}_{2}^{(1)}, \vc{M}_{2}^{(3)}), \cdots, h_{\vc{M}^{(1)}\times \vc{M}^{(3)}}(\vc{M}_{n}^{(1)}, \vc{M}_{n}^{(3)})]^T $,\\
 $\vc{h}_{\vc{M}^{(2)}\times \vc{M}^{(3)}}=[h_{\vc{M}^{(2)}\times \vc{M}^{(3)}}(\vc{M}_{1}^{(2)}, \vc{M}_{1}^{(3)}), h_{\vc{M}^{(2)}\times \vc{M}^{(3)}}(\vc{M}_{2}^{(2)}, \vc{M}_{2}^{(3)}), \cdots,h_{\vc{M}^{(2)}\times \vc{M}^{(3)}}(\vc{M}_{n}^{(2)}, \vc{M}_{n}^{(3)})]^T$, \,
 $\vc{h}_{\vc{M}^{(1)}\times \vc{M}^{(2)}\times \vc{M}^{(3)}}=[h_{\vc{M}^{(1)}\times \vc{M}^{(2)}\times \vc{M}^{(3)}}(\vc{M}_{n}^{(1)}, \vc{M}_{1}^{(2)}, \vc{M}_{1}^{(3)}), h_{ \vc{M}^{(1)}\times \vc{M}^{(2)}\times \vc{M}^{(3)}}(\vc{M}_{n}^{(1)}, \vc{M}_{2}^{(2)}, \vc{M}_{2}^{(3)}), \cdots, h_{\vc{M}^{(1)}\times \vc{M}^{(2)}\times \vc{M}^{(3)}}(\vc{M}_{n}^{(1)}, \vc{M}_{n}^{(2)}, \vc{M}_{n}^{(3)})]^T$,\, $\lambda^{(1)}$, $\lambda^{(2)}$,   $\lambda^{(3)}$, $\lambda^{(1\times 2)}$, $\lambda^{(1\times 3)}$,  $\lambda^{(2\times 3)}$ and  $\lambda^{(1\times 2\times 3)}$  are the  positive tuning  parameters that  trade-off between the model fits and its complexity. 

 By the representer theorem \cite{Kimeldorf-71, Schlkof-book} and  the fact that   the  reproduction kernel of  a product of an RKHS is the product of the reproducing kernels \cite{Aron-RKHS},  the expanded  functions of $f$ in   Eq.(\ref{mee2})  for arbitrary $\tilde{\vc{M}}^{(1)}\in \mc{M}^{(1)}$, $\tilde{\vc{M}}^{(2)}\in \mc{M}^{(2)}$ and $\tilde{\vc{M}}^{(3)}\in \mc{M}^{(3)}$   can be written as:
\[h_{\vc{M}^{(1)}}= \sum_{i=1}^n \alpha^{(1)}_ik^{(1)}(\tilde{\vc{M}}^{(1)},\vc{M}_i^{(1)}),\]
 \[h_{\vc{M}^{(2)}}= \sum_{i=1}^n \alpha^{(2)}_ik^{(2)}(\tilde{\vc{M}}^{(2)},\vc{M}_i^{(2)}),\]
\[h_{\vc{M}^{(3)}}= \sum_{i=1}^n \alpha^{(3)}_ik^{(3)}(\tilde{\vc{M}}^{(3)},\vc{M}_i^{(3)}),\] 
\[h_{\vc{M}^{(1)}\times \vc{M}^{(2)}}= \sum_{i=1}^n \alpha^{(1\times 2)}_ik^{(1)}(\tilde{\vc{M}}^{(1)},\vc{M}_i^{(1)}) k^{(3)}(\tilde{\vc{M}}^{(1)},\vc{M}_i^{(2)}),\] 
\[h_{\vc{M}^{(1)}\times \vc{M}^{(3)}} = \sum_{i=1}^n \alpha^{(1\times 3)}_ik^{(1)}(\tilde{\vc{M}}^{(1)},\vc{M}_i^{(1)}) k^{(3)}(\tilde{\vc{M}}^{(3)},\vc{M}_i^{(3)},\]  
\[h_{\vc{M}^{(2)}\times \vc{M}^{(3)}}= \sum_{i=1}^n \alpha^{(2\times 3)}_ik^{(2)}(\tilde{\vc{M}}^{(2)},\vc{M}_i^{(2)}) k^{(3)}(\tilde{\vc{M}}^{(3)},\vc{M}_i^{(3)}),\]
\[h_{\vc{M}^{(1)}\times \vc{M}^{(2)}\times \vc{M}^{(3)}}= \sum_{i=1}^n \alpha^{(1\times 2\times 3)}_ik^{(1)}(\tilde{\vc{M}}^{(1)},\vc{M}_i^{(1)})k^{(2)}(\tilde{\vc{M}}^{(2)},\vc{M}_i^{(2)}) k^{(3)}(\tilde{\vc{M}}^{(3)},\vc{M}_i^{(3)}).\]

   For each  data view, we can define the kernel  matrices: $\vc{K}^{(1)} = (k^{(1)}(M_i^1, \vc{M}_j^1))_{ij}$, $\vc{K}^{(2)} = (k^{(2)}(M_i^2, \vc{M}_j^2))_{ij}$, $\vc{K}^{(3)} = (k^{(3)}(M_i^3, \vc{M}_j^3))_{ij}$, $\vc{K}^{(1\times 2)} =\vc{K}^{(1)}\odot\vc{K}^{(2)}$,  $\vc{K}^{(1\times 3)} =\vc{K}^{(1)}\odot\vc{K}^{(3)}$, $\vc{K}^{(2\times 3)} =\vc{K}^{(2)}\odot\vc{K}^{(3)}$ and $\vc{K}^{(1\times 2\times 3)} =\vc{K}^{(1)}\odot \vc{K}^{(2)}\odot\vc{K}^{(3)}$, where $\odot$ is denoted as the  element-wise product of two matrices. Now we have
\begin{eqnarray}
\label{mee4}
\vc{h}_{\vc{M}^{(1)}} &=& \vc{K}^{(1)}\gvc{\alpha}^{(1)},\,  \vc{h}_{\vc{M}^{(2)}} = \vc{K}^{(2)}\gvc{\alpha}^{(2)}, \, \vc{h}_{\vc{M}^{(3)}} = \vc{K}^{(3)}\gvc{\alpha}^{(3)}, \, \vc{h}_{\vc{M}^{(1)}\times \vc{M}^{(2)}} =  \vc{K}^{(1\times 2)}\gvc{\alpha}^{(1\times 2)},  \nonumber\\ \vc{h}_{\vc{M}^{(1)}\times \vc{M}^{(3)}}& =&  \vc{K}^{(1\times 3)}\gvc{\alpha}^{(1\times 3)},  \vc{h}_{\vc{M}^{(2)}\times \vc{M}^{(3)}}=  \vc{K}^{(2\times 3)}\gvc{\alpha}^{(2\times 3)},  \, \vc{h}_{\vc{M}^{(1)}\times \vc{M}^{(2)}\times \vc{M}^{(3)}} =  \vc{K}^{(1\times 2\times 3)}\gvc{\alpha}^{(1\times 2\times\times 3)},\qquad\qquad
\end{eqnarray} 
where $\gvc{\alpha}^{(1)}=[\alpha^{(1)}_1, \alpha^{(1)}_2, \cdots, \alpha^{(1)}_n]^T$, $\gvc{\alpha}^{(2)}=[\alpha^{(2)}_1, \alpha^{(2)}_2, \cdots, \alpha^{(2)}_n]^T$, $\gvc{\alpha}^{(3)}=[\alpha^{(3)}_1, \alpha^{(3)}_2, \cdots, \alpha^{(3)}_n]^T$, $\gvc{\alpha}^{(1\times 2)}=[\alpha^{(1\times 2)}_1, \alpha^{(1\times 2)}_2, \cdots, \alpha^{(1\times 2)}_n]^T$, $\gvc{\alpha}^{(1\times 3)}=[\alpha^{(1\times 3)}_1, \alpha^{(1\times 3)}_2, \cdots, \alpha^{(1\times 3)}_n]^T$,   $\gvc{\alpha}^{(2\times 3)}=[\alpha^{(2\times 3)}_1, \alpha^{(2\times 3)}_2, \cdots, \alpha^{(2\times 3)}_n]^T$ and 
 $\gvc{\alpha}^{(1\times 2\times 3)}=[\alpha^{(1\times 2\times 3)}_1, \alpha^{(1\times 2\times 3)}_2, \cdots, \alpha^{(1\times 2\times 3)}_n]^T$. 

   Substituting $\vc{h}_{\vc{M}^{(1)}}$,  $\vc{h}_{\vc{M}^{(2)}}$, $\vc{h}_{\vc{M}^{(3)}}$, $\vc{h}_{\vc{M}^{(1)}\times \vc{M}^{(2)}}$, $\vc{h}_{\vc{M}^{(1)}\times \vc{M}^{(3)}}$, $\vc{h}_{\vc{M}^{(2)}\times \vc{M}^{(3)}}$  and $\vc{h}_{\vc{M}^{(1)}\times \vc{M}^{(2)}\times \vc{M}^{(3)}}$ into Eq. (\ref{mee2}), and  applying the reproducing kernel properties, we get
\begin{eqnarray}
\label{mee5}
\mc{L}(\vc{y},\gvc{\beta}, \gvc{\alpha})&=&\frac{1}{2}\gvc{\epsilon}^T\gvc{\epsilon} +\frac{\lambda^{(1)}}{2}[\gvc{\alpha}^{(1)}]^T\vc{K}^{(1)} \gvc{\alpha}^{(1)}+\frac{\lambda^{(2)}}{2}[\gvc{\alpha}^{(2)}]^T\vc{K}^{(2)} \gvc{\alpha}^{(3)}+\frac{\lambda^{(3)}}{2}[\gvc{\alpha}^{(3)}]^T\vc{K}^{(3)} \gvc{\alpha}^{(3)} \nonumber\\ &+&\frac{\lambda^{(1\times 2)}}{2}[\gvc{\alpha}^{(1\times 2)}]^T\vc{K}^{(1\times 2)} \gvc{\alpha}^{(1\times 2} +\frac{\lambda^{(1\times 3)}}{2}[\gvc{\alpha}^{(1\times 3)}]^T\vc{K}^{(1\times 3)} \gvc{\alpha}^{(1\times 3)} +\frac{\lambda^{(2\times 3)}}{2}[\gvc{\alpha}^{(2\times 3)}]^T\vc{K}^{(2\times 3)} \gvc{\alpha}^{(2\times 3)}     
\nonumber\\ &+&\frac{\lambda^{(1\times 2\times 3)}}{2}[\gvc{\alpha}^{(1\times 2\times 3)}]^T\vc{K}^{(1\times 2\times 3)}\gvc{\alpha}^{(1\times 2\times 3)}  
\end{eqnarray}
where $\epsilon=\vc{y} - \vc{X}\gvc{\beta}-\vc{K}^{(1)}\gvc{\alpha}^{(1)} - \vc{K}^{(2)}\gvc{\alpha}^{(2)}- \vc{K}^{(3)}\gvc{\alpha}^{(3)} - \vc{K}^{(1\times 2)}\gvc{\alpha}^{(1\times 2)} -  \vc{K}^{(1\times 3)}\gvc{\alpha}^{(1\times 3)}- \vc{K}^{(2\times 3)}\gvc{\alpha}^{(2\times 3)} - \vc{K}^{(1\times2\times 3)}\gvc{\alpha}^{(1\times 2\times 3)}$ and  $\gvc{\alpha}=(\gvc{\alpha}^{(1)}, \gvc{\alpha}^{(2)}, \gvc{\alpha}^{(3)}, \gvc{\alpha}^{(1\times 2)}, \gvc{\alpha}^{(1\times 3)}, \gvc{\alpha}^{(2\times 3)} , \gvc{\alpha}^{(1\times 2\times 3)})$.

The gradients of  $\mc{L}$ with respect to the parametric  coefficients $\gvc{\beta}$ and nonparametric coefficients $\gvc{\alpha}'s$  are
 \begin{eqnarray}
\label{mee6}
\frac{\partial \mc{L}}{\partial \gvc{\beta}} =\vc{X}^T\gvc{\epsilon}, \, \frac{\partial \mc{L}}{\partial \gvc{\alpha}^{(1)}}= [\vc{K}^{(1)}]^T\gvc{\epsilon}+ \lambda^{(1)}\vc{K}^{(1)}\gvc{\alpha}^{(1)}, \nonumber\\
\frac{\partial \mc{L}}{\partial \gvc{\alpha^{(2)}}}= [\vc{K}^{(2)}]^T\gvc{\epsilon}+ \lambda^{(2)}\vc{K}^{(2)}\gvc{\alpha}^{(2)},\, \frac{\partial \mc{L}}{\partial \gvc{\alpha^{(3)}}}= [\vc{K}^{(3)}]^T\gvc{\epsilon}+ \lambda^{(3)}\vc{K}^{(3)}\gvc{\alpha}^{(3)},  \nonumber\\
 \frac{\partial \mc{L}}{\partial \gvc{\alpha}^{(1\times 2)}}= [\vc{K}^{(1\times 2)}]^T\gvc{\epsilon}+ \lambda^{(1\times 2)}\vc{K}^{(1\times 2)}\gvc{\alpha}^{(1\times 2)}, 
 \frac{\partial \mc{L}}{\partial \gvc{\alpha}^{(1\times 3)}}= [\vc{K}^{(1\times 3)}]^T\gvc{\epsilon}+ \lambda^{(1\times 3)}\vc{K}^{(1\times 3)}\gvc{\alpha}^{(1\times 3)},\nonumber\\
 \frac{\partial \mc{L}}{\partial \gvc{\alpha}^{(2\times 3)}}= [\vc{K}^{(2\times 3)}]^T\gvc{\epsilon}+ \lambda^{(2\times 3)}\vc{K}^{(2\times 3)}\gvc{\alpha}^{(2\times 3)},\, \frac{\partial \mc{L}}{\partial \gvc{\alpha}^{(1\times 2\times 3)}}= [\vc{K}^{(1\times2\times 3)}]^T\gvc{\epsilon}+ \lambda^{(1\times 2\times 3)}\vc{K}^{ (1\times 2\times 3)}\gvc{\alpha}^{(1\times 2\times 3)}
\end{eqnarray}
 By setting the gradients to zero, this first-order condition is given by the linear system as follows:
 %{\scriptsize 
{\tiny
\begin{eqnarray}
\label{mee7}
\begin{bmatrix}
\vc{X}^T\vc{X}&\vc{X}^T\vc{K}^{(1)}&\vc{X}^T\vc{K}^{(2)} &\vc{X}^T\vc{K}^{(3)} &\vc{X}^T\vc{K}^{(1\times 2)}&\vc{X}^T\vc{K}^{(1\times 3)}&\vc{X}^T\vc{K}^{(2\times 3)}  &\vc{X}^T\vc{K}^{(1\times2\times 3)}\\
[\vc{K}^{(1)}]^T\vc{X}&A&[\vc{K}^{(1)}]^T\vc{K}^{(2)} &[\vc{K}^{(1)}]^T\vc{K}^{(3)} &[\vc{K}^{(1)}]^T\vc{K}^{(1\times 2)}&[\vc{K}^{(1)}]^T\vc{K}^{(1\times 3)}&[\vc{K}^{(1)}]^T\vc{K}^{(2\times 3)}  &[\vc{K}^{(1)}]^T\vc{K}^{(1\times2\times 3)}\\
[\vc{K}^{(2)}]^T\vc{X}&[\vc{K}^{(2)}]^T\vc{K}^{(1)}&B &[\vc{K}^{(2)}]^T\vc{K}^{(3)} &[\vc{K}^{(2)}]^T\vc{K}^{(1\times 2)}&[\vc{K}^{(2)}]^T\vc{K}^{(1\times 3)}&[\vc{K}^{(2)}]^T\vc{K}^{(2\times 3)}  &[\vc{K}^{(2)}]^T\vc{K}^{(1\times2\times 3)}\\
[\vc{K}^{(3)}]^T\vc{X}&[\vc{K}^{(3)}]^T\vc{K}^{(1)} &[\vc{K}^{(3)}]^T\vc{K}^{(2)}&C& [\vc{K}^{(3)}]^T\vc{K}^{(1\times 2)}&[\vc{K}^{(3)}]^T\vc{K}^{(1\times 3)}&[\vc{K}^{(3)}]^T\vc{K}^{(2\times 3)}  &[\vc{K}^{(3)}]^T\vc{K}^{(1\times2\times 3)}\\
[\vc{K}^{(1\times 2)}]^T\vc{X}&[\vc{K}^{(1\times 2)}]^T\vc{K}^{(1)} &[\vc{K}^{(1\times 2)}]^T\vc{K}^{2}&[\vc{K}^{(1\times 2)}]^T\vc{K}^{3}&D&[\vc{K}^{(1\times 2)}]^T\vc{K}^{(1\times 3)}&[\vc{K}^{(1\times 2)}]^T\vc{K}^{(2\times 3)}  &[\vc{K}^{(1\times 2)}]^T\vc{K}^{(1\times2\times 3)}\\
[\vc{K}^{(1\times 3)}]^T\vc{X}&[\vc{K}^{(1\times 3)}]^T\vc{K}^{(1)} &[\vc{K}^{(1\times 3)}]^T\vc{K}^{2}&[\vc{K}^{(1\times 3)}]^T\vc{K}^{3}&[\vc{K}^{(1\times 3)}]^T\vc{K}^{(1\times 2)}&E&[\vc{K}^{(1\times 3)}]^T\vc{K}^{(2\times 3)}  &[\vc{K}^{(1\times 3)}]^T\vc{K}^{(1\times2\times 3)}\\
[\vc{K}^{(2\times 3)}]^T\vc{X}&[\vc{K}^{(2\times 3)}]^T\vc{K}^{(1)} &[\vc{K}^{(2\times 3)}]^T\vc{K}^{2}&[\vc{K}^{(2\times 3)}]^T\vc{K}^{3}&[\vc{K}^{(2\times 3)}]^T\vc{K}^{(1\times 2)}&[\vc{K}^{(2\times 3)}]^T\vc{K}^{(1\times 3)}  &F&[\vc{K}^{(2\times 3)}]^T\vc{K}^{(1\times2\times 3)}\\
[\vc{K}^{(1\times2\times 3)}]^T\vc{X}&[\vc{K}^{(1\times2\times 3)}]^T\vc{K}^{(1)} &[\vc{K}^{ 1\times 2\times 3}]^T\vc{K}^{2}&[\vc{K}^{(1\times2\times 3)}]^T\vc{K}^{3}&[\vc{K}^{(1\times2\times 3)}]^T\vc{K}^{(1\times 2)}&[\vc{K}^{ 1\times 2\times 3}]^T\vc{K}^{(1\times 3)}  &[\vc{K}^{(1\times2\times 3)}]^T\vc{K}^{ 2\times 3} &G
\end{bmatrix} \nonumber
\end{eqnarray}
\begin{eqnarray}
\begin{bmatrix}
\gvc{\beta}\\
\gvc{\alpha}^{(1)}\\
\gvc{\alpha}^{(2)}\\
\gvc{\alpha}^{(3)}\\
\gvc{\alpha}^{1\times 2}\\
\gvc{\alpha}^{1\times 3}\\
\gvc{\alpha}^{2\times 3}\\
\gvc{\alpha}^{1\times 2\times 3}
\end{bmatrix} =  \begin{bmatrix}
\vc{X}^T\vc{y}\\
\vc{K}^{(1)}\vc{y}\\
\vc{K}^{(2)}\vc{y}\\
\vc{K}^{(3)}\vc{y}\\
\vc{K}^{(1\times 2)}\vc{y}\\
\vc{K}^{(1\times 3)}\vc{y}\\
\vc{K}^{(2\times 3)}\vc{y}\\
\vc{K}^{(1\times2\times 3)}\vc{y}
\end{bmatrix},
\end{eqnarray}
}
where
$A = [\vc{K}^{(1)}]^T\vc{K}^{(1)}+ \lambda^{(1)}\vc{K}^{(1)}$, $B = [\vc{K}^{(2)}]^T\vc{K}^{(2)}+ \lambda^{(2)}\vc{K}^{(2)}$, $C= [\vc{K}^{(3)}]^T\vc{K}^{(3)}+ \lambda^{(3)}\vc{K}^{(3)}$,  $D = [\vc{K}^{(1\times 2)}]^T\vc{K}^{(1\times 2)}+ \lambda^{(1\times 2)}\vc{K}^{(1\times 2)}$, $E = [\vc{K}^{(1\times 3)}]^T\vc{K}^{(1\times 3)}+ \lambda^{(1\times 3)}\vc{K}^{(1\times 3)}$,$F = [\vc{K}^{(2\times 3)}]^T\vc{K}^{(2\times 3)}+ \lambda^{(2\times 3)}\vc{K}^{(2\times 3)}$  $G = [\vc{K}^{(1\times 2\times 3)}]^T\vc{K}^{(1\times 2\times 3)}+ \lambda^{(1\times 2\times 3)}\vc{K}^{(1\times 2\times 3)}$. Following many derivations in the literature (e.g., \cite{Liu-07, Li-12, Ge-15}), we can show that a  first-order linear system is equivalent to the normal equation of the linear mixed effects model:
\begin{eqnarray}
\label{mee8}
\vc{y}=\vc{X}\gvc{\beta}+\vc{h}_{\vc{M}^{(1)}}+\vc{h}_{\vc{M}^{(2)}}+\vc{h}_{\vc{M}^{(3)}}+ \vc{h}_{\vc{M}^{(1)}\times \vc{M}^{(2)}} + \vc{h}_{\vc{M}^{(1)}\times \vc{M}^{(3)}}+ \vc{h}_{\vc{M}^{(2)}\times \vc{M}^{(3)}}+ \vc{h}_{\vc{M}^{(1)}\times \vc{M}^{(2)}\times \vc{M}^{(3)}}+\gvc{\epsilon},
\end{eqnarray}
where $\gvc{\beta}$ is a coefficient vector of fixed effects, 
$\vc{h}_{\vc{M}^{(1)}}$, $\vc{h}_{\vc{M}^{(2)}}$, $\vc{h}_{\vc{M}^{(3)}}$,  $\vc{h}_{\vc{M}^{(1)}\times \vc{M}^{(2)}}$,  $\vc{h}_{\vc{M}^{(1)}\times \vc{M}^{(3)}}$,  $\vc{h}_{\vc{M}^{(2)}\times \vc{M}^{(3)}}$ and $\vc{h}_{\vc{M}^{(1)}\times \vc{M}^{(2)}\times \vc{M}^{(3)}}$ are independent random effects with distribution as  $\vc{h}_{\vc{M}^{(1)}}\sim N(0, \tau^{(1)}\vc{K}^{(1)}), \tau^{(1)}= \frac{\sigma^2}{\lambda^{(1)}}$, $\vc{h}_{\vc{M}^{(2)}}\sim N(0, \tau^{(2)}\vc{K}^{(2)}), \tau^{(2)}= \frac{\sigma^2}{\lambda^{(2)}}$, $\vc{h}_{\vc{M}^{(3)}}\sim N(0, \tau^{(3)}\vc{K}^{(3)}), \tau^{(3)}= \frac{\sigma^2}{\lambda^{(3)}}$, $\vc{h}_{M^{(1\times 2)}}\sim N(0, \tau^{(1\times 2)}\vc{K}^{(1\times 2)}), \tau^{(1\times 2)}= \frac{\sigma^2}{\lambda^{(1\times 2)}}$,  $\vc{h}_{M^{(1\times 3)}}\sim N(0, \tau^{(1\times 3)}\vc{K}^{(1\times 3)}), \tau^{(1\times 3)} = \frac{\sigma^2}{\lambda^{(1\times 3)}}$,   $\vc{h}_{M^{(2\times 3)}}\sim N(0, \tau^{(2\times 3)}\vc{K}^{(2\times 3)}), \tau^{(2\times 3)} = \frac{\sigma^2}{\lambda^{(2\times 3)}}$,   $\vc{h}_{M^{(1\times 2\times 3)}}\sim N(0, \tau^{(1\times 2\times 3)}\vc{K}^{(1\times 2\times 3)}), \tau^{(1\times 2\times 3)} = \frac{\sigma^2}{\lambda^{(1\times 2\times 3)}}$. $\gvc{\epsilon}$ is  also an independent random variable with the  distribution $\gvc{\epsilon}\sim N(0, \sigma^2 \vc{I})$, where $\vc{I}$ is an identity matrix. This relationship insures that all of the  effects extracted by minimizing the loss function in Eq. (\ref{mee1}), are the same as  the best linear unbiased predictors (BLUPs) of the linear mixed effects model in Eq. (\ref{mee8}). It is possible to estimate the variance components using the restricted maximum likelihood (ReML) approach (see in the appendix for details).  The solution of the  linear system  in Eq. (\ref{mee7}) gives the coefficients of the  fixed effect, $\gvc{\beta}$, and coefficients  for the  random effect,  $\gvc{\alpha}$. By  inserting $\gvc{\alpha}$ into Eq. (\ref{mee4}),  we can estimate the random effects $\hat{\vc{h}}_{\vc{M}^{(1)}}$, $\hat{\vc{h}}_{\vc{M}^{(2)}}$, $\hat{\vc{h}}_{\vc{M}^{(3)}}$, $\hat{\vc{h}}_{M^{1\times 2}}$, $\hat{\vc{h}}_{M^{(1\times 3)}}$, $\hat{\vc{h}}_{M^{(2\times 3)}}$ and  $\hat{\vc{h}}_{M^{(1\times 2\times 3)}}$, respectively.

\section{Statistical testing}
\label{sec:test}
Using positive definite kernels,  we treat each gene-derived SNP, ROI, and gene-derived DNA methylation as a testing unit.  In the  following subsections, we study the test statistic of the  overall effect and higher order interaction effects. 
\subsection{Testing overall effect}
We known that the overall testing  effect $H_0:  h_{\vc{M}^{(1)}} (\cdot) = h_{\vc{M}^{(2)}} (\cdot) = h_{\vc{M}^{(3)}} (\cdot) =  h_{\vc{M}^{(1)}\times \vc{M}^{(2)}} (\cdot) =  h_{\vc{M}^{(1)}\times \vc{M}^{(3)}} (\cdot) =  h_{\vc{M}^{(2)}\times \vc{M}^{(3)}} (\cdot) =  h_{\vc{M}^{(1)}\times \vc{M}^{(2)}\times \vc{M}^{(3)}} (\cdot)=0$
is equivalent to test the variance components in Eq.(\ref{mee8}),    $H_0:  \tau^{(1)} =\tau^{(2)} = \tau^{(3)} = \tau^{(1\times 2)} = \tau^{(1\times 3)}=\tau^{(2\times 3)} =  \tau^{(1\times 2\times 3)}=0$. 

 Unfortunately, under the null hypothesis, the asymptotic distribution of a likelihood ratio test (LRT) statistic does not follow a chi-square distribution or a mixture chi-square distribution.  Because   the parameters in the  variance components analysis are laid on the boundary of the parameter space when the null hypothesis is true and kernel matrices are not block-diagonal, S. Li and Cui (2012) have  proposed a score test statistic based on the restricted likelihood. In this paper, we have constructed a score test statistic for the multi-view data model, Eq. (\ref{mee8}). Assuming that the linear mixed model in Eq. (\ref{mee8})  has multivariate normal distribution with mean $\vc{X}\gvc{\beta}$ and variance-covariance matrix  $\Theta (\gvc{\theta})= \sigma^2\vc{I} + \tau^{(1)}\vc{K}^{(1)}+ \tau^{(2)}\vc{K}^{(2)}+\tau^{(3)}\vc{K}^{(3)}+ \tau^{(1\times 2)}\vc{K}^{(1\times 2)}+ \tau^{(1\times 3)}\vc{K}^{(1\times 3)} + \tau^{(2\times 3)}\vc{K}^{(2\times 3)} + \tau^{(1\times 2\times 3)}\vc{K}^{(1\times 2\times 3)}$, where $\gvc{\theta} = (\sigma^2, \tau^{(1)}, \tau^{(2)},  \tau^{(3)}, \tau^{(1\times 2)}, \tau^{(2\times 3)}, \tau^{(1\times 2\times 3)})$ are the  variance components. The restricted log-likelihood function of Eq. (\ref{mee8}) can be written as
\begin{eqnarray}
\label{mee9}
\ell_R (\gvc{\theta}) = -\frac{1}{2}ln (|\Theta (\gvc{\theta})|)- \frac{1}{2} ln (|\vc{X}^T \vc{\Theta}^{-1} (\gvc{\theta}) \vc{X} |) - \frac{1}{2} (\vc{y} - \vc{X}\gvc{\hat{\beta}})^T\gvc{\Theta}^{-1} (\gvc{\theta}) (\vc{y} - \vc{X}\gvc{\beta})
\end{eqnarray}
The estimate of the variance components are obtained by the partial derivative  of Eq. (\ref{mee9}) with respect to each of the variance components (see appendix for more detail).  By considering that  the true value of $\sigma^2$ under the null hypothesis is $\sigma^2_0$, under the ReML the score test statistic is defined as
\begin{eqnarray}
\label{overall}
S(\sigma^2_0)= \frac{1}{2\sigma^2_0} (\vc{y}- \vc{X}\hat{\gvc{\beta}})^T \vc{K}(\vc{y}- \vc{X}\hat{\gvc{\beta}})
\end{eqnarray}
where $\vc{K}= \vc{K}^{(1)}+ \vc{K}^{(2)}+\vc{K}^{(3)}+ \vc{K}^{(1\times 2)} +\vc{K}^{(1\times 3)} +\vc{K}^{(2\times 3)} + \vc{K}^{(1\times 2\times 3)}$,  $\hat{\gvc{\beta}}$ is the maximum likelihood estimator (MLE) of the regression coefficient under the null hypothesis $\vc{y}= \vc{X}\gvc{\beta}+\epsilon_0$, $\sigma_0^2$ is the variance of $\epsilon_0$, and $S(\sigma_0^2)$ is the quadratic function for  the variable $\vc{y}$, which follows a mixture of the chi-square distribution under the null hypothesis.   By the Satterthwaite method \cite{Satterthwaite-46}, we can approximate the distribution of $S(\sigma_0^2)$ to a scaled chi-square distribution, i.e., $S(\sigma_0^2)\sim \gamma \chi^2_\nu$, where the scale parameter $\gamma$ and the degrees of freedom $\nu$ can be measured by the method of moments (MOM).  The mean and variance of the test statistic  $S(\sigma_0^2)$ are
\[ \rm{E}[S(\sigma_0^2)]=\rm{E}[\gamma \chi^2_\nu]= \gamma\nu,\qquad \rm{and} \qquad
  \rm{Var}[S(\sigma_0^2)= \rm{Var}[ [\gamma \chi^2_\nu]= 2\gamma^2\nu, \]
respectively. By solving the above two equations, we have $\hat{\gamma}=\frac{\rm{Var}[S(\sigma_0^2)]}{2\rm{E}[S(\sigma_0^2)]}$ and $\hat{\nu}=\frac{2\rm{E}[S(\sigma_0^2)^2]}{\rm{Var}[S(\sigma_0^2)]} $. In practices, $\sigma_0^2$ is unknown but we can replace it by its ReML under the null model denoted by $\hat{\sigma}_0^2$. Lastly, the $p-$value of an experimental  score statistic  $S(\hat{\sigma}_0^2)$ is obtained   using the  scaled chi-square distribution $\hat{\gamma}\chi^2_{\hat{\nu}}$.  

\subsection{Testing higher order interaction effect}
To test the higher order interaction effect, we show that  testing the null hypothesis  $H_0: h_{\vc{M}^{(1)}\times \vc{M}^{(2)}\times \vc{M}^{(3)}} (\cdot)=0$ is equivalent to testing the variance component:   $H_0: \tau^{(1\times 2\times 3)}=0$.  Let $\Sigma= \sigma^2\vc{I} + \tau^{(1)}\vc{K}^{(1)}+ \tau^{(2)}\vc{K}^{(2)}+\tau^{(3)}\vc{K}^{(3)}+ \tau^{1\times 2}\vc{K}^{(1\times 2)}+ \tau^{1\times 3}\vc{K}^{(1\times 3)} + \tau^{2\times 3}\vc{K}^{(2\times 3)}$, and $\tau^{(1)}, \tau^{(2)}, \tau^{(3)}, \tau^{(1\times 2)},  \tau^{(1\times 3)}, \tau^{(2\times 3)}$,  and   $\sigma^2$  are model parameters under the null model $ \vc{y}=\vc{X}\gvc{\beta}+\vc{h}_{\vc{M}^{(1)}}+\vc{h}_{\vc{M}^{(2)}}+\vc{h}_{\vc{M}^{(3)}}+ \vc{h}_{\vc{M}^{(1)}\times \vc{M}^{(2)}} + \vc{h}_{\vc{M}^{(1)}\times \vc{M}^{(3)}}+\gvc{\epsilon}$. We formulate a test statistic:
\begin{eqnarray}
\label{hin}
S_{\rm{I}}(\gvc{\tau}_{\rm{I}})= \frac{1}{2\sigma^2_0} \vc{y}^T \vc{W}_{01} \vc{K}^{(1\times 2\times 3)}\vc{W}_{01}\vc{y},\end{eqnarray}
 where $\gvc{\tau}_{\rm{I}} = (\sigma^2, \tau^{(1)}, \tau^{(2)},  \tau^{(3)}, \tau^{(1\times 2)},\tau^{(1\times 3)}, \tau^{(2\times 3)})$, and  $\vc{W}_{01}= \Sigma^{-1}- \Sigma^{-1}   \vc{X}(\vc{X}^T  \Sigma^{-1}  \vc{X})^{-1} \vc{X}^T\Sigma^{-1}$ is the projection matrix under the null hypothesis. Similarly to the  overall effect test,  we can use the  Satterthwaite method to  approximate the distribution  for the    higher order intersection test statistic $S_{\rm{I}}(\gvc{\tau}_{\rm{I}})$ by a scaled chi-square distribution  with scaled   $\gamma_{\rm{I}}$ and degree of freedom $\nu_{\rm{I}}$, i.e., $S_{\rm{I}}(\gvc{\tau}_{\rm{I}})\sim \gamma_{\rm{I}} \chi^2_{\nu_{\rm{I}}}$. The scaled parameter and degree of freedom  are estimated by the MOM, $\hat{\gamma}_{\rm{I}}=\frac{\rm{Var}[S_{\rm{I}}(\gvc{\tau}_{\rm{I}})]}{2\rm{E}[S_{\rm{I}}(\gvc{\tau}_{\rm{I}})]}$ and $\hat{\nu}_{\rm{I}}=\frac{2\rm{E}[S_{\rm{I}}(\gvc{\tau}_{\rm{I}})]}{\rm{Var}[S_{\rm{I}}(\gvc{\tau}_{\rm{I}})]}$, respectively.  In practice, the unknown  model parameters  $\tau^{(1)}, \tau^{(2)}, \tau^{(3)}, \tau^{(1\times 2)},  \tau^{(1\times 3)}, \tau^{(2\times 3)}$,  and   $\sigma^2$ are estimated by their respective  ReML estimates  $\hat{\tau}^{(1)}, \hat{\tau}^{(2)}, \hat{\tau}^{(3)}, \hat{\tau}^{(1\times 2)},  \hat{\tau}^{(1\times 3)}, \hat{\tau}^{(2\times 3)}$,  and   $\hat{\sigma}^2$ under the null hypothesis. Lastly, the $p-$value for the  observed higher order interaction effect (score statistic  $S_{\rm{I}}(\gvc{\tau}_{\rm{I}})$) is obtained   using the  scaled chi-square distribution $\hat{\gamma}_{\rm{I}}\chi^2_{\hat{\nu_{\rm{I}}}}$.

\subsection{Kernel choice}
In kernel methods, choosing a suitable kernel is indispensable. Most kernel methods suffer from poor selection of a suitable kernel. It is often the case that the kernel has parameters which may strongly influence the results.   Assuming $k:\mc{X}\times \mc{X}\to \mb{R}$ is a  positive definite kernel. Then for any $X, \tilde{X} \in \mc{X}$,   a linear  positive definite kernels on $\mb{R}$ is  defined as
\[
k(X, \tilde{X})=\langle X, \tilde{X}\rangle= X^T\tilde{X}.\]
The linear kernel is  used  by  the underlying Euclidean space to define the similarity measure. Whenever the dimensionality of $\vc{X}$ is very high, this may allow for more complexity in the function class than what we could measure and assess otherwise.  The polynomial kernel is defined as 
\[
k(X,\tilde{X})= (X^T\tilde{X}+c )^d, \, (c\geq 0, d\in \mb{N}). \nonumber
\]
Using the polynomial kernel makes it  possible to use higher order correlations between data for different purposes. This kernel incorporates every polynomial interaction up to degree $d$ (provided that $c >0$). For instance, if we want to take only the  mean and variance into account, we  only need to consider $d=2$ and $c=1$. For more emphasis on mean we need to increase the constant offset $a$. Polynomial kernels only map data into a finite dimensional space. Due to the finite bounded degree the given kernel will not provide us with guarantees for a good dependency measure.  In addition, both linear and polynomial kernels are unbounded. 

Many radial basis function kernels, such as the Gaussian  kernel,  map $X$ into a infinite dimensional space.  The  Gaussian  kernel is  defined as:  
\[
k(X,\tilde{X} )=e^{\frac{1}{2\sigma^2}-||X-\tilde{X} ||^2},\, (\sigma > 0).
\]
While the  Gaussian  kernel has a free parameter (bandwidth), it still  follows a number of theoretical   properties such as  boundedness,  consistence,  universality, robustness etc. It is the most applicable kernel of the  kernel methods \cite{Sriperumbudur-09}.  For the Gaussian kernel,  we can  use the median of the  pairwise distance as a  bandwidth \cite{ Gretton-08, Song-12}.

  For GWASs,  a kernel captures the pairwise  similarity across a number of SNPs in each gene. Kernel projects the genotype data from original space (high dimension and nonlinear) to a feature space (linear space). One of the more popular kernels used  for  genomics similarity is  the identity-by-state (IBS) kernel (nonparametric function of the genotypes) \cite{Kwee-08}:  
\[ k(\vc{M}_i, \vc{M}_j)= 1- \frac{1}{2s}\sum_{b=1}^s| M_{ib}- M_{jb} |.\]
where $s$ is the number of SNP markers of the corresponding gene. The IBS kernel does not need any assumption on these types of genetic interactions. Thus, in principle, it can capture any  effect between genetic features and their  influences on the phenotype.  In this paper, we used the  Gaussian kernel for the  quantitative  data view (imaging and epigenetics) and the  IBS kernel for the qualitative data view (genetics).

\section{Relevant methods}
\label{sec:comt}
 Li and Cui (2012)  have proposed a linear  PCA (LPCA) based regression  method  for the interaction effect between two genes. This makes it possible to extend the notion  to  three datasets.  
Let $\vc{M}^{(1)}=[M_{1}^1, M_{2}^1, \cdots,  M_{s}^1]$, $\vc{M}^{(2)}=[M_{1}^2, M_{2}^2, \cdots,  M_{r}^2]$, and  $\vc{M}^{(3)}=[M_{1}^3, M_{2}^3, \cdots,  M_{d}^3]$  be the data matrix  for the genetics, imaging and epigenetics, respectively. Using the PCA  we   can  compute the first $\ell$ principle  components:  $U_{1}^1, U_{2}^1,\cdots U_{s_\ell}^1$, $U_{1}^2, U_{2}^2,\cdots U_{r_\ell}^2$, and $U_{1}^3, U_{2}^3, \cdots U_{d_\ell}^3$ with $s_\ell \leq s$,  $r_\ell \leq r$, and $d_\ell \leq d$, for the  corresponding data matrix, respectively. We then compared the numerical, simulation and real data analysis with the following  methods: test based on  only first and  first few  principal components multiple regression,  which we are called partial principal component regression (pPCAR)  and   full  principal component regression (fPCAR)), respectively. 

\subsection{Principal component multiple regression}
By considering only the first principal component, the $3$rd order interaction model ( i.e., pPCA) can be stated  as:
\begin{eqnarray}
\label{ppca}
\vc{y}= \vc{X}\beta +\sum_{a=1}^s\alpha_a \vc{M}^{(1)}_a+\sum_{b=1}^r\alpha_b \vc{M}^{(2)}_b+\sum_{c=1}^d\alpha_c \vc{M}^{(3)}_c + \eta U_1^{(1)}U_1^{(2)}U_1^{(3)}.
\end{eqnarray}
This model is called partial  PCA regression (pPCAR).  Using the multiple   regression  in  Eq. (\ref{ppca}), the interaction of  $\vc{M}^{(1)}\times \vc{M}^{(2)}\times \vc{M}^{(3)}$ is  assessed by testing  $H_0: \eta=0.$
To consider all possible interactions of the selected principal components,  we can also replace the main effects by the first $\ell$ principal components. The number of principal components $\ell$ is selected based on the proportion of variation  explained by the principal components,  which can explain the  major variations (say, $\geq 85\%$).  The models in Eq. (\ref{ppca})  then becomes
\begin{eqnarray}
\label{fpca}
\vc{y}= \vc{X}\beta +\sum_{a=1}^{s_\ell}\alpha_a U^{(1)}_a+\sum_{b=1}^{r_\ell}\alpha_b U^{(2)}_b+\sum_{c=1}^{d_\ell}\alpha_c U^{(3)}_c + \sum_{a=1}^{s_\ell}  \sum_{b=1}^{r_\ell} \sum_{c=1}^{d_\ell}  \eta_{123} U_a^{(1)}U_b^{(2)}U_c^{(3)}
\end{eqnarray}
 Using the multiple   regression  in  Eq. (\ref{fpca}), the interaction of  $\vc{M}^{(1)}\times \vc{M}^{(2)}\times \vc{M}^{(3)}$ is  assessed by testing  $H_0: \eta_{123}=0.$ 

\subsection{Principal component sequence kernel association test}
 Over the past several years, the  sequence kernel association test (SKAT) approach  has been  widely used  in GWASs  due to its flexibility and computational efficiency. The SKAT is based on  a SNP-set (e.g., a gene or a region) level test for the association between a set of  variants and dichotomous or quantitative phenotypes.  This method aggregates individual test statistics of SNPs   and efficiently computes SNP-set level p-values, while adjusting for covariates, such as principal components to account for population stratification \cite{Wu-11, Ionita-13}.  We applied SKAT to gene-derived SNPs, ROIs, and gene-derived DNA methylations data.  To do this,  we use SKAT  in  Eq. (\ref{fpca}) and  the interaction of  $\vc{M}^{(1)}\times \vc{M}^{(2)}\times \vc{M}^{(3)}$ is  assessed by testing  $H_0: \eta_{123}=0.$

\section{Experiments}
\label{sec:exp}
We conducted experiments on both the simulation studies (numerical data and  real MCIC data) and  imaging genetics with the SZ study. We considered  the IBS kernel for the genetic data and the  Gaussian kernel for all  other data. For the Gaussian kernel,  we used the median of the  pairwise distance as the   bandwidth.  The proposed method is based on the ReML  algorithm (Fisher's scoring algorithm).  The ReML algorithm converged in less than $50$  iterations (the difference between successive log ReML values was smaller than $10^{-04}$), and in  most of the cases it converged very quickly with $10$ iterations, taking only a few seconds with  an R-program.  Solving  the ReML may be trapped by local minima. To avoid this problem, we use a set of  initial points ($0$, $0.00001$, $0.0001$, $0.001$, $0.01$, $0.1$, $1$) for the optimization algorithm and chose the best one (maximized ReML).  
\subsection{Simulation studies}
The goal of these simulation studies is  to evaluate the performance of the proposed method  and the accuracy of the score tests.   To synthesize quantitative phenotypes, we applied the following model:
\begin{multline}
\label{exper1}
 y_i=\vc{X}_i^T \beta + \alpha_1\left[h_S(S_i) + h_T(T_i) + h_C(C_i)\right] + \alpha_2\left[h_{S\times T}(S_i, T_i) + h_{S\times C}(S_i, C_i) + h_{T\times C}(T_i, C_i)\right]\\ + \alpha_3\left[h_{S\times T\times C}(S_i,T_i C_i)\right] +\sigma \epsilon_i 
\end{multline}

where $\vc{X}_i$ is a vector of covariates including an intercept (e.g., age, height,  etc.,) of  $i$th  subject ($i=1,2, \cdots, n $) and   $\beta$'s are  the coefficient. $S_i$, $T_i$,  and $C_i$   are the three data sets  and  $\epsilon_i$ is a random error that  follows the Gaussian distribution with mean  zero and unit variance,  i.e.,   $\epsilon_i\sim N(0, 1)$, and  $\sigma$ is the  standard deviation of the error and was fixed  to $10^{-02}$,  of the $i-$th  subject. For  each  function,  we designed  the following form\\
$h_S(S_i)= \sum_{a=1}^{10} S[i,a]\rm{cos}(S[i,a]),\quad h_T(T_i)= \sum_{b=1}^{2}2T[i,b]\rm{sin}(T[i,b]),\quad  h_C(C_i)= \sum_{c=1}^{10}\frac{i}{\sqrt{2}\exp(C[i,c])},\\ h_{S\times T}(S_i, T_i)= h_S(S_i) 2h_T(T_i),\qquad  h_{S\times C}(S_i, C_i)= h_S(S_i) 3h_C(C_i), \quad h_{T\times C}(S_i, T_i)= 2h_T(T_i) 3h_C(C_i),\\ h_{S\times T\times C}(S_i, T_i, C_i)= h_S(S_i) 2h_T(T_i)3h_C(C_i).$

In  simulation-I and simulation-II,  we generated data under  different values of  $(\alpha_1, \alpha_2, \alpha_3)$ to evaluate the performance  of the test. In other words, for $\alpha_1= \alpha_2= \alpha_3=0$ both main effects and all interaction effects vanish and we  examined the false positive rate of the score test of  the  over all effect.  For  $\alpha_1 \geq 0$,  $\alpha_2= 0$ ( $\alpha_2 \geq 0)$ and   $\alpha_3=0$, there are main effects (2nd order interaction effects) but no higher order interaction effects, hence we can evaluate the power of the score test.   We  also set   $(\alpha_1, \alpha_2, \alpha_3)$  to  many different values  to test the power of both score tests.   In each  setting  $500$ simulations were performed to confirm the variation of the results.

\subsubsection{Simulation-I (numerical data)}
In this simulation,  we  generated two covariates (height and weight) and  three views  (genetics, topological, and categorical data). We generated the  height and weight  by the regular sequencing of the  interval $(50, 80)$ and  $(60, 225)$ with  increment of   $2.05$ and    $4.7$ for the  $n=500$ subject, respectively. Then, we added the  noise $3 N(0, 1)$ to  each of the  variables. The  element of coefficient vector $\beta$ is fixed to  $0.5$.  For the genetics data,  we simulated  a gene with  $10$ SNPs using the latent model for $500$ subjects  as  in \cite{Parkhomenko-09, Alam-16b}. We generated data along three circles of different radii with small noise for topological features \cite{Ashad-14}:
\begin{equation}
T_i = r_i \begin{pmatrix} \cos(\vc{R}_i) \\ \sin(\vc{R}_i) \end{pmatrix}
+ \epsilon_i, \nonumber
\end{equation}
where $r_i=1$, $0.5$ and $0.25$, for $i=1,\ldots,n_1$, $i=n_1+1,\ldots,n_2$, and $i=n_2+1,\ldots, n_3$ $(n = n_1+n_2+n_3=500)$, respectively, $\vc{R}_{i}\sim U[-1, 1]$ and $\epsilon_i\sim \mc{N} (0, \,I_2)$ independently.
For the categorical data,  we  considered $10$ categories with probability $1/10$ and converted these features into the  dummy  features with levels zero and one.

In addition, to draw the receiver operating characteristic (ROC) the data was generated   by fixing $\alpha_1=1$,  $\alpha_2=1$ and  $\alpha_3$ was  allocated with probability $0.5$ for each run, whether  a random number  was  uniformly distributed on $[0, 1]$ or at  $0$ . We also only fixed $\alpha_1=1$ and  for each run $\alpha_3$ ($\alpha_2$= $\alpha_3$) was  allocated with probability $0.5$, whether  a random number is    uniformly distributed on $[0,1]$ or at $0$.  We considered three sample sizes $n\in \{100, 500, 1000\}$ and compared the ROC curves of the proposed method with the   three state-of-the-art methods  in identifying the interaction effects. 

\subsection{Simulation-II (Mind Clinical Imaging Consortium's schizophrenia data)}
\label{subsec:MCIC}
 To validate  Eq. (\ref{exper1}) under different values of $(\alpha_1, \alpha_2, \alpha_3)$, we    consider  real  data. This simulation was based on the SZ data which was collected by  the MCIC \cite{Chen-12MCIC, Liu-13, Chekouo-16}. These are  $208$ subjects including $92$ schizophrenic patients (age: $34\pm 11$, $22$ females) and $116$ (age: $32\pm 11$, $44$ females) healthy controls. All participants' symptoms were evaluated by the scale for the assessment of positive symptoms and  negative symptoms \cite{Andreasen-84}.  By filtering missing data, the number of subjects was  reduced to $182$ subjects ($79$ SZ patients and $103$ healthy controls). We considered  the age, height, and  weight as the covariates  and  gene-derived SNP, ROIs with voxels, and gene-derived DNA methylation information  as the three views. 

\par {\bf Genetics}: For each subject (SZ patients  and healthy controls) a blood sample was taken and DNA was extracted. Gene typing was performed for all subjects at the Mind Research Network using the Illumina Infinium HumanOmni1- Quad assay covering  $1140419$ SNP loci.  To form the final genotype calls and to perform a series of standard quality control procedures the bead studio and PLINK software packages were applied, respectively. The final dataset spans  $722177$ loci  with $22442$ genes of $182$ subjects. Genotypes   ``aa"  (non-minor allele), ``Aa" (one minor allele) and ``AA" (two minor alleles) were coded as $0$, $1$ and $2$ for each  SNP, respectively \cite{Alam-16b}.   A list of the top  $75$ genes  for the SZ are listed in the SZ genes database $\rm{(https://bioinfo.uth.edu/SZGR/)}$.

\par {\bf Imaging}: Participants' fMRI data were collected during a block design motor response for auditory stimulation. State-of-the-art approaches using participant feedback and expert observation  were used. The aim was to continuously monitor the patients while acquiring images with the parameters (TR=2000 ms, TE= 30ms, field of view=22cam, slice thickness=4mm, 1 mm skip, 27 slices, acquisition matrix $64\times 64$, flip angle=$90^{\circ}$) on a  Siemens 3T Trio Scanner and 1.5 T Sonata. The data comes from four different sites\ (\& scanners) with echo-planar imaging (EPI). Data were pre-processed with SPM software  and  were realigned spatially, normalized and resliced to $3\times 3 \times 3$ mm.  They were  smoothed with  a  $10\times 10 \times 10$ $\rm{mm}^3$ Gaussian kernel and then analyzed by multiple  regression that  considered the stimulus and their temporal derivatives plus an intercept term as a regressors.  Finally the stimulus-on versus stimulus-off contrast images were extracted. Next, $41236$ voxels were extracted from $116$ ROIs based on the AAL brain atlas for analysis \cite{Alam-16a}.  For  imaging  features (ROIs), we considered  $116$ ROIs. The name for the  ROIs is given by the automated anatomical labeling (AAL) template \cite{Yan-10}.

\par{\bf Epigenetics}: DNA methylation is one of the main epigenetic mechanisms to regulate gene expression, and may be involved in  the development of SZ. For this paper, we investigated $27481$ DNA methylation markers in blood from  SZ patients and healthy controls.  DNA from blood samples were measured by the Illumina Infinium Methylation27 Assay. The methylation value is calculated by taking the ratio of the methylated probe intensity and the total probe intensity.

  In this paper, the  top  $72$ genes (from $\rm{https://bioinfo.uth.edu/SZGR/}$ and genes have more than one SNP), $116$ ROIs, and form DNA methylation  $129$ genes (genes have more than 5 methylations)  are considered as gene-derived SNPs, ROIs with voxel, and  gene-derived DNA methylations  features, respectively.

\subsection{Simulation results}
Table~\ref{tab:simu} presents the simulation results (simulation-I and simulation-II) for the overall and higher order interaction  tests. The nominal $p-$value threshold was fixed to $0.05$.  By observing this table, we can see that when $\alpha_1= \alpha_2 = \alpha_3=0$, the size of the overall score test is close to  the nominal $p-$value threshold. When $\alpha_1 \geq 0$,  $\alpha_2 = 0$ (or ($\alpha_2 \geq 0$))  and $\alpha_3=0$, the false positive rate of the test for higher order interaction  effects  is also controlled. For the power analysis ($\alpha_3\geq 0$) we found that the power of the interaction test for the proposed method quickly exceeds $0.85$ and $0.90$ for simulation-I and simulation-II, respectively.  While the SKAT method has  higher power  when  compared to other relevant methods (pPCAR and fPCA) it has  lower power when compared to   the proposed method both in   simulation-I  and in simulation-II.  We observed that dimension  reduction methods (pPCAR and fPCA) can significantly   inflate the false positive rates and dramatically loses  power when compared to the proposed one  and SKAT methods. 

\begin{figure}
\begin{center}
\includegraphics[width=14cm, height=8cm]{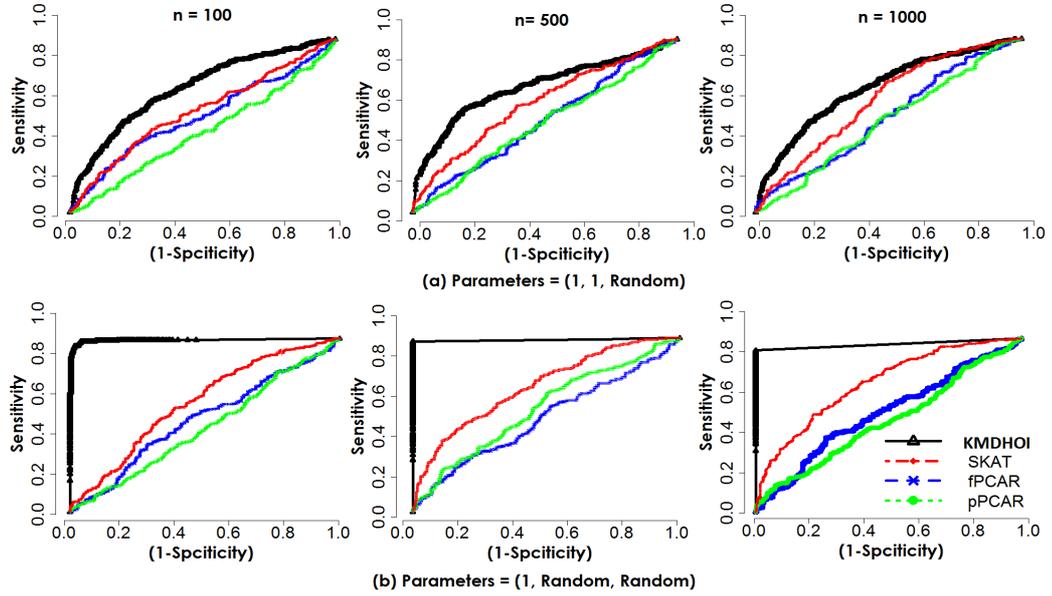}
\caption{Receiver operating characteristics (ROC)  of the kernel methods and relevant methods for higher order interaction detection with three sample sizes, $n\in \{100, 500, 1000\}$ for (a) third parameter value is random, (b) second and third parameter values are random. The sensitivity  are plotted against (1- specificity)  with the $p$-values threshold  in the range  zero to one $(0 - 1)$ with a step size $0.0001$.}
\label{fig:ROC}
\end{center}
\end{figure}

Figure~\ref{fig:ROC} shows the  receiver operating characteristics (ROC) of the proposed method and three alternative methods to detect interactions using the simulation-III with three sample sizes, $n\in \{100, 500, 1000\}$  for (a) third parameter value is random only, (b) second and third parameter values are random. The  sensitivity  are plotted against (1- specificity)  with the $p$-values threshold  in the range   $0 - 1$ with a step size $0.0001$. The power gain of the proposed method  relative to the alternative methods is evident in all situations. When the sample size was increased, and the second order interaction was equal to  one, a higher power was observed.   We also observed extremely high power for the similar  second and higher order interactions. 

\begin{table}[!tbp]
\begin{center}
\caption {Power of the overall and higher order interaction score tests, and the alternative methods for interaction detection based on  dimension reduction regression (pPCAR, fPCAR) and sequence kernel association test (SKAT). The nominal $p-$values threshold was fixed to $0.05$.}
\begin{tabular}{l|ccccc|cccccc}
\hline\hline
\rm{Parameters}&\multicolumn{5}{c|}{\rm{Simulation - I}}&\multicolumn{5}{c}{\rm{Simulation-II}}\tabularnewline
&\multicolumn{2}{c}{\rm{KMDHOI}}&\multicolumn{3}{c|}{\rm{State-of-the-art methods}} & \multicolumn{2}{c}{\rm{KMDHOI}}&\multicolumn{3}{c}{\rm{State-of-the-art methods}} \tabularnewline
&\multicolumn{2}{c}{}&\multicolumn{1}{c}{\rm{pPCAR}}&\multicolumn{1}{c}{\rm{fPCAR}} &\multicolumn{1}{c|}{\rm{SKAT}} &\multicolumn{2}{c}{}&\multicolumn{1}{c}{\rm{pPCAR}}&\multicolumn{1}{c}{\rm{fPCAR}} &\multicolumn{1}{c}{\rm{SKAT}}\tabularnewline
\multicolumn{1}{l}{($\alpha_1$, $\alpha_2$, $\alpha_3$)}&\multicolumn{1}{|c}{Overall}&\multicolumn{1}{c}{HOI}&\multicolumn{1}{c}{HOI}&\multicolumn{1}{c}{HOI}&\multicolumn{1}{c|}{HOI}&\multicolumn{1}{c}{Overall}&\multicolumn{1}{c}{HOI}&\multicolumn{1}{c}{HOI}&\multicolumn{1}{c}{HOI}&\multicolumn{1}{c}{HOI}\tabularnewline
\hline
(0, 0, 0)&$0.047$&$0.003$&$0.044$&$0.036$&$0.052$&$0.055$&$0.009$&$0.046$&$0.056$&$0.045$\tabularnewline
(0.5, 0, 0)&$1.00$&$0.000$&$0.137$&$0.041$&$0.127$&$0.999$&$0.339$&$0.172$&$0.077$&$0.188$\tabularnewline
(1, 0, 0)&$1.00$&$0.000$&$0.060$&$0.015$&$0.143$&$0.999$&$0.457$&$0.199$&$0.077$&$0.176$\tabularnewline
(0, 0.5, 0)&$1.00$&$0.336$&$0.137$&$0.053$&$0.160$&$1.000$&$0.882$&$0.189$&$0.095$&$0.261$\tabularnewline
(0, 0.5, 0.5)&$1.00$&$0.745$&$0.161$&$0.093$&$0.410$&$1.000$&$0.899$&$0.199$&$0.107$&$0.288$\tabularnewline
(0, 0.5, 1)&$1.00$&$0.843$&$0.157$&$0.082$&$0.478$&$1.000$&$0.904$&$0.173$&$0.092$&$0.305$\tabularnewline
(0, 0,0.1)&$1.00$&$0.813$&$0.173$&$0.080$&$0.787$&$1.000$&$0.924$&$0.231$&$0.132$&$0.362$\tabularnewline
(0, 0,1)&$1.00$&$0.882$&$0.213$&$0.110$&$0.748$&$1.000$&$0.918$&$0.283$&$0.115$&$0.351$\tabularnewline
(0.5, 0.5, 0.5)&$1.00$&$0.765$&$0.174$&$0.107$&$0.403$&$1.000$&$0.872$&$0.151$&$0.117$&$0.381$\tabularnewline
(1,1,1)&$1.00$&$0.785$&$0.207$&$0.111$&$0.526$&$1.000$&$0.896$&$0.190$&$0.110$&$0.405$\tabularnewline
\hline
\end{tabular}
\label{tab:simu}
\end{center}
\end{table}

\subsection{Application to imaging genetics and epigenetics with schizophrenia}
Here it is demonstrated the power of our  proposed method and SKAT utilization for imaging genetic and epigenetic SZ  data collected  by MCIC. The key to integration, here, is to characterize the underlying interactions between the genetic features (gene-derived SNPs),  human brain  features (ROIs) and epigenetic features (gene-derived DNA methylation)  with covariates (age, height, weight) on hippocampal volume derived from structural MRI scans of the SZ. To do this, we extracted significant (gene-derived SNPs)-ROI-(gene-derived DNA methylations)  interactions using the proposed  method and compared them  to the SKAT methods. 

\begin{figure}
\begin{center}
  \includegraphics[width=14cm, height=8cm]{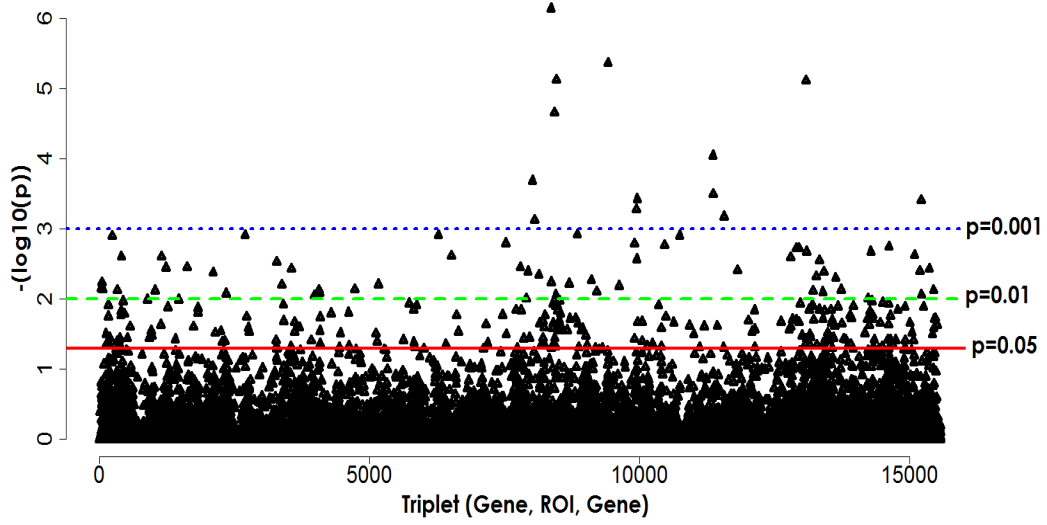}
\caption{The  plot of  $-\rm{log}_{10} (p)$ with  $15436$ triplets.}
\label{fig:pval}
\end{center}
\end{figure}

By considering $72$ genes-derived SNP, $116$ ROIs and $129$ gene-derived DNA methylation, we have $1077408\, (72\times 116\times 129)$ triplets.  By  the overall tests,  we obtained  $15436$ significant  triplets  at a $5\%$ level ($p$ $\leq 0.05$). Figure \ref{fig:pval} visualizes the index plot of  $-\rm{log}_{10} (p)$ for  $15436$ triplets (the triplets in X-axis and $-\rm{log}_{10} (p)$ in Y-axis). The vertical solid, doted  and double doted lines  indicate the  p-values at $0.05$, $0.01$,  $0.001$, respectively. Based on these lines, we observed that $272$, $72$,  and $13$  triplets are identified to have significantly higher order interactions   at $0.05$, $0.01$ and $0.001$ levels, respectively.

Table~\ref{tab:tgrg} presents  the ReML estimates of $\sigma^2$, $\tau^{(1)}$,  $\tau^{(2)}$, $\tau^{(3)}$,   $\tau^{(1\times 2)}$,  $\tau^{(1\times 3)}$,  $\tau^{(2\times 3)}$,  $\tau^{(1\times 2\times 3)}$ and  the $p$-values for the proposed  and SKAT methods on  each of the  $13$ triplets, which  are identified to have interaction  significance at a level of  $0.001$. At this  $p$-value, we have $6$  gene-derives SNPs ({\bf IL1B, MAGI2, NRG1, PDLIM5, SLC18A1, TDRD3}),  $10$ ROIs  ({\bf CRBL8.L, CRBLCrus1.L,  ORBSUP.R,  LING.L, CAU.R, IPL.L, IPL.R, PoCG.L, ITG.R, VER54}), and 6 gene-derived DNA methylations ({\bf CRABP1, FBXO28, DUSP1, FHIT, PLAGL1, TFPI2}) that have   significant interaction effects on the hippocampal volume of  SZ patients.

\begin{figure*}
\begin{center}
\includegraphics[width=14cm, height=10cm]{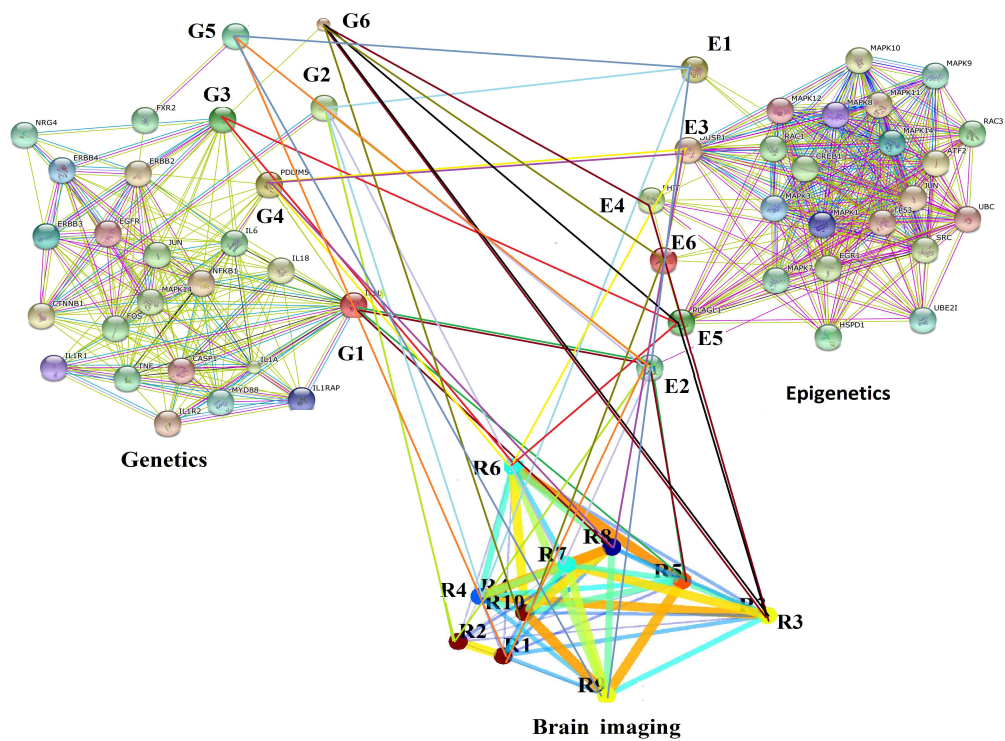}
\caption{The network graph of  features interaction  of each individual view and  features  interactions of  among others  view ($p\leq 0.001$).  Each node represents genes-derived SNP ( {\bf G1}: {\bf  IL1B}, {\bf G2}: {\bf MAGI2}, {\bf G3}: {\bf NRG1}, {\bf G4}: {\bf PDLIM5}, {\bf G5}: {\bf SLC18A1},  {\bf G6:} {\bf TDRD}), ROIs ({\bf R1}: {\bf CRBL8.L}, {\bf R2}: {\bf CRBLCrus1.L}, {\bf R3}: {\bf ORBSUP.R}, {\bf R4}: {\bf LING.L}, {\bf R5}: {\bf CAU.R},  {\bf R6}: {\bf IPL.L},  {\bf R7}: {\bf IPL.R}, {\bf R8}: {\bf PoCG.L}, {\bf R9}: {\bf ITG.R}, {\bf R10}: {\bf VER54})  and gene-derived DNA methylation ({\bf E1}: {\bf CRABP1},   {\bf E2}: {\bf FBXO28},  {\bf E3}: {\bf DUSP1},  {\bf E4}:{\bf FHIT},  {\bf E5}: {\bf PLAGL1},  {\bf E6}: {\bf TFPI2}).}
\label{fig:grg}
\end{center}
\end{figure*}

Figure~\ref{fig:grg} shows  the network within each genetics, imaging and epigentics  interactions  as well as the interactions they have   between  all others  views. Each node represents the  gene-derived SNPs, ROIs and gene-derived DNA methylations, respectively. The interacting genes-derived SNPs, ROIs and gene-derived DNA methylations are  connected with lines. The  thickness of the connection line indicates the strength of the interaction  among genes-derived SNPs, ROIs and gene-derived DNA methylations. These selected gene-derived SNPs and gene-derived DNA methylations show the interactions between several other genes. The selected ROIs also show the interaction within each selected ROI (shown in Figure~\ref{fig:grg})  as well as the  other ROIs (not shown in the figure). Following many studies in the literature, we have shown that each selected gene-derived SNPs our method has identified also has robust research discussing its role in the expression of SZ disease \cite{Siaw-16, Shibuya-13, Koide-12, Harrison-06, Moselhy-15, Bly-05}.

\begin{figure}
\begin{center}
  \includegraphics[width=14cm, height=12cm]{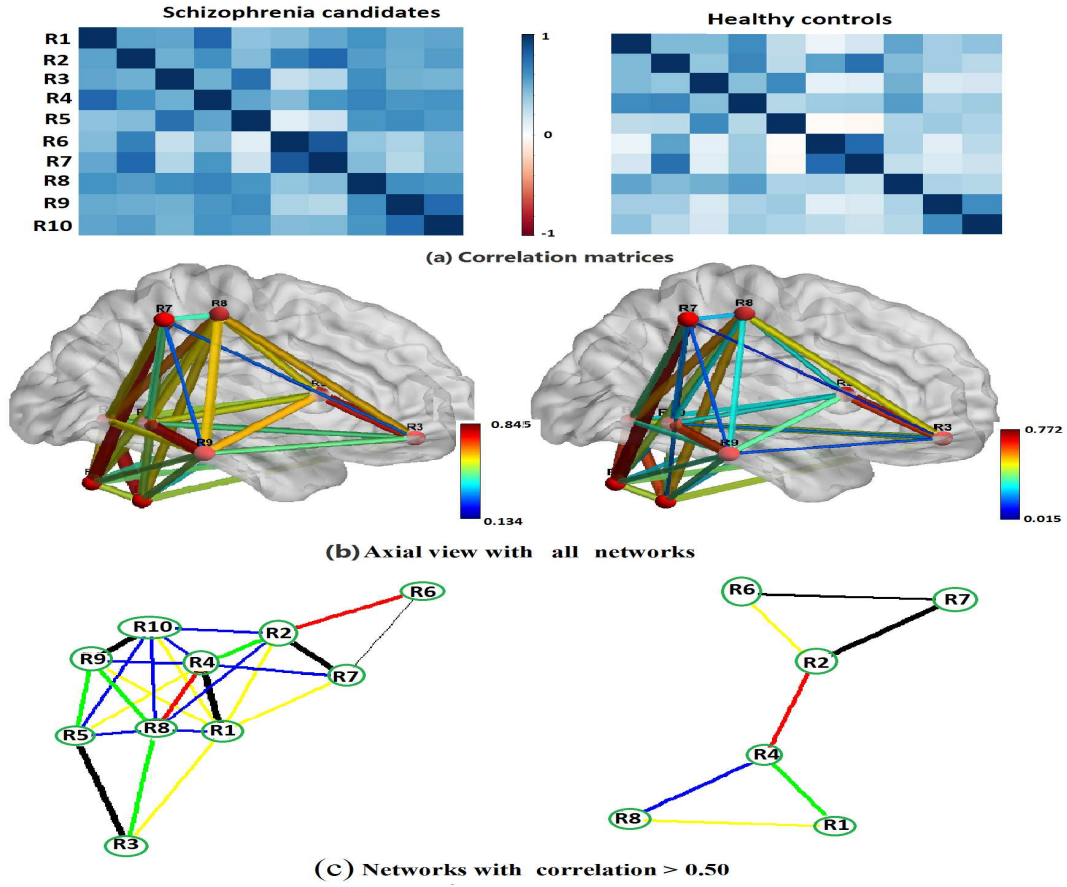}
\caption{The  visualization of  selected  $10$ ROIs for  schizophrenia candidate and healthy control groups: (a) correlation matrices, (b) axial view with all networks, (c) networks with correlation $> 0.5$.}
\label{fig:cmav}
\end{center}
\end{figure}

Recent research has also shown that the  $10$  ROIs selected by the proposed method   have  a critical role in  brain related diseases \cite{Suk-16, Chen-13, Wu-13}.  We additionally investigated the $10$  ROIs to confirm their role in  SZ.  To do this, each multidimensional variable ROI was converted to a univariate variable by taking the weighted mean.  We then evaluated  the differences between the SZ candidates and healthy controls  using  network measures and visualizations.  Table~\ref{tb:evroi} presents the transitivity, degree  and global efficiency of  each ROI for the SZ candidate and network and healthy control. From this table, we observed that the transitivity (measuring the probability that the adjacent vertices of a vertex are connected)  of the SZ candidate group  is larger  than  in  the  healthy control group (most of the ROIs and on average); this suggests that SZ tends to have more transitive triples. The degree (the number of edges incident to the vertex) of the  SZ candidate group  is larger than  in the  healthy control group for all of the ROIs; this indicates that these ROIs could have an impact on the SZ candidate.  The global efficiency, the mean of all nodal efficiencies, of the  SZ candidate group  is  different from the  healthy control group. This may suggest that functional activity of the SZ  candidate is not similar to the functional activity of the healthy control group in these regions. Figure~\ref{fig:cmav} shows the visualization of   correlation matrices, axial view with all networks and  networks with correlation $> 0.05$ for the SZ candidate and  healthy control group. From  Figure~\ref{fig:cmav}, it can be observed that the ROIs in the SZ candidate groups are more correlated and connected than the   healthy control group. Therefore,  with strong agreement, it has been shown that  the selected ROIs   have potential impact on the expression of  SZ disease. 

Table~\ref{tab:grg} lists the selected  significant gene-derived SNP, ROIs and gene-derived DNA methylation using the proposed method (KMDHOI) and SKAT at a $p\leq 0.01$.  We found that    $31$ genes-derived SNP, $35$ ROIs and $20$ genes-derived DNA methylation from $72$ triplets were  identified to have  significance on  the  hippocampal volume of the SZ patients and the healthy controls.  We also observed that $6$ gene-derived SNPs,  $10$ ROIs  and 6 gene-derived DNA methylations  were   significant at  a $p\leq 0.001$. The  underlined elements  indicated in Table~\ref{tab:grg} have  significant interaction triplets. Table~ $5$ \& $6$  (in the appendix) lists $72$ triplets, which were   significant at  a $p\leq 0.001$.      

  For the proposed KMDHOI approach, we considered triplets (gene-derived SNP, ROI, gene-derived DNA methylation)  with a $p\leq 4\time 10^{-8}$ to be statistically significant after the Bonferroni correction for  $1077408$ tests. Although the interaction (gene-derived SNP, ROI, gene-derived DNA methylation) results do not appear to be significant after adjusting for multiple comparisons, some of them appear  promising consistent results.  According to the  $p-$values,  we can determine the  gene-derived SNPs, ROIs, and gene-derived DNA methylations that have  a highly  significant hippocampal volume on  SZ patients and healthy controls. We  observed  gene-derived SNP ({\bf MAGI2, NRG1, SLC18A1, TDRD3}), ROIs ({\bf CRBL8.L, CRBLCrus1.L, ORBSUP.R, LING.L, IPL.L, IPL.R}) and gene-derived DNA methylations ({\bf CRABP1, FBXO28, FHIT, PLAGL1}) at a $p\leq 0.0001$, gene-derived SNP  ({\bf MAGI2, NRG1, TDRD3}), ROIs  ({\bf CRBL8.L, CRBLCrus1.L, IPL.L, IPL.R}) and gene-derived DNA methylation  ({\bf FBXO28,  PLAGL1})  at a  $p\leq 0.00001$,   and genes-derived SNP  ({\bf MAGI2}), ROIs  ({\bf CRBLCrus1.L}), and gene-derived DNA methylation  ({\bf FBXO28})  at  a $p\leq 0.000001$, which  are   identified to have  high  interaction effects on  hippocampal volume of  SZ patients and healthy control.

 To confirm this discovery,  we used the DAVID, and  gene ontology (GO)  enrichment analysis to find the most relevant GO terms associated with the selected $31$ genes. The selected genes are associated with a set of annotation terms. We compared $5$ annotation categories, including literature, disease, gene ontology, pathways and protein interaction using DAVID \cite{DAVID}. Table~ $7$ (in the appendix) presents five  annotation categories of the $31$ selected genes. From this table, we observed that the selected genes have had remarkable literature review  done in   past studies. According to the disease annotation, the selected genes are highly associated with complex diseases including SZ, cognitive function,  bipolar disorder, and others. By GO annotation, the selected genes have significant relationship to single-organism processes, response to stimuli, developmental processes and etc.  From the table, we observed that the selected genes have a significant pathway  to facilitate biological interpretation in a network context.  Moreover, protein interaction annotations show that the selected genes have been discussed in many biomedical papers \cite{Sanders-08, Gerhard-04, Strausberg-02}. 

Genes do not function alone.  Rather, they interact  with each other. When genes share a similar set of GO annotation terms, they are most likely to be involved in similar biological mechanisms. To  confirm this,  we extracted the (gene-derived SNPs)-(gene-derived DNA methylations)  network using STRING \cite{STRING-15}. STRING imports protein association knowledge from databases of physical interaction and databases of curated biological pathway knowledge. In STRING, the simple interaction unit is the functional association (functional relationship between two proteins/ genes) that is  most likely  contributing to a common biological purpose. In this view, the color saturation of the edges represents the confidence score of a functional association. Further network analysis shows that the number of nodes, expected number of edges, number of edges, average node degree, clustering coefficient, PPI enrichment $p$-values are $51$, $93$, $300$, $11.8$, $0.603$, and $p\leq 0\times 10^{-16}$, respectively \cite{STRING-15}. This network has significantly more interactions than expected. This means that these genes have more interactions among themselves than what would be expected for a random set of genes of similar size drawn from the genome. Such an enrichment indicates that the proteins/genes are at least  biologically connected as a group.

\begin{table}[!tbp]
\begin{center}
\caption {The selected  significant genes-derived SNP, ROIs and gene-derived DNA methylation using the proposed method (KMDHOI) and SKAT. The $p-$values threshold was fixed to $0.001$.}
\scalebox{0.6}[.9]{
\begin{tabular}{lcccccccccccccc}
\hline
&&&\multicolumn{10}{c}{\rm{KMDHOI}}&\multicolumn{1}{c}{\rm{SKAT}}\tabularnewline
\multicolumn{1}{l}{Genetics}&\multicolumn{1}{c}{Imaging}&\multicolumn{1}{c}{Epigenetics}&\multicolumn{1}{c}{$\sigma^2$}&\multicolumn{1}{c}{$\tau^{(1)}$}&\multicolumn{1}{c}{$\tau^{(2)}$}&\multicolumn{1}{c}{$\tau^{(3)}$}&\multicolumn{1}{c}{$\tau^{1\times 2}$}&\multicolumn{1}{c}{$\tau^{1\times 3}$}&\multicolumn{1}{c}{$\tau^{2\times 3}$}&\multicolumn{1}{c}{$\tau^{1\times 2\times 3}$} &\multicolumn{1}{c}{\rm{OVA}} &\multicolumn{1}{c}{\rm{HOI}} &\multicolumn{1}{c}{\rm{HOI}}\tabularnewline\hline
${\bf IL1B}$&${\bf CAU.R}$&$    {\bf FBXO28}$&$0.6755$&$0.0038$&$0.0229$&$0.1225$&$1.1013$&$0.0000$&$0.1307$&$1.3606$&$0.0383$&$2E-04$&$0.4943$\tabularnewline
${\bf IL1B}$&$ {\bf PoCG.R}$&$  {\bf FBXO28}$&$0.5837$&$0.0189$&$0.1827$&$0.1247$&$1.8403$&$0.0000$&$0.3469$&$1.0603$&$0.0202$&$7E-04$&$0.2871$\tabularnewline
${\bf MAGI2}$&${\bf CRBLCrus1.L}$&$     {\bf FBXO28}$&$0.1833$&$0.3246$&$0.0000$&$0.2693$&$1.1963$&$2.2426$&$1.3683$&$0.0100$&$0.0288$&$0E-06$&$0.01891$\tabularnewline
${\bf MAGI2}$&${\bf LING.L}$&$  {\bf CRABP1}$&$0.1813$&$0.4366$&$0.0000$&$0.0885$&$1.5370$&$2.6299$&$1.0271$&$0.0100$&$0.0470$&$0E-05$&$0.4030$\tabularnewline
${\bf MAGI2}$&$ {\bf IPL.R}$&$  {\bf FBXO28}$&$0.1833$&$0.3778$&$0.0000$&$0.3044$&$0.9808$&$2.3888$&$1.2203$&$0.0100$&$0.0457$&$0E-05$&$0.5592$\tabularnewline
$ {\bf NRG1}$&$ {\bf IPL.L}$&${\bf PLAGL1}
$&$0.3682$&$0.0024$&$0.2162$&$0.1270$&$1.1227$&$2.7930$&$0.0000$&$0.2056$&$0.0284$&$0E-05$&$0.6664$\tabularnewline
${\bf PDLIM5}$&${\bf IPL.L}$&$ {\bf DUSP1}$&$0.3648$&$0.0000$&$0.0804$&$0.2182$&$1.5177$&$1.9409$&$1.0276$&$0.0100$&$0.0183$&$5E-04$&$0.0173$\tabularnewline
${\bf PDLIM5}$&${\bf PoCG.R}$&$ {\bf DUSP1}$&$0.3598$&$0.0000$&$0.2256$&$0.0139$&$0.9189$&$1.7853$&$1.2796$&$0.0100$&$0.0498$&$Ee-04$&$0.2761$\tabularnewline
${\bf SLC18A1}$&$ {\bf ORBsup.R}$&$ {\bf FHIT}$&$0.4096$&$0.0065$&$0.5356$&$0.2276$&$1.3456$&$0.0000$&$1.0003$&$0.1323$&$0.0495$&$1E-04$&$0.0522$\tabularnewline
${\bf SLC18A1}$&${\bf ORBsup.R}$&${\bf PLAGL1}
$&$0.2869$&$0.0000$&$0.3909$&$0.1933$&$1.0186$&$1.3148$&$1.3676$&$0.0100$&$0.0373$&$3E-04$ &$0.0149$\tabularnewline
${\bf SLC18A1}$&${\bf Vermis45}$&${\bf TFPI2}$&$0.5571$&$0.0447$&$0.0815$&$0.0020$&$0.0000$&$1.0458$&$0.5579$&$0.0100$&$0.0354$&$7E-04$&$0.4234$\tabularnewline
${\bf TDRD3}$&${\bf CRBL8.L}$&$ {\bf FBXO28}$&$0.5856$&$0.2284$&$0.0022$&$0.0702$&$1.1722$&$0.7667$&$0.0000$&$0.0100$&$0.0052$&$0E-05$&$0.0807$\tabularnewline
${\bf TDRD3}$&${\bf ITG.R}$&$   {\bf CRABP1}$&$0.5291$&$0.2318$&$0.0000$&$0.0223$&$0.7052$&$0.6038$&$0.6240$&$0.0100$&$0.0033$&$4E-04$&$0.0359$
\tabularnewline
\hline
\end{tabular}
}
\label{tab:tgrg}
\end{center}
\end{table}

\begin{table*}
 \begin{center}
\caption {The network  measurements (transitivity, degree, and global efficiency) of the selected $10$ ROIs for  schizophrenia candidate and healthy control groups. }
\label{tb:evroi}
 \begin{tabular}{lcccccccccc} \hline
&\multicolumn{2}{c}{\rm{Transitivity}}&\multicolumn{2}{c}{\rm{Degree}} &\multicolumn{2}{c}{\rm{Global efficiency}}\tabularnewline
\rm{ROIs} & \rm{Schizophrenia} & \rm{Healthy} & \rm{Schizophrenia} & \rm{Healthy} & \rm{Schizophrenia} & \rm{Healthy} \tabularnewline\hline
{\bf R1 = CRBL8.L}&  $0.571$&$ 1.000$&$ 7 $&$ 2$&$ 0.800$&$ 0.317$\tabularnewline
{\bf R2 = CRBLCrus1.L}&$  0.600$&$ 0.333$&$  6 $&$ 3 $&$0.750$&$ 0.400$\tabularnewline
{\bf R3 = ORBSUP.R}&$  0.667 $&$0.000$&$ 3 $&$ 1 $&$0.5833$&$ 0.100$\tabularnewline
{\bf R4 = LING.L}&$  0.667$&$ 0.333$&$  7$&$  3 $&$0.800$&$ 0.400$\tabularnewline
{\bf R5 = CAU.R}&$  0.700$&$  0.000$&$  5$&$  1 $&$0.683$&$ 0.100$\tabularnewline
{\bf R6 = IPL.L}&$  1.000$&$ 1.000$&$  2$&$  2 $&$0.500$&$ 0.317$\tabularnewline
{\bf R7 = IPL.R}&$  0.667$&$ 1.000$&$  4$&$  2$&$ 0.650$&$0.317$\tabularnewline
{\bf R8 = PoCG.L}&$  0.667 $&$1.000$&$  7$&$  2$&$ 0.800$&$ 0.3167$\tabularnewline
{\bf R9 = ITG.R}&$  0.900$&$0.000$&$5$&$  1$&$ 0.683$&$ 0.100$\tabularnewline
{\bf R10 = VER45}&$ 0.800$&$0.000$&$  6$&$  1$&$ 0.750 $&$0.100
$\tabularnewline\hline
\end{tabular}           
\end{center}
\end{table*}

\begin{figure}[ht]
\begin{center}
  \includegraphics[width=12cm, height=8cm]{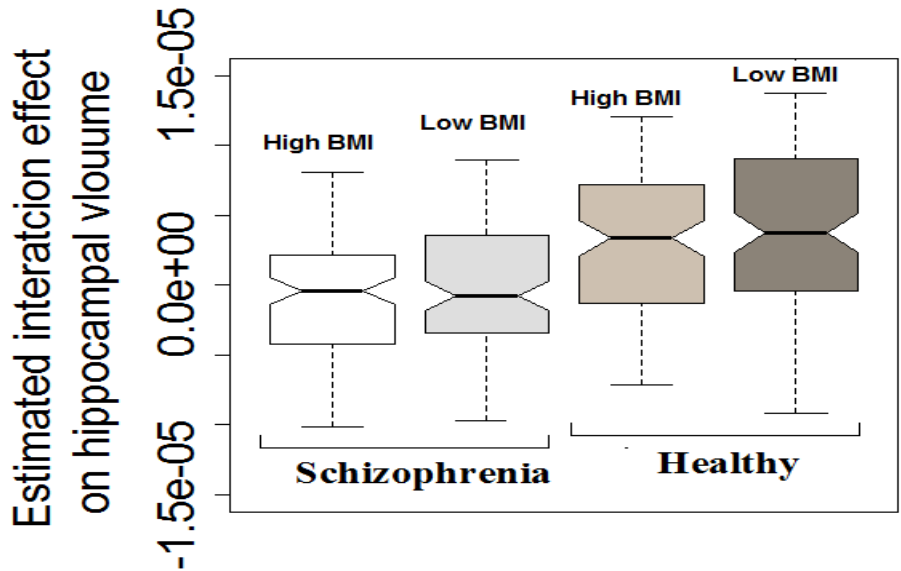}
\caption{Boxplost of  significant interaction effects in different regimes (SZ and high BMI risk, SZ and low  BMI risk, healthy control and  high BMI risk, healthy control and low  BMI risk) for the most significant   triplet  ({\bf MAGI2, CRBLCrus1.L, FBXO28}).}
\label{fig:bie}
\end{center}
\end{figure}

\begin{table}[!tbp]
\begin{center}
\caption {The selected  significant gene-derived SNPs, ROIs and gene-derived DNA methylations using the proposed method (KMDHOI) and SKAT at $p\leq 0.01$. The bold indicates  significant at $p\leq 0.001$. Note: the name of ROI is given by the AAL template.}
\scalebox{0.6}[.9]{
\begin{tabular}{lcccccccccccccccc}
%\rm{Data views}&\tabularnewline 
\tabularnewline\hline
\rm{Genetics}&{\bf IL1B}&{\bf MAGI2}& {\bf NRG1}& {\bf PDLIM5}& {\bf SLC18A1}& {\bf TDRD3}&BDNF
&CHGA&CHGB&CLINT1\tabularnewline
&COMTD1&DAOA&DISC1& DRD2& DTNBP1&ERBB4& GABBR1 & GABRB2& GRIN2B& GRM3\tabularnewline 
&HTR2A& IL10RA& MAGI1& MICB& NOS1AP& NOTCH4& NR4A2& NUMBL& PLXNA2& PPP3CC \tabularnewline
&SNAP29 \tabularnewline\hline
\rm{Imaging}
&{\bf CRBL8.L}& {\bf CRBLCrus1.L}& {\bf ORBSUP.R}& {\bf LING.L}& {\bf CAU.R}& {\bf IPL.L}&{\bf IPL.R}&{\bf PoCG.L}& {\bf ITG.R}& {\bf VER45}\tabularnewline
&AMYG.L&CRBL10.R&CRBL10.L&CRBL3.R&CRBL3.R&CRBL45.L&CRBL6.L&CRBL8.R&CRBLCrus2.R&CRBLCrus2.L\tabularnewline
&DCG.R&DCG.L&PCG.R&ORBsup.L&ORBmid.R&LING.R&ROL.R&SMA.R&TPOsup.R&TPOsup.L\tabularnewline
 &STG.L&ITG.L&Vermis10&Vermis3&MTG.R\tabularnewline\hline
\rm{Epigenetic}&{\bf CRABP1}&{\bf FBXO28}&{\bf DUSP1}&{ \bf FHIT}&{\bf PLAGL1}&{\bf TFPI2}&CCND2&CDKN1A
&EDNRB&ESR1\tabularnewline
&EYA4&FEN1&GPSN2&HOXA9&HOXB4&PTGS2&RB1&SRF&WDR37&ZNF512\tabularnewline
\hline
\end{tabular}
}
\label{tab:grg}
\end{center}
\end{table}
 Lastly, we conducted standard logistic regression analysis with covariates of  age, gender, and BMI on the outcome of SZ disease (SZ vs healthy control). We found that BMI is a significant covariate for the SZ vs healthy control  at  a $p\leq 0.0353$.  Thus,  BMI is one of the risk factors of SZ disease.  For a BMI $\geq 25$, we considered the subject to be a high risk. Based on this risk, we divided the estimated higher order interaction effect $\hat{\vc{h}}_{\vc{M}^{(1)}\times \vc{M}^{(2)}\times \vc{M}^{(3)}}$ values  into four regimes: SZ with high BMI risk,  SZ with low  BMI risk, healthy control with  high BMI risk, and healthy control with low  BMI risk. Figure~\ref{fig:bie} shows the boxplots of the  estimated interaction effect within each of the four regimes  for the most significant   triplet ({\bf MAGI2, CRBLCrus1.L, and FBXO28}). The small variation indicates a higher risk of the interaction effect (hippocampal volume).  This figure shows that the  SZ and BMI risks largely dominate the interaction effect (i.e., higher SZ  and BMI risk associated with higher risk of interaction) and vice versa. 

\section{Discussion  and future research}
 \label{sec:cond}
In this paper, we have proposed a semiparametric  kernel method for higher order interactions between multiple data sets.  Compared to the traditional PCA multiple regression and SKAT methods, the proposed method shows a more flexible and biological plausible way to model higher order epistasis among the genetic, imaging, and epigenetic  data.  While kernel based methods on multi-view data naturally produce more powerful and reproducible results, and are biologically more meaningful, the interpretation of  model parameters is often challenging.  Incorporating the gene and pathway analysis of biological information would facilitate additional improvements of model interpretation. 

 The performance of the proposed  method was evaluated  on both simulated  and real  MCIC data.   The extensive simulation studies show evidence of the power gain of the proposed  method relative to the alternative methods and  suggest that the  proposed methods  perform remarkably better than the  dimension  reduction multiple regression and SKAT methods. 

 The utility of the proposed method is further demonstrated with the application to  imaging genetics  study of SZ. According to the  $p-$values,  the proposed method is able to  rank    the triplets (gene-derived SNPs)-ROI-(gene-derived DNA methylations) and subset of triplets  can be selected which are highly related to SZ disease.  At a $p\leq 0.01$ the proposed method extract the unique   $31$ genes-derived SNP, $35$ ROIs and $20$ gene-derived DNA methylation from $72$ triplets, which are  identified to have  significant impact on  hippocampal volume of  SZ patients. By  conducting gene ontology, pathway analysis, and several network measures including visualizations,  we find evidence that the selected (gene-derived SNPs)- ROI-(gene-derived DNA methylations)  have a significant influence on the manifestation of SZ disease. The  identified triplets suggest that these statistical and biologically significant triplets may  an important  role in SZ related neurodegenerations. Our findings have indicated that genetic elements interplay with brain  regions and epigenetic factors. 

While we illustrated the proposed model using a quantitative hippocampal volume derived from structural MRI image phenotype, the utility of this model is that it can be applied to any phenotypes to detect  higher order interactions in  genetics, imaging, and epigenetic features, to include environmental covariates. The proposed model  can also be extended to qualitative phenotypes for potentially widely applicable  case-control studies (e.g., generalized kernel logistic regression).

It must be repeated that choosing a suitable kernel is indispensable. Kernel parameters may strongly influence the result desired for its application. Although  the linear kernel  does not have any free parameters, the linear kernel  has certain limitations. Using the polynomial kernel makes it  possible to detect  higher order correlations. Polynomial kernels only map data into a finite dimensional space.  In addition, both linear and polynomial kernels are unbounded.  Many radial basis function kernels, such as the Gaussian  kernel, map  input data into an infinite dimensional space. The Gaussian  kernel has a free parameter (bandwidth) but  follows a number of   properties (e.g., boundedness,  consistency, universality, and  robustness).  

In this study,  while we applied the median of the  pairwise distance as a  bandwidth for the  Gaussian kernel, future studies might also compare the higher order interaction effects using a number of different kernels with  different parameters, which may have broad implications to the detection of higher order interactions between disease phenotypes as described in the methods of this paper.

\subsection*{Acknowledgments}
%We would like to thank the reviewers for their careful reading of the manuscript and their useful comments.
 The authors wish to thank the NIH (R01 GM109068, R01 MH104680, ROI MH107354) and NSF EPSCoR program  (1539067) for support.
%\newpage
\begingroup
\bibliographystyle{apacite}
\bibliography{Ref-UKIF}

\begin{thebibliography}{}

\bibitem [\protect \citeauthoryear {%
Aberg%
, McClay%
, Nerella%
\BCBL {}\ \BBA {} et al.%
}{%
Aberg%
\ \protect \BOthers {.}}{%
{\protect \APACyear {2014}}%
}]{%
Aberg-14}
\APACinsertmetastar {%
Aberg-14}%
\begin{APACrefauthors}%
Aberg, K\BPBI A.%
, McClay, J\BPBI L.%
, Nerella, S.%
\BCBL {}\ \BBA {} et al., S\BPBI C.%
\end{APACrefauthors}%
\unskip\
\newblock
\APACrefYearMonthDay{2014}{}{}.
\newblock
{\BBOQ}\APACrefatitle {Methylome-Wide Association Study of Schizophrenia
  Identifying Blood Biomarker Signatures of Environmental Insults}
  {Methylome-wide association study of schizophrenia identifying blood
  biomarker signatures of environmental insults}.{\BBCQ}
\newblock
\APACjournalVolNumPages{JAMA Psychiatry}{71(3)}{}{255-264}.
\PrintBackRefs{\CurrentBib}

\bibitem [\protect \citeauthoryear {%
Alam%
}{%
Alam%
}{%
{\protect \APACyear {2014}}%
}]{%
Ashad-14T}
\APACinsertmetastar {%
Ashad-14T}%
\begin{APACrefauthors}%
Alam, M\BPBI A.%
\end{APACrefauthors}%
\unskip\
\newblock
\APACrefYearMonthDay{2014}{}{}.
\newblock
{\BBOQ}\APACrefatitle {Kernel Choice for Unsupervised Kernel Methods} {Kernel
  choice for unsupervised kernel methods}.{\BBCQ}
\newblock
\APACaddressPublisher{Japan}{PhD. Dissertation, The Graduate University for
  Advanced Studies}.
\PrintBackRefs{\CurrentBib}

\bibitem [\protect \citeauthoryear {%
Alam%
, Calhoun%
\BCBL {}\ \BBA {} Wang%
}{%
Alam%
, Calhoun%
\BCBL {}\ \BBA {} Wang%
}{%
{\protect \APACyear {2016}}%
}]{%
Alam-16a}
\APACinsertmetastar {%
Alam-16a}%
\begin{APACrefauthors}%
Alam, M\BPBI A.%
, Calhoun, V.%
\BCBL {}\ \BBA {} Wang, Y\BPBI P.%
\end{APACrefauthors}%
\unskip\
\newblock
\APACrefYearMonthDay{2016}{}{}.
\newblock
{\BBOQ}\APACrefatitle {Influence Function of Multiple Kernel Canonical Analysis
  to Identify Outliers in Imaging Genetics Data} {Influence function of
  multiple kernel canonical analysis to identify outliers in imaging genetics
  data}.{\BBCQ}
\newblock
\APACjournalVolNumPages{Proceedings of 7th ACM Conference on Bioinformatics,
  Computational Biology, and Health Informatics (ACM BCB),Seattle, WA,
  USA}{}{}{210-2198}.
\PrintBackRefs{\CurrentBib}

\bibitem [\protect \citeauthoryear {%
Alam%
\ \BBA {} Fukumizu%
}{%
Alam%
\ \BBA {} Fukumizu%
}{%
{\protect \APACyear {2014}}%
}]{%
Ashad-14}
\APACinsertmetastar {%
Ashad-14}%
\begin{APACrefauthors}%
Alam, M\BPBI A.%
\BCBT {}\ \BBA {} Fukumizu, K.%
\end{APACrefauthors}%
\unskip\
\newblock
\APACrefYearMonthDay{2014}{}{}.
\newblock
{\BBOQ}\APACrefatitle {Hyperparameter Selection in Kernel Principal Component
  Analysis} {Hyperparameter selection in kernel principal component
  analysis}.{\BBCQ}
\newblock
\APACjournalVolNumPages{Journal of Computer Science}{10(7)}{}{1139--1150}.
\PrintBackRefs{\CurrentBib}

\bibitem [\protect \citeauthoryear {%
Alam%
\ \BBA {} Fukumizu%
}{%
Alam%
\ \BBA {} Fukumizu%
}{%
{\protect \APACyear {2015}}%
}]{%
Ashad-15}
\APACinsertmetastar {%
Ashad-15}%
\begin{APACrefauthors}%
Alam, M\BPBI A.%
\BCBT {}\ \BBA {} Fukumizu, K.%
\end{APACrefauthors}%
\unskip\
\newblock
\APACrefYearMonthDay{2015}{}{}.
\newblock
{\BBOQ}\APACrefatitle {Higher-Order Regularized Kernel Canonical Correlation
  Analysis} {Higher-order regularized kernel canonical correlation
  analysis}.{\BBCQ}
\newblock
\APACjournalVolNumPages{International Journal of Pattern Recognition and
  Artificial Intelligence}{29(4)}{}{1551005(1-24)}.
\PrintBackRefs{\CurrentBib}

\bibitem [\protect \citeauthoryear {%
Alam%
, Komori%
, Calhoun%
\BCBL {}\ \BBA {} Wang%
}{%
Alam%
, Komori%
\BCBL {}\ \protect \BOthers {.}}{%
{\protect \APACyear {2016}}%
}]{%
Alam-16b}
\APACinsertmetastar {%
Alam-16b}%
\begin{APACrefauthors}%
Alam, M\BPBI A.%
, Komori, O.%
, Calhoun, V.%
\BCBL {}\ \BBA {} Wang, Y\BPBI P.%
\end{APACrefauthors}%
\unskip\
\newblock
\APACrefYearMonthDay{2016}{}{}.
\newblock
{\BBOQ}\APACrefatitle {Robust Kernel Canonical Correlation Analysis to Detect
  Gene-Gene Interaction for Imaging Genetics Data} {Robust kernel canonical
  correlation analysis to detect gene-gene interaction for imaging genetics
  data}.{\BBCQ}
\newblock
\APACjournalVolNumPages{Proceedings of 7th ACM Conference on Bioinformatics,
  Computational Biology, and Health Informatics (ACM BCB),Seattle, WA,
  USA}{}{}{279-288}.
\PrintBackRefs{\CurrentBib}

\bibitem [\protect \citeauthoryear {%
Andreasen%
}{%
Andreasen%
}{%
{\protect \APACyear {1984}}%
}]{%
Andreasen-84}
\APACinsertmetastar {%
Andreasen-84}%
\begin{APACrefauthors}%
Andreasen, N\BPBI C.%
\end{APACrefauthors}%
\unskip\
\newblock
\APACrefYearMonthDay{1984}{}{}.
\newblock
{\BBOQ}\APACrefatitle {Scale for the assessment of positive symptoms (SAPS)}
  {Scale for the assessment of positive symptoms (saps)}.{\BBCQ}
\newblock
\APACaddressPublisher{Iowa City, University of Iowa}{Springer}.
\PrintBackRefs{\CurrentBib}

\bibitem [\protect \citeauthoryear {%
Aronszajn%
}{%
Aronszajn%
}{%
{\protect \APACyear {1950}}%
}]{%
Aron-RKHS}
\APACinsertmetastar {%
Aron-RKHS}%
\begin{APACrefauthors}%
Aronszajn, N.%
\end{APACrefauthors}%
\unskip\
\newblock
\APACrefYearMonthDay{1950}{}{}.
\newblock
{\BBOQ}\APACrefatitle {Theory of reproducing kernels} {Theory of reproducing
  kernels}.{\BBCQ}
\newblock
\APACjournalVolNumPages{Transactions of the American Mathematical
  Society}{68}{}{337-404}.
\PrintBackRefs{\CurrentBib}

\bibitem [\protect \citeauthoryear {%
Bis%
, DeCarli%
\BCBL {}\ \BBA {} et al.%
}{%
Bis%
\ \protect \BOthers {.}}{%
{\protect \APACyear {2012}}%
}]{%
Bis-12}
\APACinsertmetastar {%
Bis-12}%
\begin{APACrefauthors}%
Bis, J\BPBI C.%
, DeCarli, C.%
\BCBL {}\ \BBA {} et al., A\BPBI S.%
\end{APACrefauthors}%
\unskip\
\newblock
\APACrefYearMonthDay{2012}{}{}.
\newblock
{\BBOQ}\APACrefatitle {Common variants at 12q14 and 12q24 are associated with
  hippocampal volume} {Common variants at 12q14 and 12q24 are associated with
  hippocampal volume}.{\BBCQ}
\newblock
\APACjournalVolNumPages{Nature Genetics}{44(5)}{}{545-551}.
\PrintBackRefs{\CurrentBib}

\bibitem [\protect \citeauthoryear {%
B.~K.~Sriperumbudur%
\ \BBA {} Sch{\"{o}}lkopf%
}{%
B.~K.~Sriperumbudur%
\ \BBA {} Sch{\"{o}}lkopf%
}{%
{\protect \APACyear {2009}}%
}]{%
Sriperumbudur-09}
\APACinsertmetastar {%
Sriperumbudur-09}%
\begin{APACrefauthors}%
B.~K.~Sriperumbudur, A\BPBI G\BPBI G\BPBI R\BPBI G\BPBI L., K.~Fukumizu%
\BCBT {}\ \BBA {} Sch{\"{o}}lkopf, B.%
\end{APACrefauthors}%
\unskip\
\newblock
\APACrefYearMonthDay{2009}{}{}.
\newblock
{\BBOQ}\APACrefatitle {Kernel Choice and Classifiability for RKHS Embeddings of
  Probability Distributions} {Kernel choice and classifiability for rkhs
  embeddings of probability distributions}.{\BBCQ}
\newblock
\APACjournalVolNumPages{Advances in Neural Information Processing
  Systems}{21}{}{1750-1758}.
\PrintBackRefs{\CurrentBib}

\bibitem [\protect \citeauthoryear {%
Bly%
}{%
Bly%
}{%
{\protect \APACyear {2005}}%
}]{%
Bly-05}
\APACinsertmetastar {%
Bly-05}%
\begin{APACrefauthors}%
Bly, M.%
\end{APACrefauthors}%
\unskip\
\newblock
\APACrefYearMonthDay{2005}{}{}.
\newblock
{\BBOQ}\APACrefatitle {Mutation in the vesicular monoamine gene, SLC18A1,
  associated with schizophrenia} {Mutation in the vesicular monoamine gene,
  slc18a1, associated with schizophrenia}.{\BBCQ}
\newblock
\APACjournalVolNumPages{Schizophrenia Research}{78}{}{337-338}.
\PrintBackRefs{\CurrentBib}

\bibitem [\protect \citeauthoryear {%
Calhoun%
\ \BBA {} Sui%
}{%
Calhoun%
\ \BBA {} Sui%
}{%
{\protect \APACyear {2016}}%
}]{%
Calhoun-16}
\APACinsertmetastar {%
Calhoun-16}%
\begin{APACrefauthors}%
Calhoun, V\BPBI D.%
\BCBT {}\ \BBA {} Sui, J.%
\end{APACrefauthors}%
\unskip\
\newblock
\APACrefYearMonthDay{2016}{}{}.
\newblock
{\BBOQ}\APACrefatitle {Multimodal fusion of brain imaging data: A key to
  finding the missing link(s) in complex mental illness} {Multimodal fusion of
  brain imaging data: A key to finding the missing link(s) in complex mental
  illness}.{\BBCQ}
\newblock
\APACjournalVolNumPages{Biol Psychiatry Cogn Neurosci
  Neuroimaging}{1}{}{230-244}.
\PrintBackRefs{\CurrentBib}

\bibitem [\protect \citeauthoryear {%
Camps-Valls%
, Rojo-Alvarex%
\BCBL {}\ \BBA {} Martinez-Romon%
}{%
Camps-Valls%
\ \protect \BOthers {.}}{%
{\protect \APACyear {2007}}%
}]{%
Camps-07}
\APACinsertmetastar {%
Camps-07}%
\begin{APACrefauthors}%
Camps-Valls, G.%
, Rojo-Alvarex, J\BPBI L.%
\BCBL {}\ \BBA {} Martinez-Romon, M.%
\end{APACrefauthors}%
\unskip\
\newblock
\APACrefYearMonthDay{2007}{}{}.
\newblock
{\BBOQ}\APACrefatitle {Kernel Methods in Bioengineering, Signal and Image}
  {Kernel methods in bioengineering, signal and image}.{\BBCQ}
\newblock
\APACaddressPublisher{London}{Idea Group publishing}.
\PrintBackRefs{\CurrentBib}

\bibitem [\protect \citeauthoryear {%
Chang%
, Kruger%
, Kustra%
\BCBL {}\ \BBA {} Zhang%
}{%
Chang%
\ \protect \BOthers {.}}{%
{\protect \APACyear {2013}}%
}]{%
Chang-13}
\APACinsertmetastar {%
Chang-13}%
\begin{APACrefauthors}%
Chang, B.%
, Kruger, U.%
, Kustra, R.%
\BCBL {}\ \BBA {} Zhang, J.%
\end{APACrefauthors}%
\unskip\
\newblock
\APACrefYearMonthDay{2013}{}{}.
\newblock
{\BBOQ}\APACrefatitle {Canonical Correlation Analysis based on Hilbert-Schmidt
  Independence Criterion and Centered Kernel Target Alignment} {Canonical
  correlation analysis based on hilbert-schmidt independence criterion and
  centered kernel target alignment}.{\BBCQ}
\newblock
\APACjournalVolNumPages{Proceedings of the $30$th International Conference on
  Ma- chine Learning, Atlanta, Georgia, USA}{}{}{}.
\PrintBackRefs{\CurrentBib}

\bibitem [\protect \citeauthoryear {%
Chekouo%
, Stingo%
, Guindani%
\BCBL {}\ \BBA {} Do%
}{%
Chekouo%
\ \protect \BOthers {.}}{%
{\protect \APACyear {2016}}%
}]{%
Chekouo-16}
\APACinsertmetastar {%
Chekouo-16}%
\begin{APACrefauthors}%
Chekouo, T.%
, Stingo, F\BPBI C.%
, Guindani, M.%
\BCBL {}\ \BBA {} Do, K\BPBI A.%
\end{APACrefauthors}%
\unskip\
\newblock
\APACrefYearMonthDay{2016}{}{}.
\newblock
{\BBOQ}\APACrefatitle {A Bayesian Predictive Model for Imaging genetics with
  Application to Schizophrenia} {A bayesian predictive model for imaging
  genetics with application to schizophrenia}.{\BBCQ}
\newblock
\APACjournalVolNumPages{The Annals of Applied Statistics}{10(3)}{}{1547-1571}.
\PrintBackRefs{\CurrentBib}

\bibitem [\protect \citeauthoryear {%
J.~Chen%
\ \protect \BOthers {.}}{%
J.~Chen%
\ \protect \BOthers {.}}{%
{\protect \APACyear {2012}}%
}]{%
Chen-12MCIC}
\APACinsertmetastar {%
Chen-12MCIC}%
\begin{APACrefauthors}%
Chen, J.%
, Calhiun, V\BPBI D.%
, Pearlson, G\BPBI D.%
, Ehrlich, S.%
, Turner, J\BPBI A.%
, Ho, B\BPBI C.%
\BDBL {}Liu, J.%
\end{APACrefauthors}%
\unskip\
\newblock
\APACrefYearMonthDay{2012}{}{}.
\newblock
{\BBOQ}\APACrefatitle {Multifaceted genomic risk for brain function in
  schizophrenia} {Multifaceted genomic risk for brain function in
  schizophrenia}.{\BBCQ}
\newblock
\APACjournalVolNumPages{NeuroImage}{61}{}{866-875}.
\PrintBackRefs{\CurrentBib}

\bibitem [\protect \citeauthoryear {%
Z.~Chen%
, Liu%
, Gross%
\BCBL {}\ \BBA {} Beaulieu%
}{%
Z.~Chen%
\ \protect \BOthers {.}}{%
{\protect \APACyear {2013}}%
}]{%
Chen-13}
\APACinsertmetastar {%
Chen-13}%
\begin{APACrefauthors}%
Chen, Z.%
, Liu, M.%
, Gross, D\BPBI W.%
\BCBL {}\ \BBA {} Beaulieu, C.%
\end{APACrefauthors}%
\unskip\
\newblock
\APACrefYearMonthDay{2013}{}{}.
\newblock
{\BBOQ}\APACrefatitle {Graph theoretical analysis of developmental patterns of
  the white matter network} {Graph theoretical analysis of developmental
  patterns of the white matter network}.{\BBCQ}
\newblock
\APACjournalVolNumPages{Frontiers in Human Neuroscience}{7}{}{199-211}.
\PrintBackRefs{\CurrentBib}

\bibitem [\protect \citeauthoryear {%
Ge%
\ \protect \BOthers {.}}{%
Ge%
\ \protect \BOthers {.}}{%
{\protect \APACyear {2015}}%
}]{%
Ge-15}
\APACinsertmetastar {%
Ge-15}%
\begin{APACrefauthors}%
Ge, T.%
, Nichols, T\BPBI E.%
, Ghoshd, D.%
, Morminoe, E\BPBI C.%
, J.~W.Smoller, a\BPBI M\BPBI R\BPBI S.%
\BCBL {}\ \BBA {} the Alzheimer's Disease Neuroimaging~Initiative.%
\end{APACrefauthors}%
\unskip\
\newblock
\APACrefYearMonthDay{2015}{}{}.
\newblock
{\BBOQ}\APACrefatitle {A kernel machine method for detecting effects of
  interaction between multidimensional variable sets: An imaging genetics
  application} {A kernel machine method for detecting effects of interaction
  between multidimensional variable sets: An imaging genetics
  application}.{\BBCQ}
\newblock
\APACjournalVolNumPages{NeuroImage}{109}{}{505-514}.
\PrintBackRefs{\CurrentBib}

\bibitem [\protect \citeauthoryear {%
Gerhard%
, Wagner%
, Feingold%
\BCBL {}\ \BBA {} et al.%
}{%
Gerhard%
\ \protect \BOthers {.}}{%
{\protect \APACyear {2004}}%
}]{%
Gerhard-04}
\APACinsertmetastar {%
Gerhard-04}%
\begin{APACrefauthors}%
Gerhard, D\BPBI S.%
, Wagner, L.%
, Feingold, E\BPBI A.%
\BCBL {}\ \BBA {} et al.%
\end{APACrefauthors}%
\unskip\
\newblock
\APACrefYearMonthDay{2004}{}{}.
\newblock
{\BBOQ}\APACrefatitle {The status, quality, and expansion of the NIH
  full-length cDNA project: the Mammalian Gene Collection (MGC)} {The status,
  quality, and expansion of the nih full-length cdna project: the mammalian
  gene collection (mgc)}.{\BBCQ}
\newblock
\APACjournalVolNumPages{The American Journal of Psychiatry}{14(10B)}{}{2121-7}.
\PrintBackRefs{\CurrentBib}

\bibitem [\protect \citeauthoryear {%
Gollub%
\ \protect \BOthers {.}}{%
Gollub%
\ \protect \BOthers {.}}{%
{\protect \APACyear {2013}}%
}]{%
Gollub-13}
\APACinsertmetastar {%
Gollub-13}%
\begin{APACrefauthors}%
Gollub, R\BPBI L.%
, Shoemaker, J\BPBI M.%
, King, M\BPBI D.%
, White, T.%
, Ehrlich, S.%
, Sponheim, S\BPBI R.%
\BDBL {}Andreasen, N\BPBI C.%
\end{APACrefauthors}%
\unskip\
\newblock
\APACrefYearMonthDay{2013}{}{}.
\newblock
{\BBOQ}\APACrefatitle {The MCIC collection: a shared repository of multi-modal,
  multi-site brain image data from a clinical investigation of schizophrenia}
  {The mcic collection: a shared repository of multi-modal, multi-site brain
  image data from a clinical investigation of schizophrenia}.{\BBCQ}
\newblock
\APACjournalVolNumPages{Front Genet}{11}{}{367-38}.
\PrintBackRefs{\CurrentBib}

\bibitem [\protect \citeauthoryear {%
Gretton%
\ \protect \BOthers {.}}{%
Gretton%
\ \protect \BOthers {.}}{%
{\protect \APACyear {2008}}%
}]{%
Gretton-08}
\APACinsertmetastar {%
Gretton-08}%
\begin{APACrefauthors}%
Gretton, A.%
, Fukumizu, K.%
, Teo, C\BPBI H.%
, Song, L.%
, Sch{\"{o}}lkopf, B.%
\BCBL {}\ \BBA {} Smola, A.%
\end{APACrefauthors}%
\unskip\
\newblock
\APACrefYearMonthDay{2008}{}{}.
\newblock
{\BBOQ}\APACrefatitle {A Kernel statistical test of independence} {A kernel
  statistical test of independence}.{\BBCQ}
\newblock
\APACjournalVolNumPages{In Advances in Neural Information Processing
  Systems}{20}{}{585-592}.
\PrintBackRefs{\CurrentBib}

\bibitem [\protect \citeauthoryear {%
Harrison%
\ \BBA {} Law%
}{%
Harrison%
\ \BBA {} Law%
}{%
{\protect \APACyear {206}}%
}]{%
Harrison-06}
\APACinsertmetastar {%
Harrison-06}%
\begin{APACrefauthors}%
Harrison, P\BPBI J.%
\BCBT {}\ \BBA {} Law, A\BPBI J.%
\end{APACrefauthors}%
\unskip\
\newblock
\APACrefYearMonthDay{206}{}{}.
\newblock
{\BBOQ}\APACrefatitle {Neuregulin 1 and Schizophrenia: Genetics, Gene
  Expression, and Neurobiology} {Neuregulin 1 and schizophrenia: Genetics, gene
  expression, and neurobiology}.{\BBCQ}
\newblock
\APACjournalVolNumPages{BIOL PSYCHIATRY}{60}{}{132-140}.
\PrintBackRefs{\CurrentBib}

\bibitem [\protect \citeauthoryear {%
Harville%
}{%
Harville%
}{%
{\protect \APACyear {1974}}%
}]{%
Harville-74}
\APACinsertmetastar {%
Harville-74}%
\begin{APACrefauthors}%
Harville, D\BPBI A.%
\end{APACrefauthors}%
\unskip\
\newblock
\APACrefYearMonthDay{1974}{}{}.
\newblock
{\BBOQ}\APACrefatitle {Bayesian inference for variance components using only
  error contrasts} {Bayesian inference for variance components using only error
  contrasts}.{\BBCQ}
\newblock
\APACjournalVolNumPages{Biometrika}{61(2)}{}{383-385}.
\PrintBackRefs{\CurrentBib}

\bibitem [\protect \citeauthoryear {%
Hieke%
, Binder%
, Nieters%
\BCBL {}\ \BBA {} Schumacher%
}{%
Hieke%
\ \protect \BOthers {.}}{%
{\protect \APACyear {2014}}%
}]{%
Hieke-14}
\APACinsertmetastar {%
Hieke-14}%
\begin{APACrefauthors}%
Hieke, S.%
, Binder, H.%
, Nieters, A.%
\BCBL {}\ \BBA {} Schumacher, M.%
\end{APACrefauthors}%
\unskip\
\newblock
\APACrefYearMonthDay{2014}{}{}.
\newblock
{\BBOQ}\APACrefatitle {Convergence analysis of kernel canonical correlation
  analysis: theory and practice} {Convergence analysis of kernel canonical
  correlation analysis: theory and practice}.{\BBCQ}
\newblock
\APACjournalVolNumPages{Computational Statistics}{29(1-2)}{}{51-63}.
\PrintBackRefs{\CurrentBib}

\bibitem [\protect \citeauthoryear {%
Hofmann%
, Sch{\"{o}}lkopf%
\BCBL {}\ \BBA {} Smola%
}{%
Hofmann%
\ \protect \BOthers {.}}{%
{\protect \APACyear {2008}}%
}]{%
Hofmann-08}
\APACinsertmetastar {%
Hofmann-08}%
\begin{APACrefauthors}%
Hofmann, T.%
, Sch{\"{o}}lkopf, B.%
\BCBL {}\ \BBA {} Smola, J\BPBI A.%
\end{APACrefauthors}%
\unskip\
\newblock
\APACrefYearMonthDay{2008}{}{}.
\newblock
{\BBOQ}\APACrefatitle {Kernel Methods in Machine Learning} {Kernel methods in
  machine learning}.{\BBCQ}
\newblock
\APACjournalVolNumPages{The Annals of Statistics}{36}{}{1171-1220}.
\PrintBackRefs{\CurrentBib}

\bibitem [\protect \citeauthoryear {%
Huang%
, Sherman%
\BCBL {}\ \BBA {} Lempicki%
}{%
Huang%
\ \protect \BOthers {.}}{%
{\protect \APACyear {2009}}%
}]{%
DAVID}
\APACinsertmetastar {%
DAVID}%
\begin{APACrefauthors}%
Huang, D.%
, Sherman, B\BPBI R.%
\BCBL {}\ \BBA {} Lempicki, R\BPBI A.%
\end{APACrefauthors}%
\unskip\
\newblock
\APACrefYearMonthDay{2009}{}{}.
\newblock
{\BBOQ}\APACrefatitle {Systematic and integrative analysis of large gene lists
  using DAVID Bioinformatics Resources} {Systematic and integrative analysis of
  large gene lists using david bioinformatics resources}.{\BBCQ}
\newblock
\APACjournalVolNumPages{Nature Protocols}{4(1)}{}{44-57}.
\PrintBackRefs{\CurrentBib}

\bibitem [\protect \citeauthoryear {%
I.~Ionita-Laza%
}{%
I.~Ionita-Laza%
}{%
{\protect \APACyear {2013}}%
}]{%
Ionita-13}
\APACinsertmetastar {%
Ionita-13}%
\begin{APACrefauthors}%
I.~Ionita-Laza, V\BPBI M\BPBI J\BPBI B\BPBI X\BPBI L\BPBI X\BPBI ., S.~Lee.%
\end{APACrefauthors}%
\unskip\
\newblock
\APACrefYearMonthDay{2013}{}{}.
\newblock
{\BBOQ}\APACrefatitle {Sequence kernel association tests for the combined
  effect of rare and common variants} {Sequence kernel association tests for
  the combined effect of rare and common variants}.{\BBCQ}
\newblock
\APACjournalVolNumPages{American Journal of Human Genetics}{92}{}{841-853}.
\PrintBackRefs{\CurrentBib}

\bibitem [\protect \citeauthoryear {%
Jahanshad%
\ \protect \BOthers {.}}{%
Jahanshad%
\ \protect \BOthers {.}}{%
{\protect \APACyear {2012}}%
}]{%
Jahanshad-12}
\APACinsertmetastar {%
Jahanshad-12}%
\begin{APACrefauthors}%
Jahanshad, N.%
, Hibar, D\BPBI P.%
, Ryles, A.%
, Toga, A\BPBI W.%
, McMahon, K\BPBI L.%
, de Zubicaray, G\BPBI I.%
\BDBL {}Thompson, P\BPBI M.%
\end{APACrefauthors}%
\unskip\
\newblock
\APACrefYearMonthDay{2012}{}{}.
\newblock
{\BBOQ}\APACrefatitle {DISCOVERY OF GENES THAT AFFECT HUMAN BRAIN CONNECTIVITY:
  A GENOME-WIDE ANALYSIS OF THE CONNECTOME} {Discovery of genes that affect
  human brain connectivity: A genome-wide analysis of the connectome}.{\BBCQ}
\newblock
\APACjournalVolNumPages{In Proceeding {{IEEE}} Int Symp Biomed
  Imaging}{}{}{542--545}.
\PrintBackRefs{\CurrentBib}

\bibitem [\protect \citeauthoryear {%
Jahanshad%
\ \BBA {} X.~Hua%
}{%
Jahanshad%
\ \BBA {} X.~Hua%
}{%
{\protect \APACyear {2013}}%
}]{%
Jahanshad-13}
\APACinsertmetastar {%
Jahanshad-13}%
\begin{APACrefauthors}%
Jahanshad, N.%
\BCBT {}\ \BBA {} X.~Hua, e\BPBI a.%
\end{APACrefauthors}%
\unskip\
\newblock
\APACrefYearMonthDay{2013}{}{}.
\newblock
{\BBOQ}\APACrefatitle {Genome-wide scan of healthy human connectome discovers
  SPON1 gene variant influencing dementia severity} {Genome-wide scan of
  healthy human connectome discovers spon1 gene variant influencing dementia
  severity}.{\BBCQ}
\newblock
\APACjournalVolNumPages{In Proceedings of the National Academy of
  Sciences}{110(12)}{}{4768-73}.
\PrintBackRefs{\CurrentBib}

\bibitem [\protect \citeauthoryear {%
Kimeldorf%
\ \BBA {} Wahhba%
}{%
Kimeldorf%
\ \BBA {} Wahhba%
}{%
{\protect \APACyear {1971}}%
}]{%
Kimeldorf-71}
\APACinsertmetastar {%
Kimeldorf-71}%
\begin{APACrefauthors}%
Kimeldorf, G.%
\BCBT {}\ \BBA {} Wahhba, G.%
\end{APACrefauthors}%
\unskip\
\newblock
\APACrefYearMonthDay{1971}{}{}.
\newblock
{\BBOQ}\APACrefatitle {Some Results on Tchebycheffian Spline Functions} {Some
  results on tchebycheffian spline functions}.{\BBCQ}
\newblock
\APACjournalVolNumPages{Journal of mathematical analysis and
  applications}{33(1)}{}{82-95}.
\PrintBackRefs{\CurrentBib}

\bibitem [\protect \citeauthoryear {%
Kircher%
\ \BBA {} Renate%
}{%
Kircher%
\ \BBA {} Renate%
}{%
{\protect \APACyear {2005}}%
}]{%
Kircher-05}
\APACinsertmetastar {%
Kircher-05}%
\begin{APACrefauthors}%
Kircher, T.%
\BCBT {}\ \BBA {} Renate, T.%
\end{APACrefauthors}%
\unskip\
\newblock
\APACrefYearMonthDay{2005}{}{}.
\newblock
{\BBOQ}\APACrefatitle {Functional brain imaging of symptoms and cognition in
  schizophrenia} {Functional brain imaging of symptoms and cognition in
  schizophrenia}.{\BBCQ}
\newblock
\APACjournalVolNumPages{Progress in Brain Research}{150}{}{299–308}.
\PrintBackRefs{\CurrentBib}

\bibitem [\protect \citeauthoryear {%
Koide%
, Banno%
, Aleksic%
\BCBL {}\ \BBA {} et al.%
}{%
Koide%
\ \protect \BOthers {.}}{%
{\protect \APACyear {2013}}%
}]{%
Koide-12}
\APACinsertmetastar {%
Koide-12}%
\begin{APACrefauthors}%
Koide, T.%
, Banno, M.%
, Aleksic, B.%
\BCBL {}\ \BBA {} et al.%
\end{APACrefauthors}%
\unskip\
\newblock
\APACrefYearMonthDay{2013}{}{}.
\newblock
{\BBOQ}\APACrefatitle {Common Variants in MAGI2 Gene Are Associated with
  Increased Risk for Cognitive Impairment in Schizophrenic Patients} {Common
  variants in magi2 gene are associated with increased risk for cognitive
  impairment in schizophrenic patients}.{\BBCQ}
\newblock
\APACjournalVolNumPages{PLoS ONE}{7(9)}{}{e36836}.
\PrintBackRefs{\CurrentBib}

\bibitem [\protect \citeauthoryear {%
Kung%
}{%
Kung%
}{%
{\protect \APACyear {2014}}%
}]{%
Kung-14}
\APACinsertmetastar {%
Kung-14}%
\begin{APACrefauthors}%
Kung, S\BPBI Y.%
\end{APACrefauthors}%
\unskip\
\newblock
\APACrefYearMonthDay{2014}{}{}.
\newblock
{\BBOQ}\APACrefatitle {Kernel Methods and Machine Learning} {Kernel methods and
  machine learning}.{\BBCQ}
\newblock
\APACaddressPublisher{New York}{Cambridge University Press}.
\PrintBackRefs{\CurrentBib}

\bibitem [\protect \citeauthoryear {%
Laid%
, Lange%
\BCBL {}\ \BBA {} Stram%
}{%
Laid%
\ \protect \BOthers {.}}{%
{\protect \APACyear {1987}}%
}]{%
Laird-87}
\APACinsertmetastar {%
Laird-87}%
\begin{APACrefauthors}%
Laid, N.%
, Lange, N.%
\BCBL {}\ \BBA {} Stram, D.%
\end{APACrefauthors}%
\unskip\
\newblock
\APACrefYearMonthDay{1987}{}{}.
\newblock
{\BBOQ}\APACrefatitle {Maximum Likelihood Computations with Repeated Measures:
  Application of the EM Algorithm} {Maximum likelihood computations with
  repeated measures: Application of the em algorithm}.{\BBCQ}
\newblock
\APACjournalVolNumPages{Journal of the American Statistical
  Association}{82(397)}{}{97-105}.
\PrintBackRefs{\CurrentBib}

\bibitem [\protect \citeauthoryear {%
L.~C.~Kwee%
}{%
L.~C.~Kwee%
}{%
{\protect \APACyear {2008}}%
}]{%
Kwee-08}
\APACinsertmetastar {%
Kwee-08}%
\begin{APACrefauthors}%
L.~C.~Kwee, X\BPBI L\BPBI D\BPBI G\BPBI M\BPBI P\BPBI E., D.~Liu.%
\end{APACrefauthors}%
\unskip\
\newblock
\APACrefYearMonthDay{2008}{}{}.
\newblock
{\BBOQ}\APACrefatitle {A powerful and flexible multilocus association test for
  quantitative traits} {A powerful and flexible multilocus association test for
  quantitative traits}.{\BBCQ}
\newblock
\APACjournalVolNumPages{Annals of Human Genetics}{82(2)}{}{386-397}.
\PrintBackRefs{\CurrentBib}

\bibitem [\protect \citeauthoryear {%
Lencz%
\ \protect \BOthers {.}}{%
Lencz%
\ \protect \BOthers {.}}{%
{\protect \APACyear {2007}}%
}]{%
Lencz-07}
\APACinsertmetastar {%
Lencz-07}%
\begin{APACrefauthors}%
Lencz, T.%
, Morgan, T\BPBI V.%
, Athanasiou, M.%
, Dain, B.%
, Reed, C\BPBI R.%
, Kane, J\BPBI M.%
\BDBL {}Malhotra, A\BPBI K.%
\end{APACrefauthors}%
\unskip\
\newblock
\APACrefYearMonthDay{2007}{}{}.
\newblock
{\BBOQ}\APACrefatitle {Converging evidence for a pseudoautosomal cytokine
  receptor gene locus in schizophrenia} {Converging evidence for a
  pseudoautosomal cytokine receptor gene locus in schizophrenia}.{\BBCQ}
\newblock
\APACjournalVolNumPages{Molecular Psychiatry}{12}{}{572-580}.
\PrintBackRefs{\CurrentBib}

\bibitem [\protect \citeauthoryear {%
J.~Li%
\ \protect \BOthers {.}}{%
J.~Li%
\ \protect \BOthers {.}}{%
{\protect \APACyear {2015}}%
}]{%
Li-15}
\APACinsertmetastar {%
Li-15}%
\begin{APACrefauthors}%
Li, J.%
, Huang, D.%
, Guo, M.%
, Liu, X.%
, Wang, C.%
, Teng, Z.%
\BDBL {}Wang, L.%
\end{APACrefauthors}%
\unskip\
\newblock
\APACrefYearMonthDay{2015}{}{}.
\newblock
{\BBOQ}\APACrefatitle {A gene-based information gain method for detecting
  gene–gene interactions in case–control studies} {A gene-based information
  gain method for detecting gene–gene interactions in case–control
  studies}.{\BBCQ}
\newblock
\APACjournalVolNumPages{European Journal of Human Genetics}{23}{}{1566-1572}.
\PrintBackRefs{\CurrentBib}

\bibitem [\protect \citeauthoryear {%
S.~Li%
\ \BBA {} Cui%
}{%
S.~Li%
\ \BBA {} Cui%
}{%
{\protect \APACyear {2012}}%
}]{%
Li-12}
\APACinsertmetastar {%
Li-12}%
\begin{APACrefauthors}%
Li, S.%
\BCBT {}\ \BBA {} Cui, Y.%
\end{APACrefauthors}%
\unskip\
\newblock
\APACrefYearMonthDay{2012}{}{}.
\newblock
{\BBOQ}\APACrefatitle {Gene-Centric gene-gene interaction: a model-based kernel
  machine method} {Gene-centric gene-gene interaction: a model-based kernel
  machine method}.{\BBCQ}
\newblock
\APACjournalVolNumPages{The Annals of Applied Statistics}{6(3)}{}{1134-1161}.
\PrintBackRefs{\CurrentBib}

\bibitem [\protect \citeauthoryear {%
Lin%
, Callhoun%
\BCBL {}\ \BBA {} Wang%
}{%
Lin%
\ \protect \BOthers {.}}{%
{\protect \APACyear {2014}}%
}]{%
Dongdong-14}
\APACinsertmetastar {%
Dongdong-14}%
\begin{APACrefauthors}%
Lin, D.%
, Callhoun, V\BPBI D.%
\BCBL {}\ \BBA {} Wang, Y\BPBI P.%
\end{APACrefauthors}%
\unskip\
\newblock
\APACrefYearMonthDay{2014}{}{}.
\newblock
{\BBOQ}\APACrefatitle {Correspondence between f{\mbox{MRI}} and {\mbox{SNP}}
  data by group sparse canonical correlation analysis} {Correspondence between
  f{\mbox{mri}} and {\mbox{snp}} data by group sparse canonical correlation
  analysis}.{\BBCQ}
\newblock
\APACjournalVolNumPages{Medical Image Analysis}{18}{}{891 - 902}.
\PrintBackRefs{\CurrentBib}

\bibitem [\protect \citeauthoryear {%
Lindstrom%
\ \BBA {} Bates%
}{%
Lindstrom%
\ \BBA {} Bates%
}{%
{\protect \APACyear {1988}}%
}]{%
Lindstrom-88}
\APACinsertmetastar {%
Lindstrom-88}%
\begin{APACrefauthors}%
Lindstrom, M\BPBI J.%
\BCBT {}\ \BBA {} Bates, M\BPBI D.%
\end{APACrefauthors}%
\unskip\
\newblock
\APACrefYearMonthDay{1988}{}{}.
\newblock
{\BBOQ}\APACrefatitle {Newton-Raphson and EM algorithms for linear
  mixed-effects models for repeated-measures data} {Newton-raphson and em
  algorithms for linear mixed-effects models for repeated-measures
  data}.{\BBCQ}
\newblock
\APACjournalVolNumPages{Journal of the American Statistical
  Association}{83(404)}{}{1014-1022}.
\PrintBackRefs{\CurrentBib}

\bibitem [\protect \citeauthoryear {%
D.~Liu%
, Lin%
\BCBL {}\ \BBA {} Ghosh%
}{%
D.~Liu%
\ \protect \BOthers {.}}{%
{\protect \APACyear {2007}}%
}]{%
Liu-07}
\APACinsertmetastar {%
Liu-07}%
\begin{APACrefauthors}%
Liu, D.%
, Lin, X.%
\BCBL {}\ \BBA {} Ghosh, D.%
\end{APACrefauthors}%
\unskip\
\newblock
\APACrefYearMonthDay{2007}{}{}.
\newblock
{\BBOQ}\APACrefatitle {Semiparametric regression of multidimensional genetics
  pathway data: least squares kernel machines and linear mixed model,}
  {Semiparametric regression of multidimensional genetics pathway data: least
  squares kernel machines and linear mixed model,}.{\BBCQ}
\newblock
\APACjournalVolNumPages{Biometrics}{630(4)}{}{1079-1088}.
\PrintBackRefs{\CurrentBib}

\bibitem [\protect \citeauthoryear {%
J.~Liu%
\ \protect \BOthers {.}}{%
J.~Liu%
\ \protect \BOthers {.}}{%
{\protect \APACyear {2014}}%
}]{%
Liu-13}
\APACinsertmetastar {%
Liu-13}%
\begin{APACrefauthors}%
Liu, J.%
, Chen, J.%
, Ehrlich, S.%
, Walton, E.%
, T.~White, N\BPBI P\BPBI B.%
, Bustillo, J.%
\BDBL {}Calhoun, V\BPBI D.%
\end{APACrefauthors}%
\unskip\
\newblock
\APACrefYearMonthDay{2014}{}{}.
\newblock
{\BBOQ}\APACrefatitle {Methylation Patterns in Whole Blood Correlate With
  Symptoms in Schizophrenia Patients} {Methylation patterns in whole blood
  correlate with symptoms in schizophrenia patients}.{\BBCQ}
\newblock
\APACjournalVolNumPages{Schizophrenia Bulletin}{40(4)}{}{769-776}.
\PrintBackRefs{\CurrentBib}

\bibitem [\protect \citeauthoryear {%
M.~Liu%
, Min%
, Y.~Gao%
\BCBL {}\ \BBA {} Shen%
}{%
M.~Liu%
\ \protect \BOthers {.}}{%
{\protect \APACyear {2016}}%
}]{%
Liu-16}
\APACinsertmetastar {%
Liu-16}%
\begin{APACrefauthors}%
Liu, M.%
, Min, R.%
, Y.~Gao, D\BPBI Z.%
\BCBL {}\ \BBA {} Shen, D.%
\end{APACrefauthors}%
\unskip\
\newblock
\APACrefYearMonthDay{2016}{}{}.
\newblock
{\BBOQ}\APACrefatitle {Multitemplate-based multiview learning for Alzheimer's
  disease diagnosis Machine Learning and Medical Imaging} {Multitemplate-based
  multiview learning for alzheimer's disease diagnosis machine learning and
  medical imaging}.{\BBCQ}
\newblock
\APACjournalVolNumPages{Machine Learning and Medical Imaging}{}{}{259-297}.
\PrintBackRefs{\CurrentBib}

\bibitem [\protect \citeauthoryear {%
Montano%
, Tauband%
, Jaffe%
, Briem%
\BCBL {}\ \BBA {} et al.%
}{%
Montano%
\ \protect \BOthers {.}}{%
{\protect \APACyear {2016}}%
}]{%
Montano-16}
\APACinsertmetastar {%
Montano-16}%
\begin{APACrefauthors}%
Montano, C.%
, Tauband, M\BPBI A.%
, Jaffe, A.%
, Briem, E.%
\BCBL {}\ \BBA {} et al.%
\end{APACrefauthors}%
\unskip\
\newblock
\APACrefYearMonthDay{2016}{}{}.
\newblock
{\BBOQ}\APACrefatitle {Association of DNA Methylation Differences With
  Schizophrenia in an Epigenome-Wide Association Study} {Association of dna
  methylation differences with schizophrenia in an epigenome-wide association
  study}.{\BBCQ}
\newblock
\APACjournalVolNumPages{JAMA Psychiatry}{73(5)}{}{506-514}.
\PrintBackRefs{\CurrentBib}

\bibitem [\protect \citeauthoryear {%
Moselhy%
, Eapenb%
, Akawi%
, Younis%
\BCBL {}\ \BBA {} et. al.%
}{%
Moselhy%
\ \protect \BOthers {.}}{%
{\protect \APACyear {2015}}%
}]{%
Moselhy-15}
\APACinsertmetastar {%
Moselhy-15}%
\begin{APACrefauthors}%
Moselhy, H.%
, Eapenb, V.%
, Akawi, N\BPBI A.%
, Younis, A.%
\BCBL {}\ \BBA {} et. al.%
\end{APACrefauthors}%
\unskip\
\newblock
\APACrefYearMonthDay{2015}{}{}.
\newblock
{\BBOQ}\APACrefatitle {Secondary association of PDLIM5 with paranoid
  schizophrenia in Emirati patients} {Secondary association of pdlim5 with
  paranoid schizophrenia in emirati patients}.{\BBCQ}
\newblock
\APACjournalVolNumPages{Meta Gene}{5}{}{135-139}.
\PrintBackRefs{\CurrentBib}

\bibitem [\protect \citeauthoryear {%
Parkhomenko%
, Tritchler%
\BCBL {}\ \BBA {} Beyene%
}{%
Parkhomenko%
\ \protect \BOthers {.}}{%
{\protect \APACyear {2009}}%
}]{%
Parkhomenko-09}
\APACinsertmetastar {%
Parkhomenko-09}%
\begin{APACrefauthors}%
Parkhomenko, E.%
, Tritchler, D.%
\BCBL {}\ \BBA {} Beyene, J.%
\end{APACrefauthors}%
\unskip\
\newblock
\APACrefYearMonthDay{2009}{}{}.
\newblock
{\BBOQ}\APACrefatitle {Sparse canonical correlation analysis with application
  to genomic data integration} {Sparse canonical correlation analysis with
  application to genomic data integration}.{\BBCQ}
\newblock
\APACjournalVolNumPages{Statistical Applications in Genetics and Molecular
  Biolog}{8(1)}{}{1-34}.
\PrintBackRefs{\CurrentBib}

\bibitem [\protect \citeauthoryear {%
Pearlson%
, Liu%
\BCBL {}\ \BBA {} Calhoun%
}{%
Pearlson%
\ \protect \BOthers {.}}{%
{\protect \APACyear {2015}}%
}]{%
Pearlson-15}
\APACinsertmetastar {%
Pearlson-15}%
\begin{APACrefauthors}%
Pearlson, G\BPBI D.%
, Liu, J.%
\BCBL {}\ \BBA {} Calhoun, V\BPBI D.%
\end{APACrefauthors}%
\unskip\
\newblock
\APACrefYearMonthDay{2015}{}{}.
\newblock
{\BBOQ}\APACrefatitle {An introductory review of parallel independent component
  analysis (p-ICA) and a guide to applying p-ICA to genetic data and imaging
  phenotypes to identify disease-associated biological pathways and systems in
  common complex disorders} {An introductory review of parallel independent
  component analysis (p-ica) and a guide to applying p-ica to genetic data and
  imaging phenotypes to identify disease-associated biological pathways and
  systems in common complex disorders}.{\BBCQ}
\newblock
\APACjournalVolNumPages{Front Genet}{6}{}{276}.
\PrintBackRefs{\CurrentBib}

\bibitem [\protect \citeauthoryear {%
Peng%
, Zhao%
\BCBL {}\ \BBA {} Xue%
}{%
Peng%
\ \protect \BOthers {.}}{%
{\protect \APACyear {2010}}%
}]{%
Peng-10}
\APACinsertmetastar {%
Peng-10}%
\begin{APACrefauthors}%
Peng, Q\BPBI N.%
, Zhao, J.%
\BCBL {}\ \BBA {} Xue, F.%
\end{APACrefauthors}%
\unskip\
\newblock
\APACrefYearMonthDay{2010}{}{}.
\newblock
{\BBOQ}\APACrefatitle {A gene-based method for detecting gene–gene
  co-association in a case–control association study} {A gene-based method for
  detecting gene–gene co-association in a case–control association
  study}.{\BBCQ}
\newblock
\APACjournalVolNumPages{European Journal of Human Genetics}{18}{}{582-587}.
\PrintBackRefs{\CurrentBib}

\bibitem [\protect \citeauthoryear {%
Potkin%
, T.~G. M.~Van%
, Ling%
, Macciardi%
\BCBL {}\ \BBA {} Xie%
}{%
Potkin%
\ \protect \BOthers {.}}{%
{\protect \APACyear {2015}}%
}]{%
Potkin-15}
\APACinsertmetastar {%
Potkin-15}%
\begin{APACrefauthors}%
Potkin, S\BPBI G.%
, T.~G. M.~Van, E.%
, Ling, S.%
, Macciardi, F.%
\BCBL {}\ \BBA {} Xie, X.%
\end{APACrefauthors}%
\unskip\
\newblock
\APACrefYearMonthDay{2015}{}{}.
\newblock
{\BBOQ}\APACrefatitle {Unanticipated Genes and Mechanisms in Serious Mental
  Illness: GWAS based Imaging Genetics Strategies} {Unanticipated genes and
  mechanisms in serious mental illness: Gwas based imaging genetics
  strategies}.{\BBCQ}
\newblock
\BIn{} (\BVOL~209).
\newblock
\APACaddressPublisher{London}{Oxford University Press}.
\PrintBackRefs{\CurrentBib}

\bibitem [\protect \citeauthoryear {%
Richfield%
, Alam%
, Calhoun%
\BCBL {}\ \BBA {} Wang%
}{%
Richfield%
\ \protect \BOthers {.}}{%
{\protect \APACyear {2017}}%
}]{%
Richfield-17}
\APACinsertmetastar {%
Richfield-17}%
\begin{APACrefauthors}%
Richfield, O.%
, Alam, M\BPBI A.%
, Calhoun, V.%
\BCBL {}\ \BBA {} Wang, Y\BPBI P.%
\end{APACrefauthors}%
\unskip\
\newblock
\APACrefYearMonthDay{2017}{}{}.
\newblock
{\BBOQ}\APACrefatitle {Learning Schizophrenia Imaging Genetics Data Via
  Multiple Kernel Canonical Correlation Analysis} {Learning schizophrenia
  imaging genetics data via multiple kernel canonical correlation
  analysis}.{\BBCQ}
\newblock
\APACjournalVolNumPages{Proceedings - 2016 IEEE International Conference on
  Bioinformatics and Biomedicine, BIBM 2016, Shenzhen, China}{5}{}{507-5011}.
\PrintBackRefs{\CurrentBib}

\bibitem [\protect \citeauthoryear {%
Sanders%
, Duan%
, Levinson%
\BCBL {}\ \BBA {} et. al.%
}{%
Sanders%
\ \protect \BOthers {.}}{%
{\protect \APACyear {2008}}%
}]{%
Sanders-08}
\APACinsertmetastar {%
Sanders-08}%
\begin{APACrefauthors}%
Sanders, A\BPBI R.%
, Duan, J.%
, Levinson, D\BPBI F.%
\BCBL {}\ \BBA {} et. al.%
\end{APACrefauthors}%
\unskip\
\newblock
\APACrefYearMonthDay{2008}{}{}.
\newblock
{\BBOQ}\APACrefatitle {No significant association of 14 candidate genes with
  schizophrenia in a large European ancestry sample: implications for
  psychiatric genetics} {No significant association of 14 candidate genes with
  schizophrenia in a large european ancestry sample: implications for
  psychiatric genetics}.{\BBCQ}
\newblock
\APACjournalVolNumPages{The American Journal of
  Psychiatry}{165(10)}{}{1359-1368}.
\PrintBackRefs{\CurrentBib}

\bibitem [\protect \citeauthoryear {%
Satterthwaite%
}{%
Satterthwaite%
}{%
{\protect \APACyear {1946}}%
}]{%
Satterthwaite-46}
\APACinsertmetastar {%
Satterthwaite-46}%
\begin{APACrefauthors}%
Satterthwaite, F\BPBI E.%
\end{APACrefauthors}%
\unskip\
\newblock
\APACrefYearMonthDay{1946}{}{}.
\newblock
{\BBOQ}\APACrefatitle {An Approximate Distribution of Estimates of Variance
  Components} {An approximate distribution of estimates of variance
  components}.{\BBCQ}
\newblock
\APACjournalVolNumPages{Biometrics Bulletin}{2(6)}{}{110-114}.
\PrintBackRefs{\CurrentBib}

\bibitem [\protect \citeauthoryear {%
Sch{\"{o}}lkopf%
\ \BBA {} Smola%
}{%
Sch{\"{o}}lkopf%
\ \BBA {} Smola%
}{%
{\protect \APACyear {2002}}%
}]{%
Schlkof-book}
\APACinsertmetastar {%
Schlkof-book}%
\begin{APACrefauthors}%
Sch{\"{o}}lkopf, B.%
\BCBT {}\ \BBA {} Smola, A\BPBI J.%
\end{APACrefauthors}%
\unskip\
\newblock
\APACrefYear{2002}.
\newblock
\APACrefbtitle {Learning with Kernels} {Learning with kernels}.
\newblock
\APACaddressPublisher{Cambridge MA}{MIT Press}.
\PrintBackRefs{\CurrentBib}

\bibitem [\protect \citeauthoryear {%
Sch{\"{o}}lkopf%
, Smola%
\BCBL {}\ \BBA {} M{\"{u}}ller%
}{%
Sch{\"{o}}lkopf%
\ \protect \BOthers {.}}{%
{\protect \APACyear {1998}}%
}]{%
Schlkof-kpca}
\APACinsertmetastar {%
Schlkof-kpca}%
\begin{APACrefauthors}%
Sch{\"{o}}lkopf, B.%
, Smola, A\BPBI J.%
\BCBL {}\ \BBA {} M{\"{u}}ller, K\BHBI R.%
\end{APACrefauthors}%
\unskip\
\newblock
\APACrefYearMonthDay{1998}{}{}.
\newblock
{\BBOQ}\APACrefatitle {Nonlinear component analysis as a kernel eigenvalue
  problem} {Nonlinear component analysis as a kernel eigenvalue
  problem}.{\BBCQ}
\newblock
\APACjournalVolNumPages{Neural Computation.}{10}{}{1299-1319}.
\PrintBackRefs{\CurrentBib}

\bibitem [\protect \citeauthoryear {%
Shibuya%
\ \protect \BOthers {.}}{%
Shibuya%
\ \protect \BOthers {.}}{%
{\protect \APACyear {2013}}%
}]{%
Shibuya-13}
\APACinsertmetastar {%
Shibuya-13}%
\begin{APACrefauthors}%
Shibuya, M.%
, Watanabe, Y.%
, Nunokawa, A.%
, Egawa, J.%
, Kaneko, N.%
, Igeta, H.%
\BCBL {}\ \BBA {} Someya, T.%
\end{APACrefauthors}%
\unskip\
\newblock
\APACrefYearMonthDay{2013}{}{}.
\newblock
{\BBOQ}\APACrefatitle {Interleukin 1 beta gene and risk of schizophrenia:
  detailed case–control and family-based studies and an updated meta-analysis}
  {Interleukin 1 beta gene and risk of schizophrenia: detailed case–control and
  family-based studies and an updated meta-analysis}.{\BBCQ}
\newblock
\APACjournalVolNumPages{Human Psychopharmacology}{29}{}{31-37}.
\PrintBackRefs{\CurrentBib}

\bibitem [\protect \citeauthoryear {%
Siawa%
, Liuc%
, Linc%
, Beend%
\BCBL {}\ \BBA {} Hsiehc%
}{%
Siawa%
\ \protect \BOthers {.}}{%
{\protect \APACyear {2016}}%
}]{%
Siaw-16}
\APACinsertmetastar {%
Siaw-16}%
\begin{APACrefauthors}%
Siawa, G\BPBI E\BHBI L.%
, Liuc, I\BHBI F.%
, Linc, P\BPBI Y.%
, Beend, M\BPBI D.%
\BCBL {}\ \BBA {} Hsiehc, T.%
\end{APACrefauthors}%
\unskip\
\newblock
\APACrefYearMonthDay{2016}{}{}.
\newblock
{\BBOQ}\APACrefatitle {DNA and RNA topoisomerase activities of Top3â are
  promoted by mediator protein Tudor domain-containing protein 3} {Dna and rna
  topoisomerase activities of top3â are promoted by mediator protein tudor
  domain-containing protein 3}.{\BBCQ}
\newblock
\APACjournalVolNumPages{Proc Natl Acad Sci USA}{113}{}{5544-5551}.
\PrintBackRefs{\CurrentBib}

\bibitem [\protect \citeauthoryear {%
Sluis%
\ \protect \BOthers {.}}{%
Sluis%
\ \protect \BOthers {.}}{%
{\protect \APACyear {2015}}%
}]{%
Sluis-15}
\APACinsertmetastar {%
Sluis-15}%
\begin{APACrefauthors}%
Sluis, S\BPBI V\BPBI D.%
, Dolan, C\BPBI V.%
, Li, J.%
, Song, Y.%
, Sham, P.%
, Posthuma1, D.%
\BCBL {}\ \BBA {} Li, M.%
\end{APACrefauthors}%
\unskip\
\newblock
\APACrefYearMonthDay{2015}{}{}.
\newblock
{\BBOQ}\APACrefatitle {MGAS: a powerful tool for multivariate gene-based
  genome-wide association analysis} {Mgas: a powerful tool for multivariate
  gene-based genome-wide association analysis}.{\BBCQ}
\newblock
\APACjournalVolNumPages{Bioinformatics}{31}{}{1007-1015}.
\PrintBackRefs{\CurrentBib}

\bibitem [\protect \citeauthoryear {%
Song%
, Smola%
, Gretton%
, Bedo%
\BCBL {}\ \BBA {} Borgwardt%
}{%
Song%
\ \protect \BOthers {.}}{%
{\protect \APACyear {2012}}%
}]{%
Song-12}
\APACinsertmetastar {%
Song-12}%
\begin{APACrefauthors}%
Song, L.%
, Smola, A.%
, Gretton, A.%
, Bedo, J.%
\BCBL {}\ \BBA {} Borgwardt, K.%
\end{APACrefauthors}%
\unskip\
\newblock
\APACrefYearMonthDay{2012}{}{}.
\newblock
{\BBOQ}\APACrefatitle {Feature Selection via Dependence Maximization} {Feature
  selection via dependence maximization}.{\BBCQ}
\newblock
\APACjournalVolNumPages{Journal of Machine Learning
  Research}{13}{}{1393--1434}.
\PrintBackRefs{\CurrentBib}

\bibitem [\protect \citeauthoryear {%
Strausberg%
, Feingold%
, Grouse%
\BCBL {}\ \BBA {} et al.%
}{%
Strausberg%
\ \protect \BOthers {.}}{%
{\protect \APACyear {2002}}%
}]{%
Strausberg-02}
\APACinsertmetastar {%
Strausberg-02}%
\begin{APACrefauthors}%
Strausberg, R\BPBI L.%
, Feingold, E\BPBI A.%
, Grouse, L\BPBI H.%
\BCBL {}\ \BBA {} et al.%
\end{APACrefauthors}%
\unskip\
\newblock
\APACrefYearMonthDay{2002}{}{}.
\newblock
{\BBOQ}\APACrefatitle {Generation and initial analysis of more than 15,000
  full-length human and mouse cDNA sequences} {Generation and initial analysis
  of more than 15,000 full-length human and mouse cdna sequences}.{\BBCQ}
\newblock
\APACjournalVolNumPages{Proceedings of the National Academy of Sciences,
  USA}{99(26)}{}{16899-903}.
\PrintBackRefs{\CurrentBib}

\bibitem [\protect \citeauthoryear {%
Suk%
, Wee%
, Lee%
\BCBL {}\ \BBA {} Shen%
}{%
Suk%
\ \protect \BOthers {.}}{%
{\protect \APACyear {2016}}%
}]{%
Suk-16}
\APACinsertmetastar {%
Suk-16}%
\begin{APACrefauthors}%
Suk, H.%
, Wee, C.%
, Lee, S.%
\BCBL {}\ \BBA {} Shen, D.%
\end{APACrefauthors}%
\unskip\
\newblock
\APACrefYearMonthDay{2016}{}{}.
\newblock
{\BBOQ}\APACrefatitle {State-spacemodel with deep learning for functional
  dynamics estimation in resting-state fMRI} {State-spacemodel with deep
  learning for functional dynamics estimation in resting-state fmri}.{\BBCQ}
\newblock
\APACjournalVolNumPages{NeuroImage}{129}{}{292-307}.
\PrintBackRefs{\CurrentBib}

\bibitem [\protect \citeauthoryear {%
S.~Yu%
\ \BBA {} Moreau%
}{%
S.~Yu%
\ \BBA {} Moreau%
}{%
{\protect \APACyear {2011}}%
}]{%
Yu-11}
\APACinsertmetastar {%
Yu-11}%
\begin{APACrefauthors}%
S.~Yu, B\BPBI D\BPBI M., L-C.~Tranchevent%
\BCBT {}\ \BBA {} Moreau, Y.%
\end{APACrefauthors}%
\unskip\
\newblock
\APACrefYearMonthDay{2011}{}{}.
\newblock
{\BBOQ}\APACrefatitle {Kernel-based Data Fusion for Machine Learning}
  {Kernel-based data fusion for machine learning}.{\BBCQ}
\newblock
\APACaddressPublisher{Verlag Berlin Heidelberg}{Springer}.
\PrintBackRefs{\CurrentBib}

\bibitem [\protect \citeauthoryear {%
Szklarczyk%
\ \protect \BOthers {.}}{%
Szklarczyk%
\ \protect \BOthers {.}}{%
{\protect \APACyear {2007}}%
}]{%
STRING-15}
\APACinsertmetastar {%
STRING-15}%
\begin{APACrefauthors}%
Szklarczyk, D.%
, Franceschini, A.%
, Wyder, S.%
, Forslund, K.%
, Heller, D.%
, Huerta-Cepas, J.%
\BDBL {}von Mering, C.%
\end{APACrefauthors}%
\unskip\
\newblock
\APACrefYearMonthDay{2007}{}{}.
\newblock
{\BBOQ}\APACrefatitle {{{STRING}} v10: Protein–protein Interaction Networks,
  Integrated over the Tree of Life} {{{STRING}} v10: Protein–protein
  interaction networks, integrated over the tree of life}.{\BBCQ}
\newblock
\APACjournalVolNumPages{Nucleic Acids Research}{43}{}{531--543}.
\PrintBackRefs{\CurrentBib}

\bibitem [\protect \citeauthoryear {%
Van%
\ \BBA {} Kapur%
}{%
Van%
\ \BBA {} Kapur%
}{%
{\protect \APACyear {2009}}%
}]{%
Van-09}
\APACinsertmetastar {%
Van-09}%
\begin{APACrefauthors}%
Van, S\BPBI O.%
\BCBT {}\ \BBA {} Kapur, S.%
\end{APACrefauthors}%
\unskip\
\newblock
\APACrefYearMonthDay{2009}{}{}.
\newblock
{\BBOQ}\APACrefatitle {Schizophrenia} {Schizophrenia}.{\BBCQ}
\newblock
\APACjournalVolNumPages{Lancet}{374 (9690)}{}{635–645}.
\PrintBackRefs{\CurrentBib}

\bibitem [\protect \citeauthoryear {%
Wan%
\ \protect \BOthers {.}}{%
Wan%
\ \protect \BOthers {.}}{%
{\protect \APACyear {2010}}%
}]{%
Wan-10}
\APACinsertmetastar {%
Wan-10}%
\begin{APACrefauthors}%
Wan, X.%
, Yang, C.%
, Yang, Q.%
, Xue, H.%
, Fan, X.%
, Tang, N\BPBI L.%
\BCBL {}\ \BBA {} Yu, W.%
\end{APACrefauthors}%
\unskip\
\newblock
\APACrefYearMonthDay{2010}{}{}.
\newblock
{\BBOQ}\APACrefatitle {BOOST: A Fast Approach to Detecting Gene-Gene
  Interactions in Genome-wide Case-Control Studies} {Boost: A fast approach to
  detecting gene-gene interactions in genome-wide case-control studies}.{\BBCQ}
\newblock
\APACjournalVolNumPages{The American Journal of Human Genetics}{87}{}{325-340}.
\PrintBackRefs{\CurrentBib}

\bibitem [\protect \citeauthoryear {%
Wen%
\ \protect \BOthers {.}}{%
Wen%
\ \protect \BOthers {.}}{%
{\protect \APACyear {2017}}%
}]{%
Wen-17}
\APACinsertmetastar {%
Wen-17}%
\begin{APACrefauthors}%
Wen, H.%
, Liu, Y.%
, Rekik, I.%
, Wang, S.%
, Chen, Z.%
, Zhang, J.%
\BDBL {}He, H.%
\end{APACrefauthors}%
\unskip\
\newblock
\APACrefYearMonthDay{2017}{}{}.
\newblock
{\BBOQ}\APACrefatitle {Multi-modal multiple kernel learning for accurate
  identification of Tourette Syndrome children} {Multi-modal multiple kernel
  learning for accurate identification of tourette syndrome children}.{\BBCQ}
\newblock
\APACjournalVolNumPages{Pattern Recognition}{63}{}{601-611}.
\PrintBackRefs{\CurrentBib}

\bibitem [\protect \citeauthoryear {%
Wockner%
\ \protect \BOthers {.}}{%
Wockner%
\ \protect \BOthers {.}}{%
{\protect \APACyear {2014}}%
}]{%
Wockner-14}
\APACinsertmetastar {%
Wockner-14}%
\begin{APACrefauthors}%
Wockner, L\BPBI F.%
, Noble, E\BPBI P.%
, Lawford, B\BPBI R.%
, Young, R\BPBI M.%
, Morris, C\BPBI P.%
, Whitehall, V\BPBI L\BPBI J.%
\BCBL {}\ \BBA {} Voisey, J.%
\end{APACrefauthors}%
\unskip\
\newblock
\APACrefYearMonthDay{2014}{}{}.
\newblock
{\BBOQ}\APACrefatitle {Genome-wide DNA methylation analysis of human brain
  tissue from schizophrenia patients} {Genome-wide dna methylation analysis of
  human brain tissue from schizophrenia patients}.{\BBCQ}
\newblock
\APACjournalVolNumPages{Transl Psychiatry}{4 (e339)}{}{1-8}.
\PrintBackRefs{\CurrentBib}

\bibitem [\protect \citeauthoryear {%
K.~Wu%
\ \protect \BOthers {.}}{%
K.~Wu%
\ \protect \BOthers {.}}{%
{\protect \APACyear {2013}}%
}]{%
Wu-13}
\APACinsertmetastar {%
Wu-13}%
\begin{APACrefauthors}%
Wu, K.%
, Taki, Y.%
, Sato, K.%
, Oi, H.%
, Kawashima, R.%
\BCBL {}\ \BBA {} Fukuda, H.%
\end{APACrefauthors}%
\unskip\
\newblock
\APACrefYearMonthDay{2013}{}{}.
\newblock
{\BBOQ}\APACrefatitle {A longitudinal study of structural brain network changes
  with normal aging} {A longitudinal study of structural brain network changes
  with normal aging}.{\BBCQ}
\newblock
\APACjournalVolNumPages{Frontiers in Human Neuroscience}{7}{}{225-236}.
\PrintBackRefs{\CurrentBib}

\bibitem [\protect \citeauthoryear {%
M\BPBI C.~Wu%
\ \protect \BOthers {.}}{%
M\BPBI C.~Wu%
\ \protect \BOthers {.}}{%
{\protect \APACyear {2011}}%
}]{%
Wu-11}
\APACinsertmetastar {%
Wu-11}%
\begin{APACrefauthors}%
Wu, M\BPBI C.%
, Lee, S.%
, Cai, T.%
, Li, Y.%
, Boehnke, M.%
\BCBL {}\ \BBA {} Lin, X.%
\end{APACrefauthors}%
\unskip\
\newblock
\APACrefYearMonthDay{2011}{}{}.
\newblock
{\BBOQ}\APACrefatitle {Rare Variant Association Testing for Sequencing Data
  Using the Sequence Kernel Association Test ({{SKAT}})} {Rare variant
  association testing for sequencing data using the sequence kernel association
  test ({{SKAT}})}.{\BBCQ}
\newblock
\APACjournalVolNumPages{American Journal of Human Genetics}{89}{}{82-93}.
\PrintBackRefs{\CurrentBib}

\bibitem [\protect \citeauthoryear {%
Xu%
, Tao%
\BCBL {}\ \BBA {} Xu%
}{%
Xu%
\ \protect \BOthers {.}}{%
{\protect \APACyear {2013}}%
}]{%
Xu-13}
\APACinsertmetastar {%
Xu-13}%
\begin{APACrefauthors}%
Xu, C.%
, Tao, D.%
\BCBL {}\ \BBA {} Xu, C.%
\end{APACrefauthors}%
\unskip\
\newblock
\APACrefYearMonthDay{2013}{}{}.
\newblock
{\BBOQ}\APACrefatitle {A survey of multi-view machine learning} {A survey of
  multi-view machine learning}.{\BBCQ}
\newblock
\APACjournalVolNumPages{Neural Computation and
  Applications}{23(7-8)}{}{2031-2038}.
\PrintBackRefs{\CurrentBib}

\bibitem [\protect \citeauthoryear {%
Yan%
\ \BBA {} Zang%
}{%
Yan%
\ \BBA {} Zang%
}{%
{\protect \APACyear {2010}}%
}]{%
Yan-10}
\APACinsertmetastar {%
Yan-10}%
\begin{APACrefauthors}%
Yan, C.%
\BCBT {}\ \BBA {} Zang, Y.%
\end{APACrefauthors}%
\unskip\
\newblock
\APACrefYearMonthDay{2010}{}{}.
\newblock
{\BBOQ}\APACrefatitle {{\mbox{DPARSF}}: a {{MATLAB}} toolbox for pipeline data
  analysis of resting-state f{{MRI}}} {{\mbox{DPARSF}}: a {{MATLAB}} toolbox
  for pipeline data analysis of resting-state f{{MRI}}}.{\BBCQ}
\newblock
\APACjournalVolNumPages{Frontiers in Systems Neuroscience}{4 (13)}{}{1-7}.
\PrintBackRefs{\CurrentBib}

\bibitem [\protect \citeauthoryear {%
Yuan%
\ \protect \BOthers {.}}{%
Yuan%
\ \protect \BOthers {.}}{%
{\protect \APACyear {2012}}%
}]{%
Zhongshang-12}
\APACinsertmetastar {%
Zhongshang-12}%
\begin{APACrefauthors}%
Yuan, Z.%
, Gao, Q.%
, He, Y.%
, Zhang, X.%
, Li, F.%
, Zhao, J.%
\BCBL {}\ \BBA {} Xue, F.%
\end{APACrefauthors}%
\unskip\
\newblock
\APACrefYearMonthDay{2012}{}{}.
\newblock
{\BBOQ}\APACrefatitle {Detection for gene-gene co-association via kernel
  canonical correlation analysis} {Detection for gene-gene co-association via
  kernel canonical correlation analysis}.{\BBCQ}
\newblock
\APACjournalVolNumPages{BMC Genetic}{13:83}{}{}.
\PrintBackRefs{\CurrentBib}

\bibitem [\protect \citeauthoryear {%
Zhao%
, Qiao%
, F.~Shi%
\BCBL {}\ \BBA {} Shen%
}{%
Zhao%
\ \protect \BOthers {.}}{%
{\protect \APACyear {2016}}%
}]{%
Zhao-16}
\APACinsertmetastar {%
Zhao-16}%
\begin{APACrefauthors}%
Zhao, F.%
, Qiao, L.%
, F.~Shi, P\BPBI T\BPBI Y.%
\BCBL {}\ \BBA {} Shen, D.%
\end{APACrefauthors}%
\unskip\
\newblock
\APACrefYearMonthDay{2016}{}{}.
\newblock
{\BBOQ}\APACrefatitle {Feature fusion via hierarchical supervised local CCA for
  diagnosis of autism spectrum disorder} {Feature fusion via hierarchical
  supervised local cca for diagnosis of autism spectrum disorder}.{\BBCQ}
\newblock
\APACjournalVolNumPages{Brain Imaging and Behavior}{}{}{1-11}.
\PrintBackRefs{\CurrentBib}

\bibitem [\protect \citeauthoryear {%
Zheng%
, Cai%
, Ding%
, Nie%
\BCBL {}\ \BBA {} Hung%
}{%
Zheng%
\ \protect \BOthers {.}}{%
{\protect \APACyear {2015}}%
}]{%
Zhen-15}
\APACinsertmetastar {%
Zhen-15}%
\begin{APACrefauthors}%
Zheng, S.%
, Cai, X.%
, Ding, C\BPBI H.%
, Nie, F.%
\BCBL {}\ \BBA {} Hung, H.%
\end{APACrefauthors}%
\unskip\
\newblock
\APACrefYearMonthDay{2015}{}{}.
\newblock
{\BBOQ}\APACrefatitle {A closed form solution to multi-view low-rank
  regression} {A closed form solution to multi-view low-rank
  regression}.{\BBCQ}
\newblock
\APACjournalVolNumPages{In Proceedings of the Twenty-Ninth AAAI Conference on
  Artificial Intelligence}{(AAAI-15)}{}{1973-1979}.
\PrintBackRefs{\CurrentBib}

\end{thebibliography}
\endgroup

%\newpage
%\section*{Appendix}
\appendix
\numberwithin{equation}{section}
\section*{Appendix}
In the following  sections,  we  present the details of the proposed method,  Satterthwaite approximation to the score test, in addition we present supplementary tables and figures on  our methods to their application to  imaging genetics and  epigenetics. 
\section{Estimation of the linear mixed effect model using ReML}
\label{ap1}
As discussed in  the  literature, we can estimate the variance components using the restricted maximum likelihood (ReML) approach \cite{Harville-74, Lindstrom-88}.  The restricted log-likelihood function of Eq. (13) is  written as:
\begin{eqnarray}
\label{amee1}
\ell_R (\gvc{\theta}) = -\frac{1}{2}ln (|\Theta (\gvc{\theta})|)- \frac{1}{2} ln (|\vc{X}^T \vc{\Theta}^{-1} (\gvc{\theta}) \vc{X} |) - \frac{1}{2} (\vc{y} - \vc{X}\gvc{\hat{\beta}})^T\gvc{\Theta}^{-1} (\gvc{\theta}) (\vc{y} - \vc{X}\gvc{\beta})
\end{eqnarray}

where  $\gvc{\hat{\beta}}$ is the BLUP  of the regression coefficients  $\gvc{\beta})$  $\Theta (\gvc{\theta})= \sigma^2\vc{I} + \tau^{(1)}\vc{K}^{(1)}+ \tau^{(2)}\vc{K}^{(2)}+\tau^{(3)}\vc{K}^{(3)}+ \tau^{(1\times 2)}\vc{K}^{(1\times 2)}+ \tau^{(1\times 3)}\vc{K}^{(1\times 3)} + \tau^{(2\times 3)}\vc{K}^{(2\times 3)} + \tau^{(1\times 2\times 3)}\vc{K}^{(1\times 2\times 3)}$, where $\gvc{\theta} = (\sigma^2, \tau^{(1)}, \tau^{(2)},  \tau^{(3)}, \tau^{1\times 2}, \tau^{2\times 3}, \tau^{1\times 2\times 3})$ are the  variance components. 
To estimate  the variance components, we need to perform the partial derivative  of Eq. (\ref{amee1}) with respective to each variance component:
\begin{eqnarray}
\label{amee2}
\frac{\partial\ell_R}{\partial \gvc{\theta}_i}&=&-\frac{1}{2} \rm{ {\bf tr}}( \vc{W}\vc{G}_i)+  \frac{1}{2} (\vc{y} - \vc{X}\gvc{\hat{\beta}})^T \gvc{\Theta}^{-1} (\gvc{\theta}) \vc{G}_i   \gvc{\Theta}^{-1} (\gvc{\theta})\vc{1}^T (\vc{y} - \vc{X}\gvc{\beta})\nonumber \\
&=&  -\frac{1}{2} \rm{ {\bf tr}}\left( \vc{W}\vc{G}_i \right)+  \frac{1}{2}\vc{y}^T \vc{W}\vc{G}_i\vc{W}\vc{y}=0
\end{eqnarray}
where $\vc{G}_i= \frac{\partial\gvc{\Theta}}{ \gvc{\theta}_i}$ and  $\vc{W}= \gvc{\Theta}^{-1}-  \gvc{\Theta}^{-1}\vc{X}(\vc{X}^T \gvc{\Theta}^{-1}\vc{X})^{-1}\vc{X}^T\gvc{\Theta}^{-1}$. We also have
$\frac{\partial \gvc{\Theta}}{\partial \sigma^2}= \vc{I}$, \,  $\frac{\partial \gvc{\Theta}}{\partial \tau^{(1)}}= \vc{K}^{(1)}$, \,  $\frac{\partial \gvc{\Theta}}{\partial \tau^{(2)}}= \vc{K}^{(2)}$, \, $\frac{\partial \gvc{\Theta}}{\partial \tau^{(3)}}= \vc{K}^{(3)}$, \, $\frac{\partial \gvc{\Theta}}{\partial \tau^{(1\times 2)}}= \vc{K}^{(1\times 2)}$,\,  $\frac{\partial \gvc{\Theta}}{\partial \tau^{(1\times 3)}}= \vc{K}^{(1\times 3)}$,\,  $\frac{\partial \gvc{\Theta}}{\partial \tau^{(1\times 2\times 3)}}= \vc{K}^{(1\times 2\times 3)}$. 
  The $(i, j)$-th element  of the observed  and expected information matrices are
\[\left[ \frac{\partial^2\ell_R}{\partial \gvc{\theta}_i\partial \gvc{\theta}_j}\right]_{ij}=-\frac{1}{2} \rm{{\bf tr}} (\vc{W}\vc{G}_i\vc{W}\vc{G}_j)+ \vc{y}^T \vc{W}\vc{G}_i\vc{W}\vc{G}_i\vc{W}\vc{y},\]
\[\rm{\vc{E}}\left[ \frac{\partial^2\ell_R}{\partial \gvc{\theta}_i\partial \gvc{\theta}_j}\right]_{ij}=-\frac{1}{2} \rm{{\bf tr}} (\vc{W}\vc{G}_i\vc{W}\vc{G}_j)=\mc{I}(\gvc{\theta}),\]
respectively. 
Using Fisher's scoring algorithm (Newton-Raphson method to solve  maximum likelihood equations numerically), given an initial value of unknown parameters at the $h$-th iteration $\gvc{\theta}_{(k+1)}$, the parameters are updated as
\[ \gvc{\theta}_{(k+1)}= \gvc{\theta}_{(k)}+[\mc{I}(\gvc{\theta}^{-1}_{(h)}][\frac{\partial\ell_R}{\partial \gvc{\theta}} ]_{\gvc{\theta}_{(k)}}.\]

In  expectation  maximization (EM) algorithm, we used a set of  initial points  ($0$, $0.00001$, $0.0001$, $0.001$, $0.01$, $0.1$, $1$) of the variance components  for the optimization algorithm and chose the best one (maximized ReML) to avoid the local minim \cite{Laird-87}. For the $\sigma^2$,  we fixed initial values to $\rm{Var}(\vc{y})$.   After  EM update, we then  conducted  the  Fisher's scoring algorithm for the left iterations until the difference between successive log ReML values ($|\ell_{R(h+1)}- \ell_{R(h)}|$) was smaller than $10^{-05}$.  In most  cases the ReML algorithm converged  in less than $50$  iterations  and in  some cases it converged very quickly with $10$ iterations, taking only a few seconds with  an R-program. 
 %\appendix
\section{Satterthwaite approximation to the score test}
\label{ap2}
The restricted score function under the null hypothesis  $H_0:  \tau^{(1)} =\tau^{(2)} = \tau^{(3)} = \tau^{(1\times 2)} = \tau^{(1\times 3)}=\tau^{(2\times 3)} =  \tau^{(1\times 2\times 3)}=0$  is given by:
\begin{eqnarray}
%\label{amee2}
\frac{\partial\ell_R}{\partial \gvc{\theta}_i}\big|_{\gvc{\tau}=0}=  -\frac{1}{2\sigma^2} \rm{ {\bf tr}}\left( \vc{W}_0\vc{G}_i\right)+  \frac{1}{2\sigma^4} (\vc{y} - \vc{X}\gvc{\hat{\beta}})^T \gvc{\Theta}^{-1}  \vc{G}_i (\vc{y} - \vc{X}\gvc{\beta})
\end{eqnarray}
where $\vc{W}_0= \vc{I}-\vc{I} (\vc{I}^T\vc{I})^{-1}\vc{I}^T$ and $\gvc{\tau} =\tau^{(1)} =\tau^{(2)} = \tau^{(3)} = \tau^{(1\times 2)} = \tau^{(1\times 3)}=\tau^{(2\times 3)}$. Since the MLE is $\sqrt{n}$ consistent, the asymptotic distribution of $ S(\hat{\sigma}_0^2)$ can still be approximated by the scaled chi-square distribution. By considering the true value of $\sigma^2$ under null hypothesis as  $\sigma^2_0$, the mean and variance of the test statistic $ S(\hat{\sigma}_0^2)$ are:
 \[\rm{E}[S(\sigma_0^2)]=\frac{1}{2} \rm{{\bf tr}}\left(\vc{W}_0 \vc{K}\right)= \rm{E}[\gamma \chi^2_\nu]=  \gamma\nu, \rm{and}\]
\[\rm{Var}[S(\sigma_0^2)]= \frac{1}{2} \rm{{\bf tr}}\left(\vc{W}_0 \vc{K}\vc{W}_0 \vc{K}\right)= \rm{Var}[\gamma \chi^2\nu]=  2\gamma^2\nu\]

where $\vc{K}= \vc{K}^{(1)}+ \vc{K}^{(2)}+\vc{K}^{(3)}+ \vc{K}^{(1\times 2)} +\vc{K}^{(1\times 3)} +\vc{K}^{(2\times 3)} + \vc{K}^{(1\times2\times 3)}$ and  $\hat{\gvc{\beta}}$. To account for this substitution, we need to estimate  $\gamma$ and $\nu$ by  replacing  the  $\rm{Var}[S(\hat{\sigma}_0^2)]$  based on the efficient information. The elements of the Fisher information matrix $\gvc{\tau}$ are written as:
%\rotatebox{90}{
%{\supptiny
\begin{eqnarray}
\scalemath{0.75}{
%\rotatebox{90}{
\mc{I}_{\gvc{\tau}\gvc{\tau}}=
%\begin{bmatrix}
\begin{bmatrix}
\rm{{\bf tr}} (\vc{A}^{(1)}\vc{A}^{(1)})&\rm{{\bf tr}} (\vc{A}^{(1)}\vc{A}^{(2)}) &\rm{{\bf tr}} (\vc{A}^{(1)} \vc{A}^{(3)})&\rm{{\bf tr}} (\vc{A}^{(1)}\vc{A}^{(1\times 2)})  &\rm{{\bf tr}} (\vc{A}^{(1)} \vc{A}^{(1\times 3)})&\rm{{\bf tr}} (\vc{A}^{(1)}\vc{A}^{(2\times 3)})&\rm{{\bf tr}} (\vc{A}^{(1)}\vc{A}^{(1\times 2 \times 3)})\\
\rm{{\bf tr}} (\vc{A}^{(2)}\vc{A}^{(1)})&\rm{{\bf tr}} (\vc{A}^{(2)}\vc{A}^{(2)}) &\rm{{\bf tr}} (\vc{A}^{(2)} \vc{A}^{(3)})&\rm{{\bf tr}} (\vc{A}^{(2)}\vc{A}^{(1\times 2)})  &\rm{{\bf tr}} (\vc{A}^{(2)} \vc{A}^{(1\times 3)})&\rm{{\bf tr}} (\vc{A}^{(2)}\vc{A}^{(2\times 3)})&\rm{{\bf tr}} (\vc{A}^{(2)}\vc{A}^{(1\times 2 \times 3)})\\
\rm{{\bf tr}} (\vc{A}^{(3)}\vc{A}^{(1)})&\rm{{\bf tr}} (\vc{A}^{(3)}\vc{A}^{(2)}) &\rm{{\bf tr}} (\vc{A}^{(3)} \vc{A}^{(3)})&\rm{{\bf tr}} (\vc{A}^{(3)}\vc{A}^{(1\times 2)})  &\rm{{\bf tr}} (\vc{A}^{(3)} \vc{A}^{(1\times 3)})&\rm{{\bf tr}} (\vc{A}^{(3)}\vc{A}^{(2\times 3)})&\rm{{\bf tr}} (\vc{A}^{(3)}\vc{A}^{(1\times 2 \times 3)})\\
\rm{{\bf tr}} (\vc{A}^{(1\times 2)}\vc{A}^{(1)})&\rm{{\bf tr}} (\vc{A}^{(1\times 2)}\vc{A}^{(2)}) &\rm{{\bf tr}} (\vc{A}^{(1\times 2)} \vc{A}^{(3)})&\rm{{\bf tr}} (\vc{A}^{(1\times 2)}\vc{A}^{(1\times 2)})  &\rm{{\bf tr}} (\vc{A}^{(1\times 2)} \vc{A}^{(1\times 3)})&\rm{{\bf tr}} (\vc{A}^{(1\times 2)}\vc{A}^{(2\times 3)})&\rm{{\bf tr}} (\vc{A}^{(1\times 2)}\vc{A}^{(1\times 2 \times 3)})\\
\rm{{\bf tr}} (\vc{A}^{(1\times 3)}\vc{A}^{(1)})&\rm{{\bf tr}} (\vc{A}^{(1\times 3)}\vc{A}^{(2)}) &\rm{{\bf tr}} (\vc{A}^{(1\times 3)} \vc{A}^{(3)})&\rm{{\bf tr}} (\vc{A}^{(1\times 3)}\vc{A}^{(1\times 2)})  &\rm{{\bf tr}} (\vc{A}^{(1\times 3)} \vc{A}^{(1\times 3)})&\rm{{\bf tr}} (\vc{A}^{(1\times 3)}\vc{A}^{(2\times 3)})&\rm{{\bf tr}} (\vc{A}^{(1\times 3)}\vc{A}^{(1\times 2 \times 3)})\\
\rm{{\bf tr}} (\vc{A}^{(2\times 3)}\vc{A}^{(1)})&\rm{{\bf tr}} (\vc{A}^{(2\times 3)}\vc{A}^{(2)}) &\rm{{\bf tr}} (\vc{A}^{(2\times 3)} \vc{A}^{(3)})&\rm{{\bf tr}} (\vc{A}^{(2\times 3)}\vc{A}^{(1\times 2)})  &\rm{{\bf tr}} (\vc{A}^{(2\times 3)} \vc{A}^{(1\times 3)})&\rm{{\bf tr}} (\vc{A}^{(2\times 3)}\vc{A}^{(2\times 3)})&\rm{{\bf tr}} (\vc{A}^{(2\times 3)}\vc{A}^{(1\times 2 \times 3)})\\
\rm{{\bf tr}} (\vc{A}^{(1\times 2\times 3)}\vc{A}^{(1)})&\rm{{\bf tr}} (\vc{A}^{(1 \times 2\times 3)}\vc{A}^{(2)}) &\rm{{\bf tr}} (\vc{A}^{(1\times 2 \times 3)} \vc{A}^{(3)})&\rm{{\bf tr}} (\vc{A}^{(1\times 2\times 3)}\vc{A}^{(1\times 2)})  &\rm{{\bf tr}} (\vc{A}^{(1\times 2\times 3)} \vc{A}^{(1\times 3)})&\rm{{\bf tr}} (\vc{A}^{(1\times 2\times 3)}\vc{A}^{(2\times 3)})&\rm{{\bf tr}} (\vc{A}^{(1\times 2\times 3)}\vc{A}^{(1\times 2 \times 3)})\\
\end{bmatrix}}, \nonumber
\end{eqnarray}
%}
%{\tiny
\[\scalemath{0.75}{
 \mc{I}_{\gvc{\tau} \sigma^2} =\frac{1}{2}[\rm{{\bf tr}} (\vc{A}^{(1)}) \qquad\rm{{\bf tr}} (\vc{A}^{(2)})\qquad \rm{{\bf tr}} (\vc{A}^{(1)})\qquad \rm{{\bf tr}} (\vc{A}^{(1\times 2)}) \qquad \rm{{\bf tr}}  (\vc{A}^{(1\times 3)}) \qquad \rm{{\bf tr}}  (\vc{A}^{(2\times 3)}) \qquad \rm{{\bf tr}}  (\vc{A}^{(1\times 2\times 3)}}],\]

 and   $\mc{I}_{\sigma^2 \sigma^2}= \frac{1}{2} \rm{{\bf tr}} (\vc{W}_0 \vc{W}_0)$, where $\vc{A}^{(1)}= \vc{W}_0 \vc{K}^{(1)}$,\, $\vc{A}^{(2)}= \vc{W}_0 \vc{K}^{(2)}$,\, $\vc{A}^{(3)}= \vc{W}_0 \vc{K}^{(3)}$, \,  $\vc{A}^{(1\times 2)}= \vc{W}_0 \vc{K}^{(1\times 2)}$,\, $\vc{A}^{(1\times 3)}= \vc{W}_0 \vc{K}^{(1\times 3)},$\, $\vc{A}^{(2\times 3)}= \vc{W}_0 \vc{K}^{(2\times 3)}$,\, $\vc{A}^{(1\times 2\times 3)}= \vc{W}_0 \vc{K}^{(1\times 2\times 3)}$. Using these information matrices, we have the efficient information  $\tilde{\mc{I}}_{\gvc{\tau}\gvc{\tau}}= \mc{I}_{\gvc{\tau},\gvc{\tau}}- \mc{I}_{\gvc{\tau} \sigma^2}^T \mc{I}_{\sigma^2 \sigma^2}^{-1}  \mc{I}_{\gvc{\tau} \sigma^2}$ and 
$\rm{Var}[S(\hat{\sigma}_0^2)]=\rm{SUM}[\tilde{\mc{I}}_{\gvc{\tau}\gvc{\tau}}]$, where the operator "SUM" indicates the sum of all the  element in  the matrix.  By considering the adjusted parameters $\hat{\gamma}=\frac{\rm{Var}[S(\hat{\sigma}_0^2)]}{2\rm{E}[S(\sigma_0^2)]}$ and $\hat{\nu}=\frac{2\rm{E}[S(\sigma_0^2)^2]}{\rm{Var}[S(\hat{\sigma}_0^2)]} $, the $p-$value of an experimental  score statistic  $S(\hat{\sigma}_0^2)$ is obtained   using the  scaled chi-square distribution $\hat{\gamma}\chi^2_{\hat{\nu}}$.  

The score test statistic  $S_{\rm{I}}(\tilde{\tau_{\rm{I}}})$ defined in Eq. (16) for the higher order interaction effect that testing the null hypothesis   $H_0: \tau^{(1\times 2\times 3)}=0$ is approximated by a scaled chi-square distribution $\hat{\gamma}_{\rm{I}}\chi^2_{\hat{\nu_I}}$. To do this,   let $\Sigma= \sigma^2\vc{I} + \tau^{(1)}\vc{K}^{(1)}+ \tau^{(2)}\vc{K}^{(2)}+\tau^{(3)}\vc{K}^{(3)}+ \tau^{1\times 2}\vc{K}^{(1\times 2)}+ \tau^{(1\times 3)}\vc{K}^{(1\times 3)} + \tau^{(2\times 3)}\vc{K}^{(2\times 3)}$, and $\tau^{(1)}, \tau^{(2)}, \tau^{(3)}, \tau^{(1\times 2)},  \tau^{(1\times 3)}, \tau^{(2\times 3)}$,  and   $\sigma^2$  are model parameters under the null model $  \vc{y}=\vc{X}\gvc{\beta}+\vc{h}_{\vc{M}^{(1)}}+\vc{h}_{\vc{M}^{(2)}}+\vc{h}_{\vc{M}^{(3)}}+ \vc{h}_{\vc{M}^{(1)}\times \vc{M}^{(2)}} + \vc{h}_{\vc{M}^{(1)}\times \vc{M}^{(3)}}+\gvc{\epsilon}$. The score function  Eq. (\ref{amee1}) under the null hypothesis becomes
\begin{eqnarray}
\frac{\partial\ell_R}{\partial \tau^{(1\times 2\times 3)}}\big|_{\tau^{(1\times 2\times 3)}=0} &=&  -\frac{1}{2\sigma^2}[ \rm{ {\bf tr}}\left( \vc{W}_{01}\vc{K}^{(1\times 2\times 3)}\right)- (\vc{y} - \vc{X}\gvc{\hat{\beta}})^T \vc{\Sigma}^{-1}\vc{K}^{(1\times 2\times 3)} \vc{\Sigma}^{-1}(\vc{y} - \vc{X}\gvc{\beta})]\nonumber\\
&=& -\frac{1}{2}[\rm{ {\bf tr}}\left( \vc{W}_{01}\vc{K}^{(1\times 2\times 3)}\right)- \vc{y}^T\vc{W}_{01}\vc{K}^{(1\times 2\times 3)} \vc{W}_{01}\vc{y}) 
\end{eqnarray}
where  $\vc{W}_{01}= \Sigma^{-1}- \Sigma^{-1}   \vc{X}(\vc{X}^T  \Sigma^{-1}  \vc{X})^{-1} \vc{X}^T\Sigma^{-1}$ is the projection matrix under the null hypothesis.  The  test statistic for the higher order interaction effect is   as follows:
\begin{eqnarray}
S_{\rm{I}}(\gvc{\tau}_{\rm{I}})= \frac{1}{2\sigma^2_0} \vc{y}^T \vc{W}_{01} \vc{K}^{(1\times2\times 3)}\vc{W}_{01}\vc{y},\nonumber 
\end{eqnarray}
where   $\gvc{\tau}_{\rm{I}} = (\tau^{(1)},\tau^{(2)},\tau^{(3)}, \tau^{(1\times 2)}).$  Similarly for overall effect test,  we can use the  Satterthwaite method to  approximate the distribution of   higher order intersection test statistic  $S_{\rm{I}}(\tau_{\rm{I}})$ by a scaled chi-square distribution  with a scaled   $\gamma_{\rm{I}}$ and degree of freedom $\nu_{\rm{I}}$, i.e., $S_{\rm{I}}(\tau_{\rm{I}})\sim \gamma_{\rm{I}} \chi^2_{\nu_{\rm{I}}}$.  The mean and variance of the test statistic $S_{\rm{I}}(\gvc{\tau}_{\rm{I}})$ are:
 \[\rm{E}[S_{\rm{I}}(\gvc{\tau}_{\rm{I}})]=\frac{1}{2} \rm{{\bf tr}}\left(\vc{W}_{01} \vc{K}^{(1\times 2\times 3)}\right)= \rm{E}[\gamma_{\rm{I}} \chi^2_{\nu_{\rm{I}}}]=  \gamma_{\rm{I}}\nu_{\rm{I}}, \rm{and}\]
\[\rm{Var}[S_{\rm{I}}(\gvc{\tau}_{\rm{I}})]= \frac{1}{2} \rm{{\bf tr}}\left(\vc{W}_0 \vc{K}^{(1\times 2\times 3)}\vc{W}_0 \vc{K}^{(1\times 2\times 3)}\right)= \rm{Var}[\gamma_{\rm{I}} \chi^2_{\nu_{\rm{I}}}]=  2\gamma_{\rm{I}}^2\nu_{\rm{I}}\]
where $\vc{K}= \vc{K}^{(1)}+ \vc{K}^{(2)}+\vc{K}^{(3)}+ \vc{K}^{(1\times 2)} +\vc{K}^{(1\times 3)} +\vc{K}^{(2\times 3)} + \vc{K}^{(1\times2\times 3)}$, 
$\gamma_{\rm{I}}=\frac{\rm{Var}[S_{\rm{I}}(\gvc{\tau}_{\rm{I}}]}{2\rm{E}[S_{\rm{I}}(\gvc{\tau}_{\rm{I}}]}$ and $\nu_{\rm{I}}=\frac{2\rm{E}[S_{\rm{I}}(\gvc{\tau}_{\rm{I}}]}{\rm{Var}[S_{\rm{I}}(\gvc{\tau}_{\rm{I}}]}$, receptively.  In practice, the unknown  model parameters  $\tau^{(1)}, \tau^{(2)}, \tau^{(3)}, \tau^{(1\times 2)},  \tau^{(1\times 3)}, \tau^{(2\times 3)}$,  and   $\sigma^2$ are estimated by their respective  ReML estimates  $\hat{\tau}^{(1)}, \hat{\tau}^{(2)}, \hat{\tau}^{(3)}, \hat{\tau}^{(1\times 2)},  \hat{\tau}^{(1\times 3)}, \hat{\tau}^{(2\times 3)}$,  and   $\hat{\sigma}^2$ under the null hypothesis.  The scaled parameter and degree of freedom  are estimated by the MOM.   Specifically,
 $\hat{\gamma}_{\rm{I}}= \frac{\widehat{\rm{Var}[S_{\rm{I}}(\gvc{\tau}_{\rm{I}})]}} {2\widehat{\rm{E}[S_{\rm{I}}(\gvc{\tau}_{\rm{I}})]}},$  and  $\hat{\nu}_{\rm{I}}=\frac{2\widehat{\rm{E}[S_{\rm{I}}(\gvc{\tau}_{\rm{I}})]}}{\widehat{\rm{Var}[S_{\rm{I}}(\gvc{\tau}_{\rm{I}})]}}$, 
 where  $\widehat{\rm{E}[S_{\rm{I}}(\gvc{\tau}_{\rm{I}})]}= \frac{1}{2} \rm{{\bf tr}}\left(\vc{W}_{01} \vc{K}^{(1\times 2\times 3)}\right)$ and  $\widehat{\rm{Var}[S_{\rm{I}}(\gvc{\tau}_{\rm{I}})]}=  \frac{1}{2} \rm{{\bf tr}}\left(\vc{W}_0 \vc{K}^{(1\times 2\times 3)}\vc{W}_0 \vc{K}^{(1\times 2\times 3)}\right)- \frac{\gvc{\Delta} \gvc{\xi}^{-1}\gvc{\Delta}}{2}$, in which

\[ \scalemath{0.7}{ \gvc{\Delta}=[\rm{{\bf tr}} ( \vc{B}^{(1\times 2\times 3)}\vc{W}_{01}) \qquad\rm{{\bf tr}} (\vc{B}^{(1\times 2\times 3)}\vc{B}^{(1)}) \qquad \rm{{\bf tr}} ( \vc{B}^{(1\times 2\times 3)} \vc{B}^{(2)})  \qquad \rm{{\bf tr}} (\vc{B}^{(1\times 2\times 3)} \vc{B}^{(3)})  \qquad \rm{{\bf tr}} (\vc{B}^{(1\times 2\times 3)}\vc{B}^{(1\times 2)})\qquad\rm{{\bf tr}} ( \vc{B}^{(1\times 2\times 3)} \vc{B}^{(1\times 3)})\qquad \rm{{\bf tr}} ( \vc{B}^{(1\times 2\times 3)} \vc{B}^{(2\times 3)})]}\]

 %{\scriptsize 
%{\tiny 
\begin{eqnarray}
\scalemath{0.75}{
\gvc{\xi}=
\begin{bmatrix}
\rm{{\bf tr}} (\vc{W}_{01}^2) &\rm{{\bf tr}} (\vc{W}_{01}\vc{B}^{(1)})&\rm{{\bf tr}} (\vc{W}_{01}\vc{B}^{(2)}) &\rm{{\bf tr}} (\vc{W}_{01}\vc{B}^{(3)})&\rm{{\bf tr}} (\vc{W}_{01}\vc{B}^{(1\times 2)})&\rm{{\bf tr}} (\vc{W}_{01}\vc{B}^{(1\times 3)})&\rm{{\bf tr}} (\vc{W}_{01}\vc{B}^{(2\times 3)})\\
\rm{{\bf tr}} (\vc{W}_{01} \vc{B}^{(1)}) &\rm{{\bf tr}} (\vc{B}^{(1)} \vc{B}^{(1)})&\rm{{\bf tr}} (\vc{B}^{(1)} \vc{B}^{(2)})&\rm{{\bf tr}} (\vc{B}^{(1)} \vc{B}^{(3)})&\rm{{\bf tr}} (\vc{B}^{(1)} \vc{B}^{(1\times 2)})&\rm{{\bf tr}} (\vc{B}^{(1)} \vc{B}^{(1\times 3)})&\rm{{\bf tr}} (\vc{B}^{(1)} \vc{B}^{(2\times 3)})\\
\rm{{\bf tr}} (\vc{W}_{01} \vc{B}^{(2)}) &\rm{{\bf tr}} (\vc{B}^{(2)} \vc{B}^{(1)})&\rm{{\bf tr}} (\vc{B}^{(2)} \vc{B}^{(2)})&\rm{{\bf tr}} (\vc{B}^{(2)} \vc{B}^{(3)})&\rm{{\bf tr}} (\vc{B}^{(2)} \vc{B}^{(1\times 2)})&\rm{{\bf tr}} (\vc{B}^{(2)} \vc{B}^{(1\times 3)})&\rm{{\bf tr}} (\vc{B}^{(2)} \vc{B}^{(2\times 3)})\\
\rm{{\bf tr}} (\vc{W}_{01} \vc{B}^{(3)}) &\rm{{\bf tr}} (\vc{B}^{(3)} \vc{B}^{(1)})&\rm{{\bf tr}} (\vc{B}^{(3)} \vc{B}^{(2)})&\rm{{\bf tr}} (\vc{B}^{(3)} \vc{B}^{(3)})&\rm{{\bf tr}} (\vc{B}^{(3)} \vc{B}^{(1\times 2)})&\rm{{\bf tr}} (\vc{B}^{(3)} \vc{B}^{(1\times 3)})&\rm{{\bf tr}} (\vc{B}^{(3)} \vc{B}^{(2\times 3)})\\
\rm{{\bf tr}} (\vc{W}_{01} \vc{B}^{(1\times 2)}) &\rm{{\bf tr}} (\vc{B}^{(1\times 2)} \vc{B}^{(1)})&\rm{{\bf tr}} (\vc{B}^{(1\times 2)} \vc{B}^{(2)})&\rm{{\bf tr}} (\vc{B}^{(1\times 2)} \vc{B}^{(3)})&\rm{{\bf tr}} (\vc{B}^{(1\times 2)} \vc{B}^{(1\times 2)})&\rm{{\bf tr}} (\vc{B}^{(1\times 2)} \vc{B}^{(1\times 3)})&\rm{{\bf tr}} (\vc{B}^{(1\times 2)} \vc{B}^{(2\times 3)})\\
\rm{{\bf tr}} (\vc{W}_{01} \vc{B}^{(1\times 3)}) &\rm{{\bf tr}} (\vc{B}^{(1\times 3)} \vc{B}^{(1)})&\rm{{\bf tr}} (\vc{B}^{(1\times 3)} \vc{B}^{(2)})&\rm{{\bf tr}} (\vc{B}^{(1\times 3)} \vc{B}^{(3)})&\rm{{\bf tr}} (\vc{B}^{(1\times 3)} \vc{B}^{(1\times 2)})&\rm{{\bf tr}} (\vc{B}^{(1\times 3)} \vc{B}^{(1\times 3)})&\rm{{\bf tr}} (\vc{B}^{(1\times 3)} \vc{B}^{(2\times 3)})\\
\rm{{\bf tr}} (\vc{W}_{01} \vc{B}^{(2\times 3)}) &\rm{{\bf tr}} (\vc{B}^{(2\times 3)} \vc{B}^{(1)})&\rm{{\bf tr}} (\vc{B}^{(2\times 3)} \vc{B}^{(2)})&\rm{{\bf tr}} (\vc{B}^{(2\times 3)} \vc{B}^{(3)})&\rm{{\bf tr}} (\vc{B}^{(2\times 3)} \vc{B}^{(1\times 2)})&\rm{{\bf tr}} (\vc{B}^{(2\times 3)} \vc{B}^{(1\times 3)})&\rm{{\bf tr}} (\vc{B}^{(2\times 3)} \vc{B}^{(2\times 3)})\\
\end{bmatrix},} \nonumber
\end{eqnarray}
%}
where $\vc{B}^{(1)}= \vc{W}_{01} \vc{K}^{(1)}$,\, $\vc{B}^{(2)}= \vc{W}_{01} \vc{K}^{(2)}$,\, $\vc{B}^{(3)}= \vc{W}_{01} \vc{K}^{(3)}$, \,  $\vc{B}^{(1\times 2)}= \vc{W}_{01} \vc{K}^{(1\times 2)}$,\, $\vc{B}^{(1\times 3)}= \vc{W}_{01} \vc{K}^{(1\times 3)},$\, $\vc{B}^{(2\times 3)}= \vc{W}_{01} \vc{K}^{(2\times 3)}$,\, $\vc{B}^{(1\times 2\times 3)}= \vc{W}_{01} \vc{K}^{(1\times 2\times 3)}$. The $p-$value of an  observed higher order interaction effect test  score statistic $S_{\rm{I}}(\tau_{\rm{I}})$ is obtained   using the  scaled chi-square distribution $\hat{\gamma}_{\rm{I}}\chi^2_{\hat{\nu_{\rm{I}}}}$.
\section{Supplementary  figures and tables}
\label{ap3}

\begin{table}
\begin{center}
\caption {The selected  significant genes-derived SNP, ROIs and gene-derived DNA methylation using the proposed method (KMDHOI). The $p-$values threshold was fixed to $0.01$.}
\scalebox{0.8}[0.9]{
\begin{tabular}{lcccccccccccccc}
\hline
&&&\multicolumn{10}{c}{\rm{KMDHOI}}\tabularnewline
\multicolumn{1}{l}{Genetics}&\multicolumn{1}{c}{Imaging}&\multicolumn{1}{c}{Epigenetics}&\multicolumn{1}{c}{$\sigma^2$}&\multicolumn{1}{c}{$\tau^{(1)}$}&\multicolumn{1}{c}{$\tau^{(2)}$}&\multicolumn{1}{c}{$\tau^{(3)}$}&\multicolumn{1}{c}{$\tau^{1\times 2}$}&\multicolumn{1}{c}{$\tau^{1\times 3}$}&\multicolumn{1}{c}{$\tau^{2\times 3}$}&\multicolumn{1}{c}{$\tau^{1\times 2\times 3}$} &\multicolumn{1}{c}{\rm{OVA}} &\multicolumn{1}{c}{\rm{HOI}}\tabularnewline\hline
${\bf BDNF}$&$ {\bf AMYG.L}$&${\bf DUSP1}
$&$0.5658$&$0.0559$&$0.0887$&$0.0041$&$0.0494$&$2.1147$&$0.0000$&$0.0100$&$0.0345$&$0.0069$\tabularnewline
$ {\bf BDNF}$&${\bf CRBL10.L} $&$ {\bf FBXO28}
$&$0.5589$&$0.0334$&$0.1414$&$0.0157$&$1.2657$&$0.0000$&$0.2034$&$0.0100$&$0.0102$&$0.0058$\tabularnewline
$ {\bf BDNF}$&$ {\bf CRBL3.R} $&$ {\bf DUSP1}$&$0.4255$&$0.2077$&$0.0000$&$0.0556$&$0.7339$&$2.1418$&$1.0345$&$0.0100$&$0.0414$&$0.0072$\tabularnewline
$ {\bf BDNF}$&${\bf ORBsup.L} $&$  {\bf FBXO28}$&$0.5515$&$0.0033$&$0.4506$&$0.0535$&$0.8534$&$0.6955$&$0.0000$&$0.0100$&$0.0078$&$0.0013$\tabularnewline
$ {\bf BDNF}$&$ {\bf LING.R} $&$ {\bf CRABP1}
$&$0.5510$&$0.5258$&$0.2416$&$0.0640$&$0.9032$&$0.7442$&$0.0000$&$0.0100$&$0.0114$&$0.0074$\tabularnewline
$ {\bf BDNF}$&$ {\bf IPL.R} $&$  {\bf FBXO28}$&$0.4471$&$0.0388$&$0.1458$&$0.0644$&$1.2503$&$2.3689$&$0.0000$&$1.9045$&$0.0367$&$0.0024$\tabularnewline
$ {\bf CHGA}$&$  {\bf AMYG.L}$&$ {\bf HOXA9}
$&$0.4326$&$0.0027$&$0.3130$&$0.3598$&$1.5185$&$0.0000$&$1.1948$&$0.7093$&$0.0474$&$0.0075$\tabularnewline
$  {\bf  CHGB}$&$ {\bf CRBL3.L} $&$ {\bf DUSP1}$&$0.6779$&$0.0321$&$0.1444$&$0.1232$&$0.2505$&$0.5545$&$0.0000$&$0.0100$&$0.0454$&$0.0025$\tabularnewline
$  {\bf  CHGB}$&$ {\bf DCG.L} $&$ {\bf DUSP1}$&$0.5253$&$0.0247$&$0.0213$&$0.0570$&$1.2794$&$1.0467$&$0.0000$&$0.0100$&$0.0408$&$0.0035$\tabularnewline
$ {\bf  CHGB}$&$ {\bf STG.L} $&$ {\bf DUSP1}$&$0.5415$&$0.1053$&$0.0360$&$0.0017$&$1.1720$&$1.2059$&$0.0000$&$0.0100$&$0.0293$&$0.0035$\tabularnewline
${\bf CLINT1}$&$ {\bf IPL.R} $&${\bf GPSN2}
$&$0.4074$&$0.6728$&$0.0404$&$0.0514$&$1.2476$&$1.6953$&$0.0000$&$0.0100$&$0.0228$&$0.0042$\tabularnewline
${\bf COMTD1}$&$ {\bf ROL.R}$&$  {\bf FBXO28}$&$0.5717$&$0.0740$&$0.0095$&$0.0479$&$2.5561$&$0.3378$&$0.0000$&$2.3490$&$0.0370$&$0.0083$\tabularnewline
${\bf DAOA}$&$ {\bf ORBsup.L}$&$ {\bf DUSP1}$&$0.3721$&$0.1079$&$0.0731$&$0.0097$&$0.5586$&$3.5344$&$0.0000$&$0.0100$&$0.0490$&$0.0012$\tabularnewline
${\bf DISC1}$&$ {\bf IPL.R} $&${\bf PLAGL1}
$&$0.3370$&$0.0161$&$0.0547$&$0.1417$&$1.5286$&$3.2871$&$0.0000$&$1.4004$&$0.0439$&$0.0029$\tabularnewline
${\bf DRD2}$&$ {\bf ORBsup.R}$&$ {\bf DUSP1}$&$0.3772$&$0.0000$&$0.3189$&$0.0073$&$1.1268$&$2.0353$&$0.9757$&$0.0100$&$0.0445$&$0.0062$\tabularnewline
${\bf DTNBP1}$&${\bf IPL.R} $&${\bf SRF}
$&$0.3940$&$0.1987$&$0.0637$&$0.2312$&$1.4537$&$1.8458$&$0.0000$&$0.0100$&$0.0267$&$0.0037$\tabularnewline
${\bf ERBB4}$&${\bf CRBLCrus2.R}$&$ {\bf DUSP1}$&$0.2576$&$0.0000$&$0.0413$&$0.0401$&$1.2244$&$3.1904$&$1.0830$&$0.0100$&$0.0459$&$0.0088$\tabularnewline
${\bf ERBB4}$&$ {\bf IPL.R} $&${\bf FEN1}
$&$0.4484$&$0.0000$&$0.0762$&$0.0357$&$0.8799$&$2.9942$&$0.1061$&$0.0100$&$0.0294$&$0.0074$\tabularnewline
${\bf ERBB4}$&$ {\bf IPL.R} $&$ {\bf HOXB4}
$&$0.3308$&$0.6704$&$0.0669$&$0.0222$&$2.3489$&$1.9820$&$0.0000$&$0.0100$&$0.0396$&$0.0082$\tabularnewline
${\bf GABBR1}$&$ {\bf LING.R}$&$ {\bf CRABP1}$&$0.5610$&$0.0068$&$0.4045$&$0.0563$&$1.8524$&$0.0000$&$0.7076$&$0.0100$&$0.0166$&$0.0072$\tabularnewline
${\bf GABRB2}$&$ {\bf SMA.R}$&$ {\bf EDNRB}
$&$0.5276$&$0.0139$&$0.1276$&$0.0364$&$1.3359$&$2.4832$&$0.0000$&$2.0971$&$0.0283$&$0.0061$\tabularnewline
${\bf GRIN2B}$&$ {\bf IPL.R} $&$ {\bf HOXB4}
$&$0.3523$&$0.2359$&$0.0352$&$0.0297$&$2.1599$&$2.1357$&$0.0000$&$0.0100$&$0.0421$&$0.0012$\tabularnewline
${\bf GRM3}$&$ {\bf SMA.R}$&$ {\bf EYA4}
$&$0.6435$&$0.0088$&$0.3928$&$0.0384$&$0.5390$&$1.5425$&$0.0000$&$0.4976$&$0.0237$&$0.0024$\tabularnewline
${\bf HTR2A}$&${\bf ITG.L}$&$  {\bf FBXO28}$&$0.4496$&$0.3074$&$0.0087$&$0.0970$&$0.1874$&$2.0080$&$0.0000$&$0.0100$&$0.0211$&$0.0016$\tabularnewline
${\bf IL10RA}$&$ {\bf LING.L}$&$ {\bf CRABP1}$&$0.5128$&$0.1846$&$0.1055$&$0.0436$&$1.3817$&$0.0000$&$0.1076$&$0.0100$&$0.0373$&$0.0035$\tabularnewline
${\bf IL10RA}$&$ {\bf TPOsup.R}$&$ {\bf CRABP1}$&$0.5644$&$0.2024$&$0.0000$&$0.0364$&$1.3936$&$1.1762$&$0.0700$&$0.0100$&$0.0425$&$0.0098$\tabularnewline
${\bf IL1B}$&$ {\bf AMYG.L}$&$  {\bf FBXO28}$&$0.5708$&$0.0130$&$0.4484$&$0.0755$&$1.8572$&$0.0000$&$0.3580$&$1.0757$&$0.0445$&$0.0040$\tabularnewline
${\bf IL1B}$&${\bf CAU.R}$&$  {\bf FBXO28}$&$0.6755$&$0.0038$&$0.0229$&$0.1225$&$1.1013$&$0.0000$&$0.1307$&$1.3606$&$0.0383$&$0.0002$\tabularnewline
${\bf IL1B}$&$ {\bf PoCG.R}$&$  {\bf FBXO28}$&$0.5837$&$0.0189$&$0.1827$&$0.1247$&$1.8403$&$0.0000$&$0.3469$&$1.0603$&$0.0202$&$0.0007$\tabularnewline
${\bf MAGI1}$&$ {\bf CRBL6.L}$&$ {\bf CRABP1}$&$0.3178$&$0.5873$&$0.0201$&$0.0249$&$1.7962$&$1.7014$&$0.0000$&$0.0100$&$0.0346$&$0.0045$\tabularnewline
${\bf MAGI2}$&$ {\bf CRBLCrus1.L}$&$  {\bf FBXO28}$&$0.1833$&$0.3246$&$0.0000$&$0.2693$&$1.1963$&$2.2426$&$1.3683$&$0.0100$&$0.0288$&$0.0000$\tabularnewline
${\bf MAGI2}$&$ {\bf DCG.R}$&$  {\bf FBXO28}$&$0.3211$&$0.0000$&$0.1485$&$0.0513$&$2.0150$&$2.5252$&$0.3083$&$0.0100$&$0.0298$&$0.0057$\tabularnewline
\hline
\end{tabular}}
\end{center}
\label{tb:Lgene768}
\end{table}

\begin{table}[!tbp]
\begin{center}
\caption{\textbf{Table~$5$ continued:}}
\scalebox{0.8}[0.95]{
\begin{tabular}{lcccccccccccccc}
\hline
&&&\multicolumn{10}{c}{\rm{KMDHOI}}\tabularnewline
\multicolumn{1}{l}{Genetics}&\multicolumn{1}{c}{Imaging}&\multicolumn{1}{c}{Epigenetics}&\multicolumn{1}{c}{$\sigma^2$}&\multicolumn{1}{c}{$\tau^{(1)}$}&\multicolumn{1}{c}{$\tau^{(2)}$}&\multicolumn{1}{c}{$\tau^{(3)}$}&\multicolumn{1}{c}{$\tau^{1\times 2}$}&\multicolumn{1}{c}{$\tau^{1\times 3}$}&\multicolumn{1}{c}{$\tau^{2\times 3}$}&\multicolumn{1}{c}{$\tau^{1\times 2\times 3}$} &\multicolumn{1}{c}{\rm{OVA}} &\multicolumn{1}{c}{\rm{HOI}}\tabularnewline\hline
${\bf MAGI2}$&$ {\bf LING.L}$&$ {\bf CRABP1}$&$0.1813$&$0.4366$&$0.0000$&$0.0885$&$1.5370$&$2.6299$&$1.0271$&$0.0100$&$0.0470$&$0.0000$\tabularnewline
${\bf MAGI2}$&$ {\bf IPL.R} $&${\bf SRF}
$&$0.2510$&$0.2349$&$0.0325$&$0.2015$&$1.4611$&$3.1453$&$0.0000$&$0.0100$&$0.0198$&$0.0087$\tabularnewline
${\bf MAGI2}$&$ {\bf IPL.L}$&$  {\bf FBXO28}$&$0.1833$&$0.3778$&$0.0000$&$0.3044$&$0.9808$&$2.3888$&$1.2203$&$0.0100$&$0.0457$&$0.0000$\tabularnewline
${\bf MICB}$&${\bf Vermis3}$&$  {\bf FBXO28}$&$0.6579$&$0.0806$&$0.0199$&$0.0444$&$0.0000$&$0.2964$&$0.4610$&$0.0100$&$0.0458$&$0.0060$\tabularnewline
${\bf NOS1AP}$&$ {\bf IPL.R} $&$ {\bf DUSP1}
$&$0.2939$&$0.0000$&$0.2870$&$0.3240$&$0.8961$&$1.8125$&$1.2330$&$0.0100$&$0.0185$&$0.0012$\tabularnewline
${\bf NOTCH4}$&${\bf Vermis3}$&$  {\bf FBXO28}$&$0.7878$&$0.0000$&$0.0183$&$0.0896$&$0.3783$&$0.2783$&$0.0057$&$1.1039$&$0.0256$&$0.0054$\tabularnewline
${\bf NR4A2}$&$ {\bf PAL.R}$&$ {\bf CRABP1}$&$0.4953$&$0.1173$&$0.0000$&$0.0237$&$1.4544$&$0.3867$&$0.7814$&$0.0100$&$0.0491$&$0.0078$\tabularnewline
${\bf NRG1}$&$ {\bf IPL.R} $&$ {\bf PLAGL1}
$&$0.3682$&$0.0024$&$0.2162$&$0.1270$&$1.1227$&$2.7930$&$0.0000$&$0.2056$&$0.0284$&$0.0000$\tabularnewline
${\bf NUMBL}$&$ {\bf ORBsup.R}$&${\bf CDKN1A}
$&$0.5418$&$0.0585$&$0.4680$&$0.0865$&$0.6606$&$0.4765$&$0.0000$&$0.0100$&$0.0479$&$0.0065$\tabularnewline
${\bf PDLIM5}$&$ {\bf ORBsup.R}$&${\bf CCND2}
$&$0.4108$&$0.0109$&$0.4396$&$0.2792$&$0.0000$&$2.5753$&$0.2739$&$2.2025$&$0.0437$&$0.0016$\tabularnewline
${\bf PDLIM5}$&$ {\bf IPL.R} $&$ {\bf DUSP1}$&$0.3648$&$0.0000$&$0.0804$&$0.2182$&$1.5177$&$1.9409$&$1.0276$&$0.0100$&$0.0183$&$0.0005$\tabularnewline
${\bf PDLIM5}$&$ {\bf IPL.R} $&$ {\bf PLAGL1}
$&$0.3134$&$0.0000$&$0.2173$&$0.2447$&$1.5550$&$1.8347$&$1.5020$&$0.0100$&$0.0460$&$0.0027$\tabularnewline
${\bf PDLIM5}$&$ {\bf PoCG.R}$&$ {\bf DUSP1}$&$0.3598$&$0.0000$&$0.2256$&$0.0139$&$0.9189$&$1.7853$&$1.2796$&$0.0100$&$0.0498$&$0.0004$\tabularnewline
${\bf PLXNA2}$&$ {\bf SMA.R}$&$ {\bf RB1}
$&$0.2386$&$0.2186$&$0.2257$&$0.0341$&$0.8229$&$3.2966$&$0.0000$&$0.0100$&$0.0432$&$0.0017$\tabularnewline
${\bf PPP3CC}$&$ {\bf IPL.R} $&$  {\bf FBXO28}$&$0.7594$&$0.0000$&$0.1627$&$0.0338$&$0.8919$&$0.5701$&$0.6865$&$0.0100$&$0.0199$&$0.0012$\tabularnewline
${\bf SLC18A1}$&$ {\bf ORBsup.R}$&$ {\bf FHIT}
$&$0.4096$&$0.0065$&$0.5356$&$0.2276$&$1.3456$&$0.0000$&$1.0003$&$0.1323$&$0.0495$&$0.0001$\tabularnewline
${\bf SLC18A1}$&$ {\bf ORBsup.R}$&$ {\bf PLAGL1}
$&$0.2869$&$0.0000$&$0.3909$&$0.1933$&$1.0186$&$1.3148$&$1.3676$&$0.0100$&$0.0373$&$0.0003$\tabularnewline
${\bf SLC18A1}$&${\bf Vermis45}$&$ {\bf TFPI2}
$&$0.5571$&$0.0447$&$0.0815$&$0.0020$&$0.0000$&$1.0458$&$0.5579$&$0.0100$&$0.0354$&$0.0007$\tabularnewline
${\bf SNAP29}$&$ {\bf ORBsup.R}$&$ {\bf DUSP1}$&$0.5384$&$0.0333$&$0.3235$&$0.0685$&$0.2128$&$1.3135$&$0.0000$&$0.0100$&$0.0417$&$0.0039$\tabularnewline
$ {\bf TDRD3}$&$ {\bf CRBL10.R}$&$ {\bf EDNRB}
$&$0.5890$&$0.3319$&$0.0000$&$0.0310$&$0.9546$&$0.5359$&$0.8984$&$0.0100$&$0.0236$&$0.0025$\tabularnewline
${\bf TDRD3}$&$  {\bf CRBL3.L} $&$ {\bf EYA4}
$&$0.5162$&$0.1651$&$0.1782$&$0.0612$&$0.4628$&$0.4991$&$0.0000$&$0.0100$&$0.0180$&$0.0019$\tabularnewline
${\bf TDRD3}$&$ {\bf  CRBL45.L}$&$ {\bf CCND2}$&$0.5346$&$0.0523$&$0.2672$&$0.1649$&$0.3055$&$1.1277$&$0.0000$&$0.0100$&$0.0124$&$0.0019$\tabularnewline
${\bf TDRD3}$&$ {\bf CRBL8.L}$&$  {\bf FBXO28}$&$0.5856$&$0.2284$&$0.0022$&$0.0702$&$1.1722$&$0.7667$&$0.0000$&$0.0100$&$0.0052$&$0.0000$\tabularnewline
${\bf TDRD3}$&$ {\bf CRBL8.L}$&$ {\bf ZNF512}
$&$0.5627$&$0.2690$&$0.0000$&$0.0375$&$0.7126$&$0.7087$&$1.3420$&$0.0101$&$0.0196$&$0.0021$\tabularnewline
${\bf TDRD3}$&${\bf  CRBLCrus1.L
}$&$ {\bf WDR37}
$&$0.5287$&$0.1848$&$0.0563$&$0.0201$&$1.4667$&$0.6084$&$0.0000$&$0.0100$&$0.0423$&$0.0076$\tabularnewline
${\bf TDRD3}$&${\bf CRBLCrus2.L}$&$ {\bf DUSP1}$&$0.4862$&$0.2265$&$0.3120$&$0.0235$&$0.0000$&$1.2779$&$0.6463$&$0.0100$&$0.0185$&$0.0047$\tabularnewline
${\bf TDRD3}$&$ {\bf DCG.R}$&$ {\bf EYA4}
$&$0.5874$&$0.1843$&$0.3626$&$0.0010$&$0.5656$&$0.1007$&$0.0000$&$0.0100$&$0.0135$&$0.0028$\tabularnewline
${\bf TDRD3}$&${\bf PCG.R}$&$ {\bf DUSP1}$&$0.8170$&$0.1434$&$0.0746$&$0.0690$&$0.0000$&$0.3873$&$0.0623$&$0.0100$&$0.0082$&$0.0079$\tabularnewline
${\bf TDRD3}$&$ {\bf PCG.R}$&$ {\bf PTGS2}
$&$0.8223$&$0.1481$&$0.1251$&$0.0182$&$0.1414$&$0.5447$&$0.0000$&$0.0100$&$0.0256$&$0.0040$\tabularnewline
${\bf TDRD3}$&$ {\bf ORBsup.L}$&$  {\bf FBXO28}$&$0.5762$&$0.1398$&$0.3331$&$0.0679$&$0.8560$&$0.4102$&$0.0000$&$0.0100$&$0.0014$&$0.0050$\tabularnewline
$TDRD3$&${\bf ORBmid.R}$&${\bf HOXA9}
$&$0.7235$&$0.3465$&$0.0000$&$0.2839$&$0.8353$&$0.5320$&$0.1433$&$0.0100$&$0.0067$&$0.0073$\tabularnewline
${\bf TDRD3}$&${\bf LING.L}$&${\bf ZNF512}
$&$0.6416$&$0.0308$&$0.0000$&$0.1771$&$1.4260$&$0.3854$&$0.0717$&$0.0100$&$0.0312$&$0.0098$\tabularnewline
${\bf TDRD3}$&${\bf CAU.R}$&$  {\bf CCND2
}$&$0.6381$&$0.1076$&$0.0170$&$0.0635$&$0.7395$&$1.2557$&$0.0000$&$0.0100$&$0.0096$&$0.0021$\tabularnewline
${\bf TDRD3}$&${\bf  IPL.L
}$&${\bf ZNF512}$&$0.6093$&$0.2597$&$0.0520$&$0.0380$&$1.0761$&$0.9126$&$0.0000$&$0.0100$&$0.0180$&$0.0018$\tabularnewline
${\bf TDRD3}$&$ {\bf TPOsup.L}$&$ {\bf PLAGL1
}$&$0.6437$&$0.1242$&$0.0000$&$0.1318$&$0.0209$&$0.8248$&$0.3640$&$1.6934$&$0.0374$&$0.0024$\tabularnewline
${\bf TDRD3}$&${\bf MTG.R}$&$ {\bf CCND2
}$&$0.2905$&$0.3650$&$0.0000$&$0.5177$&$0.8603$&$1.2746$&$1.6902$&$0.0100$&$0.0334$&$0.0040$\tabularnewline
${\bf TDRD3}$&${\bf ITG.R}$&$ {\bf CRABP1
}$&$0.5291$&$0.2318$&$0.0000$&$0.0223$&$0.7052$&$0.6038$&$0.6240$&$0.0100$&$0.0033$&$0.0004$\tabularnewline
${\bf TDRD3}$&${\bf ITG.R}$&$ {\bf EDNRB
}$&$0.4778$&$0.3792$&$0.0000$&$0.0265$&$1.0527$&$1.1174$&$0.9571$&$0.0100$&$0.0243$&$0.0085$\tabularnewline
${\bf TDRD3}$&${\bf Vermis10}$&$ {\bf FEN1
}$&$0.3894$&$0.5985$&$0.0000$&$0.0103$&$0.7989$&$1.4191$&$0.9878$&$0.0100$&$0.0377$&$0.0037$\tabularnewline
${\bf TDRD3}$&${\bf Vermis45}$&${\bf PTGS2
}$&$0.7504$&$0.1816$&$0.0718$&$0.0085$&$0.0000$&$0.4176$&$0.3236$&$0.0100$&$0.0121$&$0.0073$\tabularnewline
\hline
\end{tabular}}
\end{center}
\end{table}

%\newpage
 \begin{table}
 \begin{center}
\caption {A part of $31$ genes-derived SNP annotation   using DAVID software.}
\label{tb:pathwayall}
\scalebox{0.8}[.9]{
 \begin{tabular}{ccccccccccc} \hline
\rm{Annotation} &\rm{Database}&\rm{Term}&\rm{No. genes}&\rm{P-Value}&\rm{Benjamini}\\\hline
\rm{Literature} &\rm{Pubmed-ID}& $19367581$ &$12$&$1.6E-22$&$ 5.8E-19$\\%\cline{4-7}
&&$18198266$&$7$&$3.0E-15$& $5.5E-12$ \\%\cline{4-7}
&&$19328558$&$8$&$1.1E-13$& $1.4E-10$ \\%\cline{4-7}
&&$19086053$&$11$&$5.3E-13$& $4.9E-102$ \\%\cline{4-7}
&&$12477932$&$31$&$4.4E-6$& $1.8E-3$ \\%\cline{4-7}
&&$15489334$&$21$&$5.5E-5 $& $.3E-2$ \\\hline
\rm{Disease} &\rm{GADB-disease-class}& $\rm{Schizophrenia}$ &$26$&$1.1E-26$&$4.4E-24$\\%\cline{4-7}
&&$\rm{Cognitive~ function}$&$6$&$4.4E-6$& $8.9E-4$ \\%\cline{4-7}
&&$\rm{Bipolar~ disorder}$&$6$&$1.5E-3$& $9.8E-2$ \\\hline
\rm{Gene-Ontology}&\rm{GOTERM-BP-1}&\rm{multicellular~ organismal process}&$17$&$1.3E-4$&$2.4E-3$\\%\cline{4-7}
&&$\rm{Developmental~ process}$&$13$&$1.7E-3$& $2.4E-3$ \\\hline
\rm{Pathways}  &\rm{KEGG}& $\rm{Neuroactive ~ligand-receptor~ interaction}$ &$6$&$2.0E-3$&$6.6E-2$\\\cline{2-6}
&\rm{PANTHER}& $\rm{P05912}$ &$4$&$7.8E-3$&$2.6E-1$\\%\cline{4-7}
&& $\rm{P00001}$ &$3$&$9.0E-3$&$1.6E-1$\\\cline{2-6}
\rm{Protein Interactions}&\rm{UCSC}& $\rm{HFH3}$ &$22$&$2.5E-5 $&$4.4E-3$\\%\cline{4-7}
&& $\rm{BRN2}$ &$24$&$4.8E-5$&$4.2E-3$\\%\cline{4-7}
&& $\rm{CDP}$ &$24$&$1.6E-3$&$9.0E-3$\\%\cline{4-7}
&& $\rm{GATA}$ &$20$&$1.8E-3$&$9.2E-3$\\\hline	
\end{tabular}	
}	
\end{center}
\end{table}

%\begin{figure}
%\begin{center}
%\label{fig:10ROIs}
 %\includegraphics[width=16cm, height=16cm]{ROI10.eps}
  %\includegraphics[width=\columnwidth]{ROI10.eps}
%\caption{The selected $10$ ROIs on axial view.}
%\end{center}
%\end{figure}

%\begin{figure}
%\begin{center}
%\label{fig:GGN}
 %\includegraphics[width=12cm, height=12cm]{gg51.eps}
  %\includegraphics[width=\columnwidth]{gg51.eps}
%\caption{The  network of the selected genes-derived SNP and genes-derived DNA methylation.}
%\end{center}
%\end{figure}
\end{document}